\documentclass[screen]{acmart}
\AtBeginDocument{%
  \providecommand\BibTeX{{%
    \normalfont B\kern-0.5em{\scshape i\kern-0.25em b}\kern-0.8em\TeX}}}

\setcopyright{acmcopyright}
\copyrightyear{2023}
\acmYear{2023}
\usepackage{subcaption}

\begin{document}

\newtheorem{hyp}{Hypothesis} 

\title[An Information Bottleneck Characterization of the Understanding-Workload Tradeoff]{An Information Bottleneck Characterization of the Understanding-Workload Tradeoff in Human-Centered Explainable AI}

\author{Lindsay Sanneman}
\authornote{Both authors contributed equally to this research.}
\email{lindsays@csail.mit.edu}
\orcid{1234-5678-9012}
\author{Mycal Tucker}
\authornotemark[1]
\email{mycal@mit.edu}
\affiliation{%
  \institution{MIT}
  \streetaddress{77 Massachusetts Ave.}
  \city{Cambridge}
  \state{Massachusetts}
  \country{USA}
  \postcode{02139}
}

\author{Julie A. Shah}
\affiliation{%
  \institution{MIT}
  \streetaddress{77 Massachusetts Ave.}
  \city{Cambridge}
  \country{USA}}
\email{julie_a_shah@csail.mit.edu}

\renewcommand{\shortauthors}{Sanneman and Tucker, et al.}

\begin{abstract}

Recent advances in artificial intelligence (AI) have underscored the need for explainable AI (XAI) to support human understanding of AI systems. Consideration of human factors that impact explanation efficacy, such as mental workload and human understanding, is central to effective XAI design. Existing work in XAI has demonstrated a tradeoff between understanding and workload induced by different types of explanations. Explaining complex concepts through abstractions (hand-crafted groupings of related problem features) has been shown to effectively address and balance this workload-understanding tradeoff. In this work, we characterize the workload-understanding balance via the Information Bottleneck method: an information-theoretic approach which automatically generates abstractions that maximize informativeness and minimize complexity. In particular, we establish empirical connections between workload and complexity and between understanding and informativeness through human-subject experiments. This empirical link between human factors and information-theoretic concepts provides an important mathematical characterization of the workload-understanding tradeoff which enables user-tailored XAI design. 
\end{abstract}

\ccsdesc[500]{Human-centered computing~User studies}
\ccsdesc[500]{Human-centered computing~User models}
\ccsdesc[500]{Human-centered computing~HCI theory, concepts and models}
\ccsdesc[500]{Human-centered computing~Empirical studies in HCI}
\ccsdesc[500]{Mathematics of computing~Information theory}

\keywords{explainable AI, workload, human factors, information bottleneck}



\newcommand{\mycal}[1] {\textcolor{blue}{(Mycal: #1)}}
\newcommand{\kl}{D_{\mathrm{KL}}}

\maketitle

\section{Introduction}
With the rapid development of powerful yet opaque artificial intelligence systems, enabling AI transparency through approaches that effectively explain the outputs of these systems to humans has become increasingly important. Recent advances in explainable AI (XAI), defined as ``AI systems that can explain their rationale to a human user, characterize their strengths and weaknesses, and convey an understanding of how they will behave in the future" \cite{gunning2019darpa}, have aimed to address the problem of AI transparency. Accounting for human factors related to information processing is at the core of producing effective explanations that support human understanding of these AI systems. Such explanations require not only computer science expertise, but also cross-disciplinary efforts with the fields of cognitive science, human factors, and the social sciences, more generally \cite{miller2017explainable, miller2019explanation, sanneman2022situation}.

Human-centered explainable AI research has begun to explore human factors with respect to XAI, both through proposed frameworks \cite{sanneman2022situation, chen2014situation, endsley2023supporting, amershi2019guidelines, liao2020questioning, ehsan2023charting, liao2023ai} and experimental or interview-based analyses \cite{sanneman2022empirical, paleja2021utility, ghai2021explainable, lai2023selective, zhang2020effect, kim2023help}. In particular, recent experiments have drawn upon validated assessments from human factors to study the impacts of XAI on human mental workload, trust, and conceptual understanding \cite{sanneman2022empirical, paleja2021utility, lage2019evaluation, hoffman2018metrics, lai2023selective} and have demonstrated tradeoffs between these factors in explanation design \cite{sanneman2022empirical, paleja2021utility, lai2023selective, ghai2021explainable}. 
Such a tradeoff exists between increasing human workload and supporting human understanding of an AI system: 
one study found that explaining an autonomous agent's goals through hand-crafted abstractions of key problem features effectively balanced human workload and understanding as compared with other XAI techniques~\citep{sanneman2022empirical}. 
Further work in cognitive science has suggested that humans approach complex problem solving through the use of abstract information as well \cite{ho2019value}; 
this suggests that providing abstract representations of key information may be an effective means of providing AI explanations to humans.
However, generating such abstractions automatically and quantifying how they trade off human workload and understanding at the explanation design stage remain open problems.

Concurrent with such XAI and human factors research, recent work in cognitive science has identified the key role of information-theoretic abstractions in human cognition.
For example, in a wide variety of languages and semantic domains, human naming systems are near-optimal according to an Information Bottleneck (IB) tradeoff between maximizing informativeness (how well a listener can reconstruct a speaker's meaning) and minimizing complexity (how many bits about an input are encoded in a word)~\citep{Zaslavsky2018efficient,Zaslavsky2019containers,Mollica2021forms,Zaslavsky2021person}.
Our key insight in this work is connecting concepts from IB literature to analogs in human factors for analyzing explanations.
By connecting notions from human factors, such as workload or understanding, to information-theoretic quantities, such as complexity and informativeness, we may leverage theoretical models for tradeoffs between these terms, while simultaneously using existing IB tools for automated explanation-generation and assessment.

In this work, we leverage IB to automatically generate abstract explanations, and establish empirical connections between the aforementioned human factors constructs and information-theoretic quantities through a set of human-subject experiments. In particular, we consider the relationships between human workload and explanation complexity and between human understanding and a concept closely related to explanation informativeness called distortion (in other words, how well a particular concept can be predicted from a given abstraction of that concept). This lays the groundwork for modeling human factors quantities (and their associated tradeoffs) using the theoretically-rich IB framework, which will enhance our ability to automatically generate and analyze user-tailored explanations that account for differing human informational and workload needs.

Here, we focus on explaining complex functions through abstractions of these functions, since many AI systems — including large language models (LLMs), reinforcement learning-based robotic systems, and machine learning-based recommender systems — are built on complex functions. 
Specifically, we study the problem of explaining reward functions (discussed at length in Section \ref{RW:Reward}) to humans, with a focus on reward functions due to their applicability across many applications \cite{silver2021reward}, such as autonomous agent planning problems \cite{sutton2018reinforcement, russell2010artificial} and reinforcement learning from human feedback (RLHF), which is used to tune LLMs \cite{ziegler2019fine, bai2022training}.

We performed experiments across two domains — a grid-navigation domain and a color-based sample-collection domain — and considered both continuously and discontinuously varying reward functions in 
each.
Our results indicate significant correlations between complexity and human workload as well as distortion and a feature-based measure of human understanding (discussed further in Section~\ref{sec:human_factors}) across both domains.
We also observed significant correlations between a policy-based measure of human understanding and distortion within the grid-navigation domain, but not the color domain, which involved more complex visualizations of abstract explanations.
Together, these findings suggest that the complexity-distortion balance in IB can be effectively applied to model the workload-understanding tradeoff in human-centered XAI design and to generate user-tailored explanations, but that care must be taken to ensure such abstract explanations are visualized effectively.   

\section{Related Work}
Here we provide an overview of the literature related to the human factors constructs of workload and situation awareness as they apply to explainable AI as well as existing XAI approaches to explaining functions (including reward functions) to humans. We also discuss the Information Bottleneck method, which we leverage to automatically generate abstract explanations of reward functions. In our experiments, we hypothesize that the information-theoretic concepts used to generate these abstract explanations correlate with the human factors constructs of workload and situation awareness.


\subsection{Human Factors and Explainable AI}

\subsubsection{Human Mental Workload}
\label{RW:workload}
Mental workload is a construct that has been widely studied in the human factors literature, and can be defined as the relationship between the mental resources demanded by a task and the resources available to be supplied by the human performing the task \cite{parasuraman2008situation}. It has been researched in domains such as aviation \cite{brand2021workload, katins2023exploring}, healthcare \cite{tubbs2018nasa, said2020validation}, usability in human-computer interaction \cite{longo2018experienced}, and workplace tasks \cite{midha2021measuring}, among others \cite{longo2022human}, and has been shown to correlate with task performance across a variety of settings  \cite{wright2016agent, chen2016effects, chen2018situation, chen2017situation}. Various models of mental workload have been proposed, such as the widely cited and applied multiple resource model (MRM) \cite{wickens2008multiple}, which categorizes human cognitive resources into different ``pools'' that are available for information processing and can be filled independently. These pools are defined according to four dimensions: the modality of information representation presented to the human, the form of information encoding in the brain, stages of information processing, and response modality. 

In the context of XAI, addressing workload considerations through application of models such as the MRM can help to ensure that explanations are effective and that the information they communicate is comprehensible to the recipient of the explanation. If XAI systems communicate too much information, a person may process only partial information (or none at all) from the explanation. At the same time, explanations must provide adequate information in order to be useful for the task at hand. 
A person's available mental capacity, therefore, informs both the choice of how much information to include in a given explanation and the frequency with which to provide explanations in effective XAI design \cite{sanneman2022situation}. A person's context, including all other tasks they must perform when they receive an explanation, must also be considered in explanation design.
Beyond this, accounting for individual differences in baseline cognitive capacity between people is critical to supporting human task performance \cite{yu2014individual, wright2014individual}. 

The impact of XAI on workload has been widely researched, with some studies finding that the addition of AI transparency reduces workload due to increased access to critical information \cite{chen2014situation, schaefer2017communicating, zakershahrak2020online}, some finding that additional information provided by XAI systems increases workload \cite{sanneman2022empirical, lai2023selective, ghai2021explainable, paleja2021utility}, and others finding little impact of XAI on workload \citep{wright2016agent, chen2016effects, chen2018situation, chen2017situation}. 
In these cases, workload was measured according to a variety of objective and subjective assessments. Objective measurements quantify workload in terms of a person's performance of primary and secondary tasks with varying cognitive demands \citep{wickens2000attention}; subjective assessments often draw upon participant responses to workload scales, such as the widely applied and validated NASA TLX scale, which asks respondents to answer a set of Likert scale-based questions about their workload after completing a task \citep{hart1988development}. In addition to these assessments, physiological measures of heart rate, skin-based properties, and ocular movements have also been applied to measure workload \cite{charles2019measuring}, although these assessments tend to be more invasive and difficult to implement outside of experimental settings. 
In this paper, since workload has been shown to be a critical factor in the design of effective XAI systems, we study how workload relates to an information-theoretic concept of complexity, and how this can inform explanation design.

\subsubsection{Situation Awareness and Human Understanding}
\label{RW:SA}
Situation awareness (SA) is a human factors construct commonly defined according to the three-level framework proposed by \citet{endsley1995}: the perception of elements in the environment within a volume of time and space (level 1), the comprehension of their meaning (level 2), and the projection of their status in the near future (level 3). 
It is, essentially, a structured way of defining a person's understanding of a given scenario and their ability to use this information to perform required tasks within the given context. 
In other words, the quality of a person's context-related understanding can be assessed according to how well that person perceives, comprehends, and projects the status of relevant elements of their environment.

Since SA defines a person's contextual informational needs, it has been also applied to the development of transparency frameworks for defining the informational requirements for XAI users that XAI systems must meet through the information they explain \cite{chen2014situation, sanneman2022situation}. Within the field of human factors, validated measures such as the Situation Awareness Global Assessment Technique (SAGAT), which probes a user's understanding of critical information at various points throughout that user's performance of a task \cite{endsley1995, endsley2017direct, endsley1988situation}, have been proposed and leveraged for SA evaluation. Through application of these measures, SA has been shown to correlate with task-related measures such as performance and error frequency \citep{endsley2015situation}.

At the same time, XAI researchers have sought ways of effectively assessing user understanding of AI decision making processes as a measure of explanation efficacy \cite{lage2019evaluation, hoffman2018metrics, miller2019explanation}. 
To date, understanding has been assessed through a variety of means (most of which have not been validated through human-subject experiments), such as scales for explanation goodness \cite{hoffman2018metrics, lage2019evaluation} and a user's ability to simulate an agent's optimal behavior or next decision \cite{lage2019evaluation, huang2019enabling}.
Since the SA construct and SAGAT have been validated, a SAGAT-like approach can readily be applied to measure understanding in the context of XAI; however, the question of how to effectively probe a person's understanding of abstract concepts such as reward functions within the context of SAGAT remains.  One recent study employed human-subject evaluations to validate reward alignment metrics that could be applied to measure a person's understanding of a reward function in the context of XAI \cite{sanneman2023validating}. We leverage a subset of these metrics (discussed further in Section \ref{sec:human_factors_understanding}) in our analysis to study how a person's understanding relates to an information-theoretic concept of informativeness (which we call distortion), and how this can be leveraged to design explanations that effectively trade off workload and understanding through the proxy measures of complexity and informativeness.

\subsection{XAI Approaches to Explaining Functions}
Many existing approaches to XAI strive to explain the often-complex functions that characterize models used in AI. For example, \textit{feature importance techniques} explain the most important predictive features in regression \cite{ribeiro2016should, lundberg2017unified}, \textit{saliency maps} unveil information about the gradients of the functions that characterize neural networks and related models \cite{adebayo2018sanity, huber2021local, michaud2020understanding}, and \textit{rationalizations} summarize agent policies that can be computed based on agent reward functions \cite{ehsan2019automated, ehsan2018rationalization}, among others.
While some approaches have accounted for human workload by enabling explanations to provide variable amounts of information to users depending on their cognitive capacity \cite{ribeiro2016should, chakraborti2019plan, sreedharan2018hierarchical}
to our knowledge, none has formally accounted for the tradeoff between human workload and understanding by leveraging mathematical characterizations of these human factors constructs in XAI design. In this paper, we empirically demonstrate the links between these constructs and mathematically-grounded information-theoretic concepts, which enables the automatic generation of explanations that trade off workload and understanding differently depending on user needs and capacities.
We focus specifically on explanations of reward functions, which characterize desired autonomous agent behaviors in sequential decision-making problems such as reinforcement learning. 

\subsubsection{XAI and Reward Functions}
\label{RW:Reward}
Reward functions are one of the primary components of Markov decision processes (MDPs), which are often used to model autonomous agent planning problems \cite{sutton2018reinforcement}. Within an MDP, the reward function characterizes the reward an agent receives for taking different actions from different states; in other words, reward functions dictate what optimal agent behavior (often referred to as the agent's policy) will look like within a given domain. 

Reward functions are often defined as follows:
\begin{equation}
\label{reward}
    R(s) = \omega^{T} \Phi(s).
\end{equation}

Here,  $\Phi(s)$ is a set of features whose values can be calculated based on the agent's state in the world ($s$), and $\omega$ is a set of weights indicating the trade-offs between these features.

Within existing XAI literature, reward functions have been explained through means including policy summaries which demonstrate roll-outs of optimal agent behavior originating from a variety of world states \cite{hayes2017improving, amitai2023asq, amir2018highlights, amir2018agent}, language-based rationalizations of agent policies \cite{ehsan2018rationalization, ehsan2019automated}, techniques that reconcile a human's reward function with that of an agent \cite{tabrez2019explanation}, counterfactual demonstration-based explanations of key reward features \cite{lee2022reasoning}, and decompositions of interpretable reward components provided to human users \cite{anderson2019explaining, juozapaitis2019explainable}, among others. One recent study found that explaining reward functions through abstractions of reward features effectively balanced a workload-understanding tradeoff among different reward explanations \cite{sanneman2022empirical}. Since the abstraction-based approach proved effective in that study, we also evaluate abstract explanations of reward features in this work.

\subsection{Information Bottleneck}

We leverage methods from Information Bottleneck (IB) literature to formalize a tradeoff between complexity and reward distortion, which we then connect to human factors.
In canonical IB settings, one seeks to generate (lossy) representations, $Z$, of inputs, $X$, which are used to predict a downstream quantity, $Y$~\citep{Tishby1999,alemi2016deep}.
In this work, we only consider predicting a reward, $Y$, from features, $X$, but the IB framework is more widely applicable.
The IB maximization problem is formulated as a tradeoff between two information-theoretic terms:

\begin{equation}
    \text{maximize} \quad I(Y; Z) - \beta I(X; Z) 
\end{equation}

where $\beta$ is a scalar parameter, $I(Y; Z)$ is the informativeness (measured as the number of bits about the reward $Y$ retained in $Z$), and $I(X; Z)$ is the complexity (measured as the number of bits about the features $X$ in $Z$).
In other words, the IB formulation seeks to maximize informativeness while minimizing complexity.
Notably, there is a theoretical limit for the maximum informativeness for a given complexity, but this limit shifts as a function of $\beta$.
In our work, since we aim to explain reward functions, $X$ is the features of the reward function, $\phi(s)$, $Y$ is the reward value, and $Z$ are the abstract representations grouping $X$ to predict $Y$.
Therefore, as $\beta$ increases and complexity decreases, the above optimization will group features with similar rewards in the same abstraction.
Lastly, we note that IB work is closely related to rate distortion theory where, rather than computing informativeness ($I(Y; Z)$), one measures the distortion, or error, in predicting $Y$ from $Z$~\citep{Zaslavsky2018efficient}.
In our work, we measure the distortion in predicting a reward value, which we dub \textit{reward distortion}.

Beyond a purely mathematical formulation, several works in the fields of cognitive science, psychology, and behavioral economics have investigated aspects of IB tradeoffs in human cognition.
Across domains and languages, naming systems (e.g., words for colors) are nearly perfectly efficient in the IB sense: maximizing the ability of listeners to reconstruct a speaker's meaning at a given complexity level~\citep{Zaslavsky2018efficient,Zaslavsky2019containers,Mollica2021forms,Zaslavsky2021person}.
In vision-based domains, people similarly create compressed representations of images via sketches that capture functionally useful details at the expense of visual fidelity~\cite{sketches}; this type of behavior is consistent with an IB system under complexity constraints.
In economics, recent research points to the importance of information constraints in human behavior~\citep{aridor2023information}.
Even within XAI, \citet{bang2021explaining} briefly explored the role of penalizing complexity to create more ``interpretable'' AI models for humans to understand, but they only used a fixed complexity in experiments.
This evidence, collected across multiple fields, suggests that IB tradeoffs play an important role in human cognition; in our work, we connect notions from IB to human factors measures of explanation understanding.

\section{Research Aims}
In this paper, we aim to establish an empirical link between human factors concepts that are relevant to the design of effective explainable AI systems and information-theoretic concepts. 
This link will enable a mathematical characterization of tradeoffs occurring between relevant human factors in XAI design, as well as provide tools to automatically generate abstraction-based explanations which trade off these factors, which is useful for meeting the varying informational and workload needs of individual users of AI systems.
Specifically, we perform human-subject experiments in order to validate an information-theoretic measure of explanation complexity as an indicator of human workload and an information-theoretic measure of reward distortion as an indicator of human understanding. We then detail the human factors concepts and information-theoretic metrics we studied in greater detail, along with the hypotheses assessed in our experiments.

\subsection{Measures of Human Workload and Human Understanding from the Field of Human Factors}
\label{sec:human_factors}
\subsubsection{Workload}
As discussed in Section \ref{RW:workload}, multiple measures of human workload (both objective and subjective) have been applied to the study of workload within the field of human factors. One of the measures most commonly applied in the literature is the NASA Task Load Index (TLX) scale \cite{hart1988development}, which has also been applied to the study of techniques for explainable AI  \cite{zakershahrak2019online, sanneman2022empirical, paleja2021utility,lai2023selective, ghai2021explainable}.  
We therefore used the NASA TLX scale to assess workload in our own set of experiments.

\subsubsection{Human Understanding} 
\label{sec:human_factors_understanding}
While a number of approaches have been proposed for assessing human understanding in the context of XAI \cite{lage2019evaluation, hoffman2018metrics, huang2019enabling}, as discussed in Section \ref{RW:SA}, few have been validated through human-subject experiments. One recent set of experiments validated reward alignment metrics which capture the similarity between a human's reward function and that of an autonomous agent \cite{sanneman2023validating}; these metrics can be applied to study either how aligned an agent's reward function is with a human's after a reward-learning process on the one hand, or how aligned a human's understanding of an agent's reward function is with the agent's true reward after an explanation of the reward is provided on the other. 
As our aim is to study human understanding of reward functions resulting from the provision of reward explanations at different levels of abstraction, we consider the latter metric in this work. 

\citet{sanneman2023validating} identified two categories of alignment in their experiments: feature alignment, which captures how aligned human and agent reward features and weights are; and policy alignment, which captures how aligned human and agent policies corresponding to these reward functions are. We leverage one of the validated metrics from each of these categories in our assessments of human understanding.

The feature understanding metric we apply in this work is a similarity metric called \textit{feature ranking}, which can be defined as follows: 
\begin{equation}
    FR = \dfrac{(W_H \cap W_{GT})}{(W_H \cup W_{GT})}
\end{equation}

Here, $W_{GT}$ is the set of pairwise comparisons of the magnitudes of the weights $w$ of a set of reward features, $\phi(s)$, as in Equation \ref{reward} (e.g., one of these comparisons could be $w_A > w_B$, where $w_A$ is the weight of feature $A$, which is higher than $w_B$, the weight of feature $B$). $W_{GT}$ specifically captures the pairwise rankings of the feature weights in the ground truth reward function, and $W_H$ is the set of pairwise rankings from the human's reward function. In our experiments, we assessed the human's reward function by asking the human participants to rank a set of predetermined features in order of importance according to their understanding of the reward function upon receiving an abstract explanation of this reward.
(We include examples of this assessment in Appendix \ref{app:survey_questions}.) The \textit{feature ranking} metric is the intersection over union of the ground truth rankings and the human's rankings, and thus captures the similarity between how important the human believes a set of reward features are relative to each other versus the ground truth relative importance of each feature. 

The policy understanding metric we apply is a regret-based metric called \textit{best demonstration}, defined as follows:
\begin{equation}
    BD = 1 -  \dfrac{R(\xi^{*}) - R(\xi^{H})}{R(\xi^{*}) - R(\xi^{-})}
\end{equation} 

Here, $\xi^{*}$ is the optimal demonstration (i.e., a set of state-actions pairs) of a given task according to a ground truth reward function — which, in our case, is the reward function being explained. $\xi^{H}$ is the human's best demonstration according to the reward function they understood from the explanation. Finally, $\xi^{-}$ is the worst possible demonstration in terms of the ground truth reward, which we assume to be calculable for a finite-horizon task. $R(\cdot)$ evaluates each of these trajectories according to the ground truth reward function. 

In order to evaluate this metric in our experiments, we ask human participants to provide their optimal demonstration of a task given their understanding of the associated reward function after having received an explanation of it. (Examples of this assessment are also included in Appendix \ref{app:survey_questions}.) The \textit{best demonstration} metric essentially captures how close the human's reward function is to the ground truth reward in terms of the policies that result from these rewards for a given task.

\subsection{Information-Theoretic Measures of Explanation Complexity and Reward Distortion}

\subsubsection{Complexity}
Drawing upon prior literature, we define \textit{complexity} as the mutual information between an input, $X$, and an abstraction, $Z$~\citep{Zaslavsky2018efficient,tucker2022trading}.
This measure is defined via the Kullback–Leibler divergence of the conditional distribution of $Z$ given $X$ from the prior over $Z$:

\begin{equation}
    I(X; Z) = \kl[\mathbb{P}(Z|X)\|\mathbb{P}(Z)].
\end{equation}

Complexity is minimized at 0 if all $X$ are represented via the same $Z$; beyond such uninformative representations, more complex representations include additional information about $X$ in $Z$.
For example, more complex color naming systems use more distinct words: naming systems using the word ``crimson'' convey more information about a precise color than naming systems that only use less specific words like ``red''~\citep{Zaslavsky2018efficient}.
More generally, in IB literature, by penalizing the complexity of representations, one imposes a bottleneck on how much information is stored in $Z$, which in turn induces lossy representations that do not enable perfect reconstructions of $X$ from $Z$~\citep{Tishby1999,alemi2016deep}.
In our work, we both use existing IB methods to generate representations across a spectrum of complexity, as well as calculate complexity as a metric which we then relate to human workload.

\subsubsection{Reward Distortion}

We define reward distortion as a measure for how well one can predict a reward value, $Y$, from an abstract representation, $Z$.
Formally, we measure reward distortion as the minimum mean squared error (MSE) for predictions of $Y$ from $Z$:

\begin{equation}
    D(Z; Y) = \frac{1}{|Y|} \sum_{(X, Y)} || \hat{Y}(Z(X)) - Y ||^2
\end{equation}

where, assuming access to a dataset of reward features $(X)$ and rewards $(Y)$, $Z(X)$ represents the abstraction generated from $X$, and $\hat{Y}(Z(X))$ represents the optimal prediction of $Y$ given $Z(X)$.
With a small set of discrete representations, $Z$, computing an optimal predictor is equivalent to traditional methods for MSE regression.
We note that reward distortion is similar to notions of informativeness ($I(Y; Z)$) from traditional IB literature (see \citet{Zaslavsky2018efficient} for discussion of connections to rate distortion theory).
Informativeness measures, however, do not explicitly represent how some prediction errors are better or worse than others.
Given the continuous nature of reward values, we therefore use reward distortion as our preferred metric.
In our paper, we seek to connect reward distortion to \textit{feature rank (FR)} and \textit{best demonstration (BD)} metrics of human understanding.
Given that we always measure the distortion in predicting reward, we at times refer to \textit{reward distortion} simply as \textit{distortion}.

\subsection{Hypotheses}
We investigated the impact of varying complexity and distortion on human workload and human reward understanding by evaluating the following hypotheses:

\begin{hyp}
    The distortion of the abstraction-based explanations will be negatively correlated with human reward understanding, including both feature and policy understanding.
\end{hyp}

\begin{hyp}
    The complexity of the abstraction-based explanations will be positively correlated with human mental workload. 
\end{hyp}

Jointly, these hypotheses state that decreasing distortion will improve human understanding (H1), but increasing the complexity of abstractions will result in greater workload (H2). In other words, we aim to evaluate whether (the inverse of) distortion can serve as a suitable proxy for human understanding and whether complexity can serve as a suitable proxy for human workload in explanation design. 
Given theoretical bounds from IB literature showing a minimum distortion for a given complexity, this suggests optimal explanations will also trade off these two competing factors: minimizing distortion to achieve a desired level of understanding, while subject to bounds on workload (and therefore complexity).

\section{Methodology}

\subsection{Domains}
\label{domains}
We leveraged two domains for our set of experiments: a grid-based navigation domain and a color domain.
\begin{figure}
    \centering
    \begin{subfigure}[c]{0.31\textwidth}
        \centering
        \includegraphics[trim={0cm 1.5cm 0cm 1.5cm},clip=true,width=\textwidth]{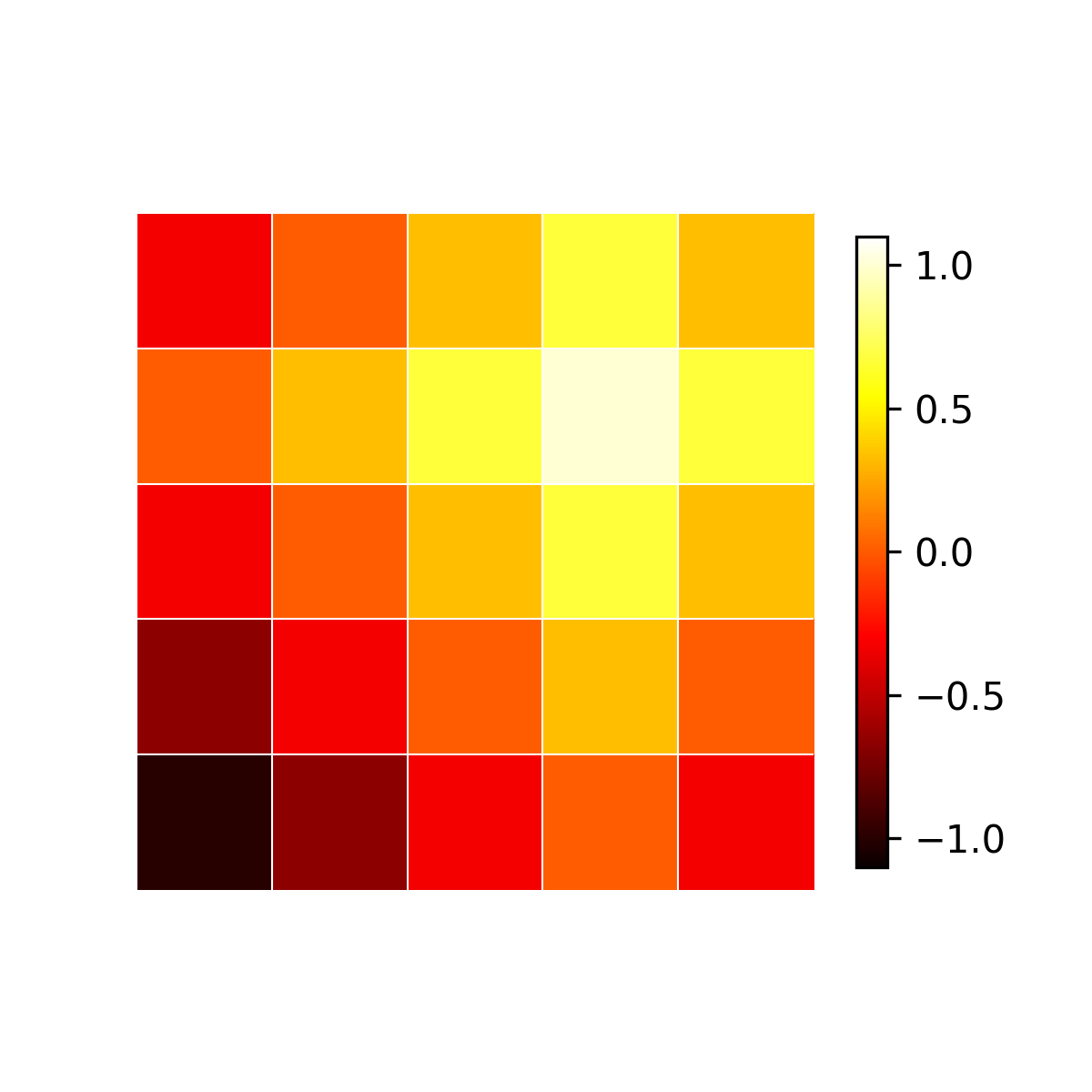}
        \caption{Manhattan Grid}
    \end{subfigure}
    \begin{subfigure}[c]{0.31\textwidth}
        \centering
        \includegraphics[trim={0cm 1.5cm 0cm 1.5cm},clip=true,width=\textwidth]{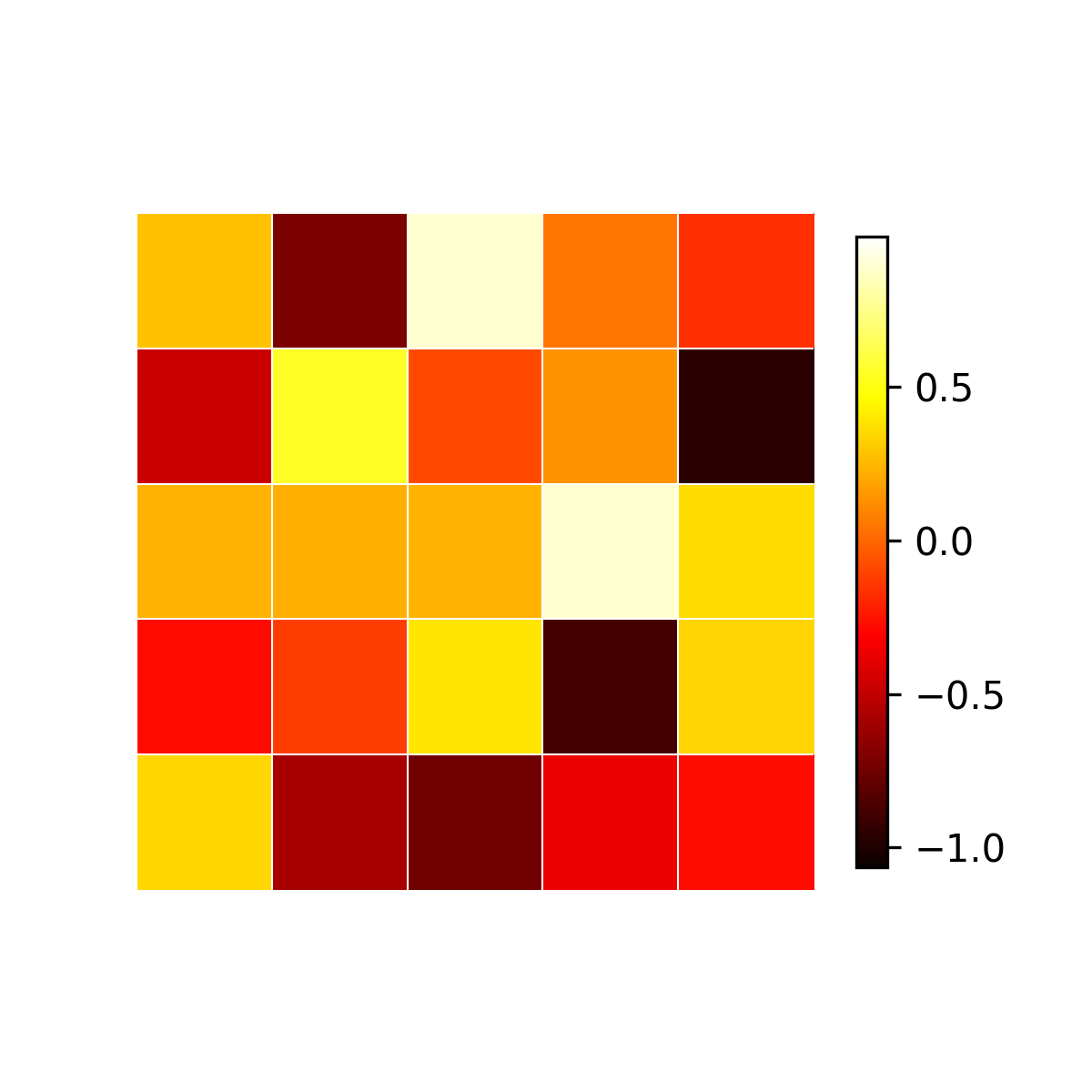}
        \caption{Random Grid}
    \end{subfigure}
    \begin{subfigure}[c]{0.31\textwidth}
        \centering
        \includegraphics[trim={0.5cm 0.25cm 0.5cm 0.25cm},clip=true,width=\textwidth]{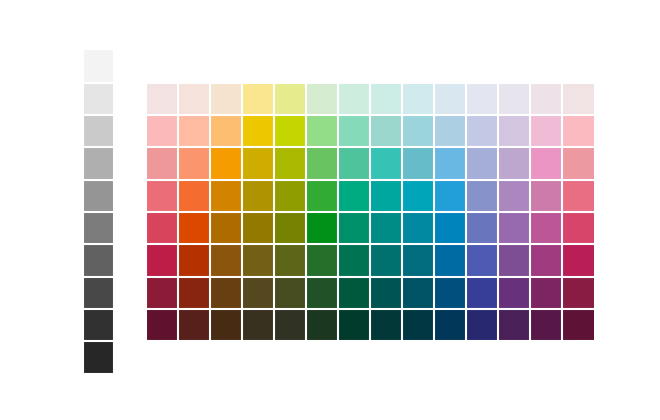}
        \caption{Color Domain}
    \end{subfigure}
    \caption{We conducted experiments in three domains. In the two grid-based domains (a-b), reward was based on location: either Manhattan distance from a fixed point, or randomly-distributed reward. In the color domain (c), reward was based on the blue value in a color's RGB representation. }
    \label{fig:domains}
\end{figure}

\subsubsection{Grid Navigation Domain}
In the grid-navigation domain, different reward values between -1 and +1 were assigned to squares in a 5x5 grid, as depicted in Figure~\ref{fig:domains}. The task was to navigate between a start square and goal square while maximizing the reward accumulated along the path. In this set of experiments, we considered two different reward functions: the first, the ``Manhattan grid'' depicted in Figure~\ref{fig:domains}~(a), had a maximal reward value of +1 at one of the grid squares, with the reward values of other squares in the grid decreasing according to the Manhattan distance from that square. In the second, the ``random grid'' depicted in Figure \ref{fig:domains}~(b), the reward for each location was sampled uniformly at random within the range $[-1, 1]$. We studied these two reward functions as examples of: (1) a continuously varying reward function where adjacent grid regions have similar reward values (the Manhattan grid); and (2) a discontinuously varying reward function, where adjacent grid regions do not necessarily form natural abstraction groups (the random grid). All abstract representations of the grid regions were provided as heat maps, as shown in Appendix \ref{app:abs_viz}

\subsubsection{Color Domain}
\label{sec:color_domain}
In the color domain, we applied continuous and discontinuous reward functions to the colors depicted in Figure~\ref{fig:domains}~c. 
We drew the colors from the World Colorchip Survey (WCS) dataset~\citep{wcs}, and represented them via their RGB values. 
Notably, prior literature has identified how languages represent colors at different abstraction levels, which led us to hope that abstraction-based explanations could improve human understanding of color-based reward functions within this domain~\citep{Zaslavsky2018efficient}.
For the continuous reward function, we set each color's reward equal to that of the blue value in the RGB representation (between 0 and 1). 
For the discontinuous reward function, we used a hand-specified function that divided the blue values into eight bins, which we assigned different rewards between -1 and 1; the exact function is in Appendix~\ref{app:color}. 
Thus, the color domain largely mirrored the grid navigation domain by establishing both continuous and discontinuous reward functions.
The task that participants needed to perform for the \textit{best demonstration (BD)} assessment in the color domain was a sample collection task, where the objective was to navigate through a grid and maximize the total value of samples collected along the path. The samples were represented by one of the colors in the original color grid (shown in Figure \ref{fig:domains}~c), and abstract representations of the color regions were provided as heat maps, as shown in Appendix~\ref{app:abs_viz}. 

\subsection{Information-Bottleneck Explanation Generation}
\label{sec:ib_method}
We used existing methods from prior literature to generate explanations of our domain reward functions at different complexity and distortion levels.
The \texttt{embo} package provides an implementation of the deterministic information bottleneck, which, given a joint distribution $\mathbb{P}(X, Y)$, generates abstractions of inputs $X$ over the range of possible complexities~\citep{embo,strouse2017deterministic}.
For each domain, therefore, we computed this joint distribution by iterating over all possible inputs $X$ (e.g., the $(x, y)$ location of a cell in the grid world) and computing the associated reward $Y$ (e.g., the value at that location in the grid world).
Further details of this process are included in Appendix~\ref{app:ib_impl}.

\begin{figure}
    \centering
    \includegraphics[trim={0cm 2.5cm 0cm 5cm},clip=true,width=\textwidth]{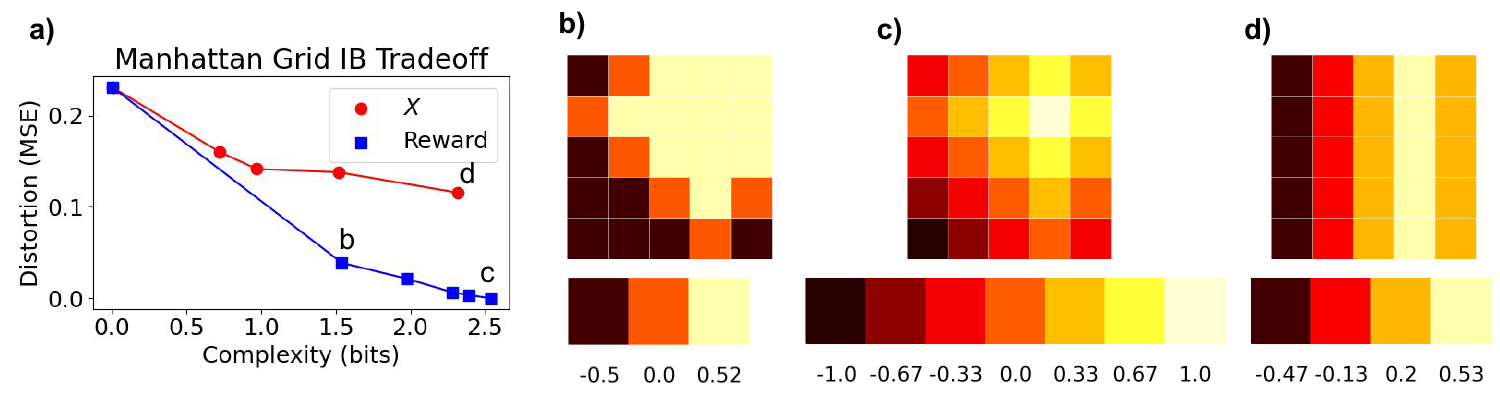}
    \caption{Complexity-distortion curves (\textbf{a)}) and corresponding abstractions (\textbf{b-d}) for the Manhattan grid navigation domain. Using the true grid reward leads to low distortion as complexity increases (blue ``Reward'' curve), and more fine-grained abstractions (b-c), eventually recovering the underlying reward grid. Generating abstractions to recover the $x$ coordinate in the grid (red ``X'' curve and d), rather than the reward, led to higher distortion due to abstractions that did not align with the true reward structure.}
    \label{fig:grid_full}
\end{figure}

For a given reward function, the IB method generated abstractions representing different optimal solutions trading off distortion and complexity; we dubbed such abstractions the ``reward-optimal'' abstractions.
Increasing the complexity of reward-optimal abstractions led to more fine-grained abstractions and lower distortion, as shown in Figures~\ref{fig:grid_full}~b (low complexity) and c (high complexity).
In our experiments, however, we wished to explore the effects of varying distortion and complexity independently.
Therefore, in addition to the reward-optimal abstractions, we generated additional abstractions using different ``training objectives:'' alternate reward functions, which were not necessarily relevant to structure of the reward function being explained.
The abstractions generated by these alternate training objectives resulted in higher distortion (with respect to the reward function being explained) for the same complexity as the reward-optimal abstractions.

For example, in the Manhattan grid, one training objective we used was predicting the $x$ location of each cell as the reward value.
Abstractions generated from this function represented vertical strips in the grid, as shown in Figure~\ref{fig:grid_full}~d.
At the same time, we evaluated the distortion of such abstractions by measuring the MSE in predicting the actual reward in the Manhattan grid.
Unsurprisingly, the $x$-location abstractions led to high distortion in predicting reward.
Figure~\ref{fig:grid_full}~a shows how, in general, using $x$-based abstractions led to higher distortion, for the same complexity, than using reward-optimal abstractions based on the true Manhattan reward.

The same trends held in the color domains as well, where we generated reward-optimal abstractions based on the continuous or discontinuous reward functions of a color's blue value (from its RGB representation), as discussed in Section \ref{sec:color_domain}.
To generate non-reward-optimal baseline abstractions in these cases, we used the color's red value (again, from RGB) as an alternate training objective; such abstractions were not useful in predicting the true reward value (which depended only on the blue value), leading to high distortion regardless of complexity.
(See Figures~\ref{fig:app_color_full} for complexity-distortion curves in the color domain.)
Overall, by using a variety of training objectives to generate abstractions, we could test explanations using abstractions at the same complexity, but different distortion, levels.
Appendix~\ref{app:abs_viz} includes examples of abstractions at different complexity levels, for different training objectives, in all domains.

\subsection{Experiment Design}
In order to empirically study the relationships between abstraction complexity and human cognitive workload and between distortion and human understanding, we performed human-subject experiments in both the grid navigation and color domains. 
In each domain, we studied explanations of two different types of reward function (continuous and discontinuous), as discussed in Section \ref{domains}. For each type of reward function, we generated a set of abstract explanations spanning a range of complexities and distortions. (We include examples of abstractions at different complexity and distortions in Appendix~\ref{app:abs_viz}.) 
The independent variables were the complexities and distortions of the abstractions; 
dependent variables included both human mental workload (as measured via the NASA TLX scale) and reward understanding, measured according to \textit{feature ranking (FR)} and \textit{best demonstration (BD)} assessments.

\subsection{Procedure}
For each domain, participants were first asked a set of demographic questions, including a question about whether they were colorblind, since successful performance of the provided tasks relied upon interpretation of color. Next, participants received an overview of the experiment. In order to reduce learning effects, they were also presented with a set of example abstract explanations within the given domain, along with corresponding examples of correct responses to the two reward understanding questions that they were asked throughout the experiment (\textit{feature rank} and \textit{best demonstration} as detailed in Section \ref{sec:human_factors_understanding}). 

Following the examples, participants were presented with two different scenarios in the given domain. In the grid domain, one scenario involved the Manhattan grid (based on a continuously varying distance from a point), and one scenario involved the random grid (a fundamentally discontinuous reward function). In the color domain, one scenario involved the continuous reward function, while the other involved the discontinuous reward function. The order of presentation of scenarios was counterbalanced for each experiment across all participants.

For each scenario, a participant was shown abstractions based upon a combination of training objective and complexity level.
In the color domain, abstractions were selected from the set of three training objectives (continuous blue, discontinuous blue, and red) and five possible complexities , for a total of 15 possible abstractions. The five complexity values were chosen such that the abstract colors regions spanned a range from one to eight regions. In the grid navigation domain, there were similarly three training objectives (Manhattan or random reward, $y$-based reward, and $x$-based reward) with five possible levels of complexity, for a total of 15 possible abstractions. The number of abstract color regions again spanned a range of one to eight.
Since all abstractions trained according to the same objective were the same regardless of underlying reward function, there was overlap in the set of possible abstractions for each reward function in both domains (e.g., the abstractions trained to predict the ‘’X” values in the navigation domain were the same for both the random and Manhattan reward functions). Therefore, we ensured no participant received the same abstraction across both scenarios.

Following each scenario, participants were asked the feature rank and best demonstration questions; they were also asked the six NASA TLX scale questions in order to assess cognitive workload, along with seven additional questions related to subjective assessment of the abstract explanation quality, which were adapted from a scale for team fluency proposed by \citet{hoffman2019evaluating}. 
At the end of the experiment, participants were asked a set of open-ended feedback questions about the experiment, including what they found to be most challenging and whether they had additional feedback to provide about the experiment. In addition, we asked two attention-check questions during the survey: one before the two scenarios, and another immediately after.

We administered both experiments through the Qualtrics platform, and recruited our participants via the Prolific platform. Participants received no time limit, and took an average of 28 minutes to complete the color survey and of 19.5 minutes to complete the grid navigation survey. They were compensated with \$7 for completing the grid navigation survey and \$10 for completing the color survey, with bonus payments of \$20 provided to the highest-performing participants in each case. As this experiment was survey-based and involved minimal risk, it qualiﬁed for exempt human-subject evaluation status according to the policies outlined by the institutional review board (IRB) at the university where this research was conducted.

\section{Results}

\subsection{Grid Navigation Domain}
Fifty-one participants completed the grid navigation survey (20 women, 30 men, and 1 non-binary individual). The median age was 37 years (min=19 years, max=76 years). Data from six participants was omitted from analysis due to failed attention-check questions or incomplete responses.
We first analyzed the Spearman correlations between distortion and understanding and between complexity and workload for the each of the Random and Manhattan grids separately. We leveraged Spearman correlations due to the non-normality of the underlying datasets in this analysis, as well as the expected monotonic relationships between the correlated variables. We then analyzed the combined Random and Manhattan grid data through a linear mixed-effects analysis to account for individual differences in participants’ responses to the reward understanding and workload questions. 

\begin{figure}
    \centering
    \begin{subfigure}[c]{0.31\textwidth}
        \centering
        \includegraphics[trim={0cm 0cm 0cm 0cm},clip=true,width=\textwidth]{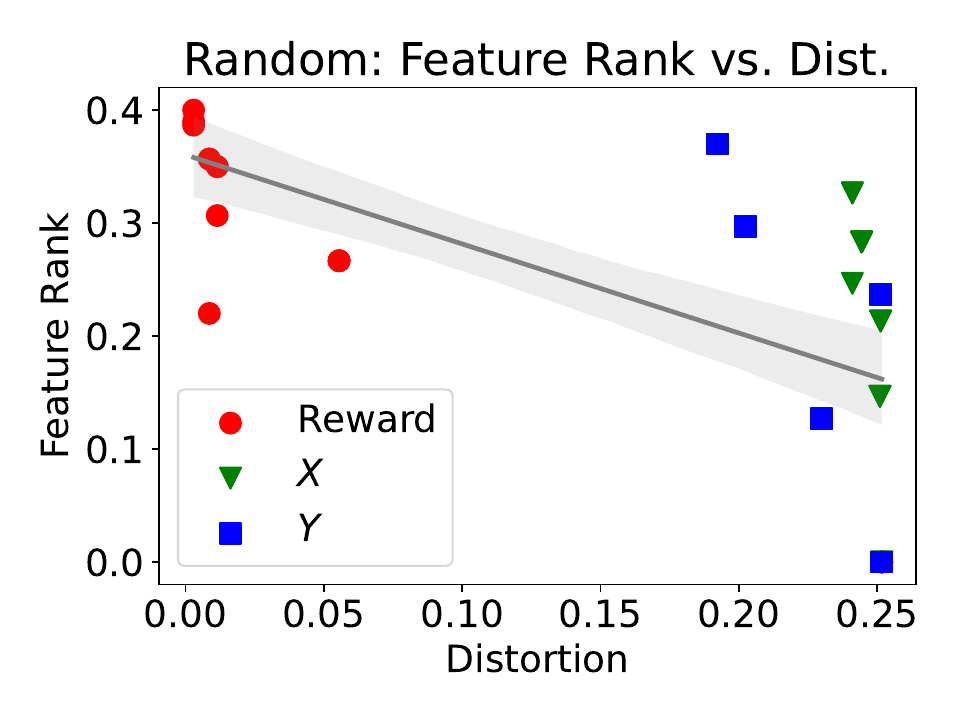}
        \caption{Feature Rank vs. Distortion}
    \end{subfigure}
    \begin{subfigure}[c]{0.31\textwidth}
        \centering
        \includegraphics[trim={0cm 0cm 0cm 0cm},clip=true,width=\textwidth]{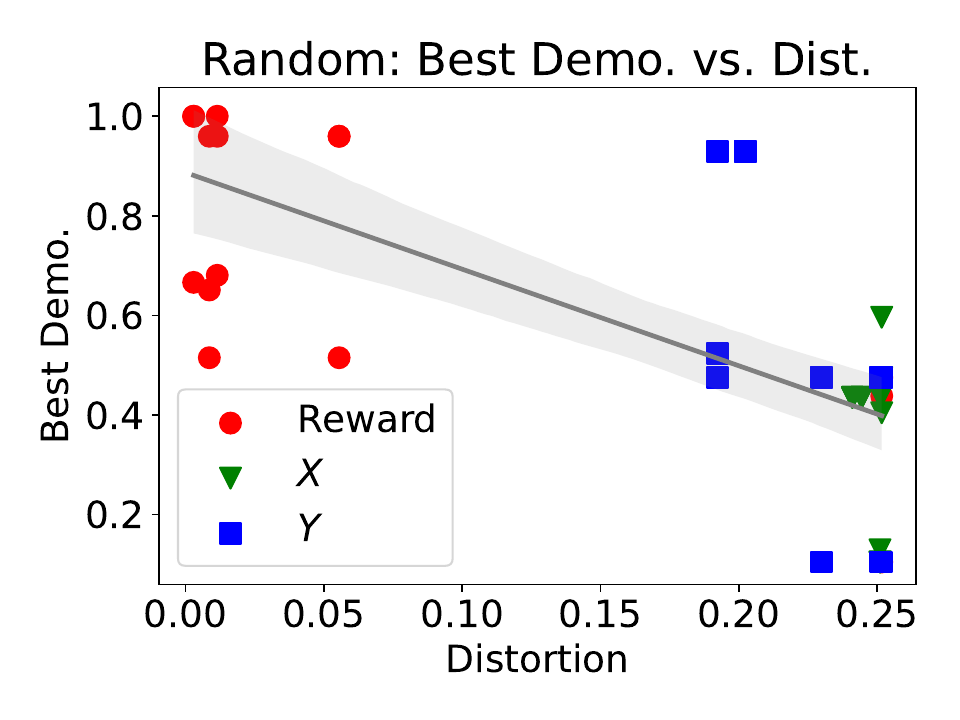}
        \caption{Best Demonstration vs. Distortion}
    \end{subfigure}
    \begin{subfigure}[c]{0.31\textwidth}
        \centering
        \includegraphics[trim={0cm 0cm 0cm 0cm},clip=true,width=\textwidth]{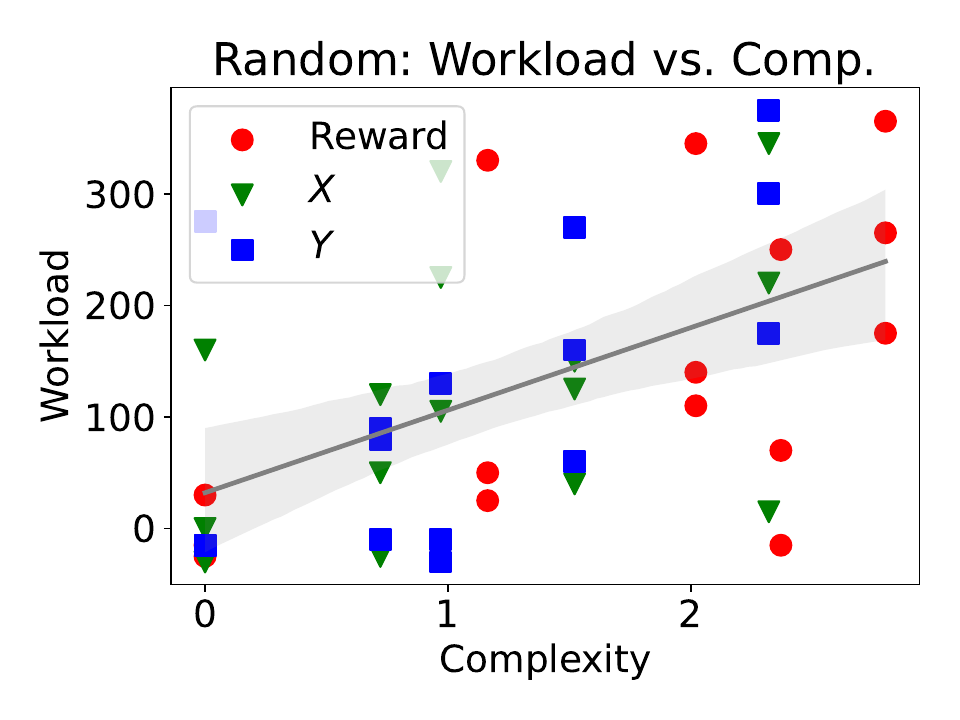}
        \caption{Workload vs. Complexity}
    \end{subfigure}
    \caption{Results from the Random grid domain. As distortion increased, explanation understanding, as measured by Feature Rank (a) or Best Demonstration (b), decreased. At the same time, as complexity increased, workload increased (c).}
    \label{fig:random_results}
\end{figure}

We present results from the Random grid domain in Figure~\ref{fig:random_results}. 
Figures~\ref{fig:random_results}~a and b demonstrate that both feature rank ($FR$) and best demonstration ($BD$) scores were negatively correlated with reward distortion. 
Intuitively, this supports our hypothesis that metrics of understanding would decrease as reward distortion increased (H1). 
Quantitatively, these results were significant: using the Spearman correlation coefficient, we found $\rho(\text{FR}, \text{Dist}) = -0.83, \hspace{0.5em} (p < 0.001)$ and $\rho(\text{BD}, \text{Dist}) = -0.68, \hspace{0.5em} (p < 0.001)$.
At the same time, Figure~\ref{fig:random_results}~c shows a significantly positive correlation between workload and complexity ($\rho(\text{Work.}, \text{Comp.}) = 0.53,\hspace{0.5em}  (p < 0.001)$); 
that is, as the complexity of abstractions increased, so did the workload, supporting H2.

\begin{figure}
    \centering
    \begin{subfigure}[c]{0.31\textwidth}
        \centering
        \includegraphics[trim={0cm 0cm 0cm 0cm},clip=true,width=\textwidth]{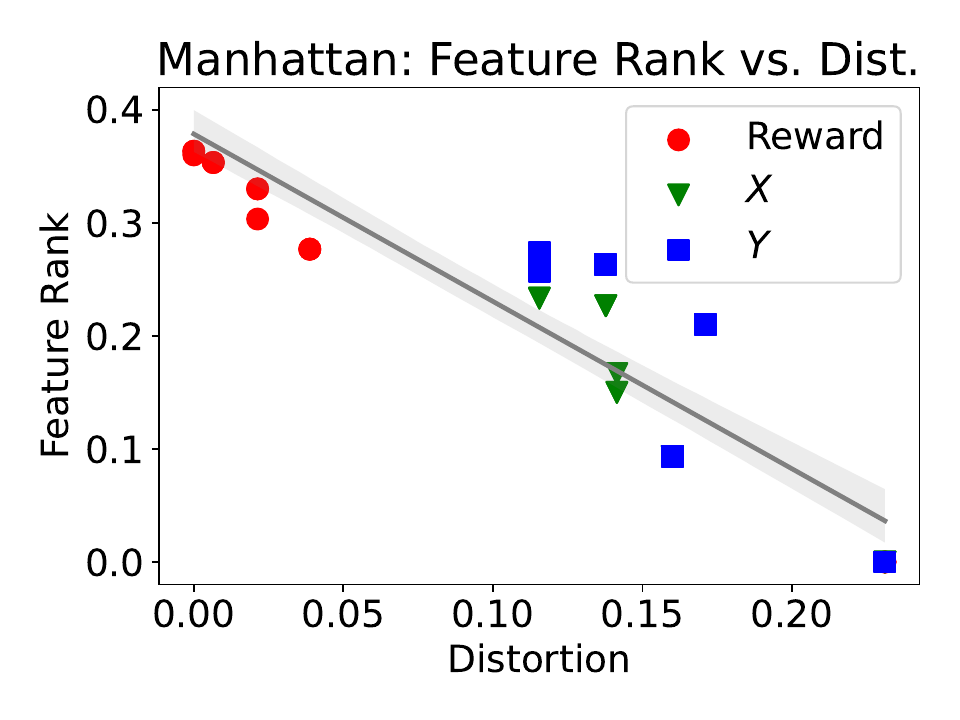}
        \caption{Feature Rank vs. Distortion}
    \end{subfigure}
    \begin{subfigure}[c]{0.31\textwidth}
        \centering
        \includegraphics[trim={0cm 0cm 0cm 0cm},clip=true,width=\textwidth]{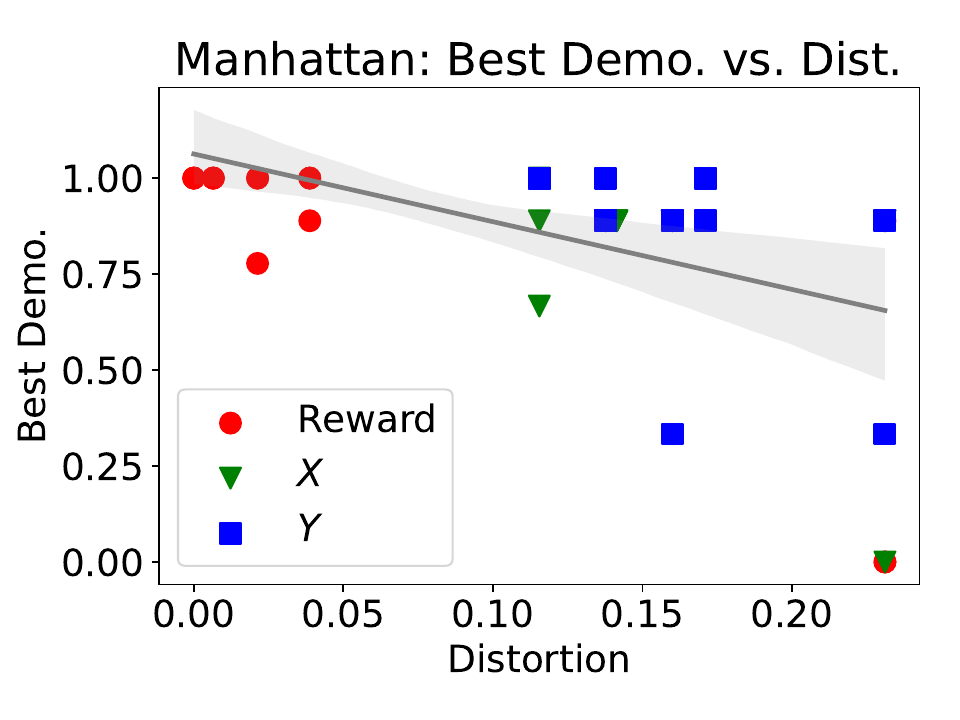}
        \caption{Best Demonstration vs. Distortion}
    \end{subfigure}
    \begin{subfigure}[c]{0.31\textwidth}
        \centering
        \includegraphics[trim={0cm 0cm 0cm 0cm},clip=true,width=\textwidth]{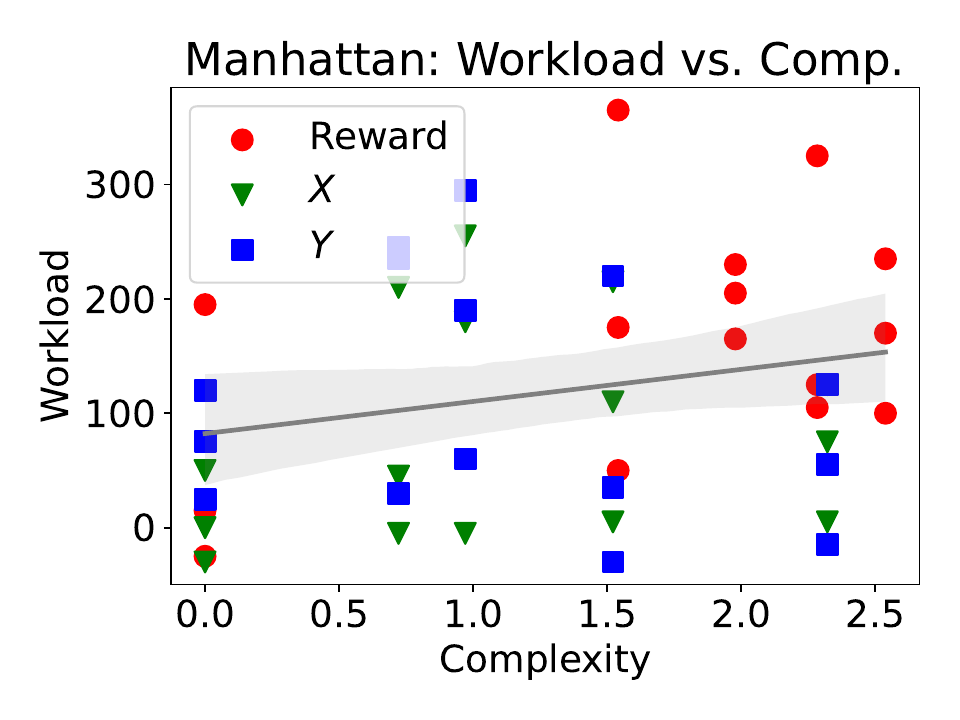}
        \caption{Workload vs. Complexity}
    \end{subfigure}
    \caption{Manhattan domain results, connecting metrics of understanding to distortion (a-b) and workload to complexity (c). Increased distortion led to worsened understanding, and increased complexity led to increased workload.}
    \label{fig:manhattan_results}
\end{figure}

We observed similar trends in the Manhattan grid domain, depicted in Figure~\ref{fig:manhattan_results}.
Both $FR$ and $BD$ scores decreased as distortion increased (Figures~\ref{fig:manhattan_results}~a and b). 
Concretely, $\rho($ FR, dist$) = -0.97 \hspace{0.5em} (p < 0.001)$, and $\rho($ BD, dist$) = -0.63 \hspace{0.5em} (p < 0.001)$.
That is, each measure of understanding was significantly negatively correlated with distortion and, as before, the correlation was stronger for FR than for BD.
Unlike with the Random grid, we did not identify a significant correlation between workload and complexity, although the trend was still positive: $\rho($ Work., Comp.$) = 0.23 \hspace{0.5em} p = 0.13$.

Following the analysis of Spearman correlations for each grid type separately, we performed linear mixed effects modeling (LMEM) on the joint data from the Manhattan and Random grids and found significant trends supporting all our hypotheses.
The models we applied for this analysis were formulated according to the following equations in Wilkinson notation \cite{wilkinson1973symbolic}: $FR \sim Distortion + (1 | Participant)$, $BD \sim Distortion + (1 | Participant)$, and $Workload \sim Complexity + (1 | Participant)$. Here, the models were fit using $Participant$ as a grouping variable, with a random intercept to account for the individual differences between participants, which were not accounted for by the Spearman correlations 
(e.g., perhaps one participant would consistently report higher workload). 
In our joint analysis, we leveraged the fact that each participant answered one question about the Manhattan grid and one about the Random grid.
We observed significant main effects for distortion and complexity within each model at the $p<0.001$ level, with the following effect sizes and intercepts for each:
$\text{$FR$} = -0.92 * \text{Dist.} + 0.35$, $\text{$BD$} = -2.19 * \text{Dist.} + 1.02$, $\text{Work.} = 33.5 * \text{Comp.} + 79.80$.

Overall, the results from our grid domain experiments strongly support our hypotheses that 1) decreased distortion would be associated with increased understanding and 2) increased complexity would be associated with increased workload. 
We found statistically significant support for all but one of our hypothesized results via Spearman correlation tests assessing each grid-based reward function separately, and we found support for all of our hypothesized results when evaluating the combined datasets through a linear mixed effects analysis which accounted for individual differences in participant responses.

\subsection{Color Domain}
We applied the same set of analyses as in the grid navigation domain to the color domain: again analyzing the results for the continuous and discontinuous reward functions separately via the Spearman correlation, then performing a joint analysis of the data together with a LMEM. Forty-seven participants completed the grid navigation survey (11 women, 33 men, 1 non-binary individual, and 2 not reporting gender). The median age was 33 years (min=20 years, max=66 years). We omitted data collected from two participants from analysis due to failed attention-check questions or incomplete responses.

\begin{figure}
    \centering
    \begin{subfigure}[c]{0.31\textwidth}
        \centering
        \includegraphics[trim={0cm 0cm 0cm 0cm},clip=true,width=\textwidth]{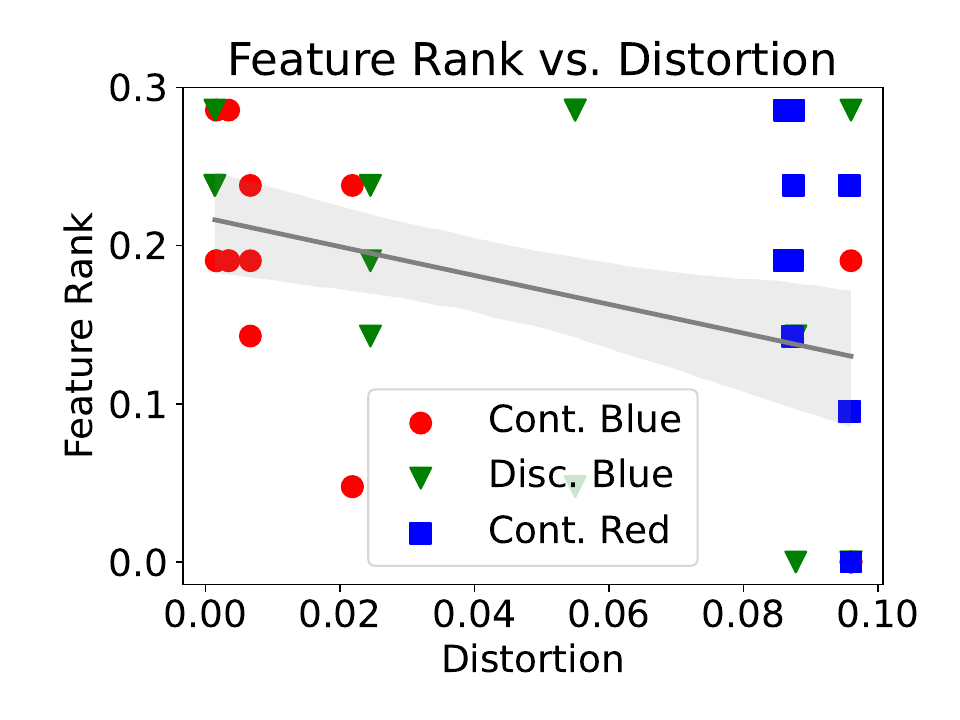}
        \caption{Feature Rank vs. Distortion}
    \end{subfigure}
    \begin{subfigure}[c]{0.31\textwidth}
        \centering
        \includegraphics[trim={0cm 0cm 0cm 0cm},clip=true,width=\textwidth]{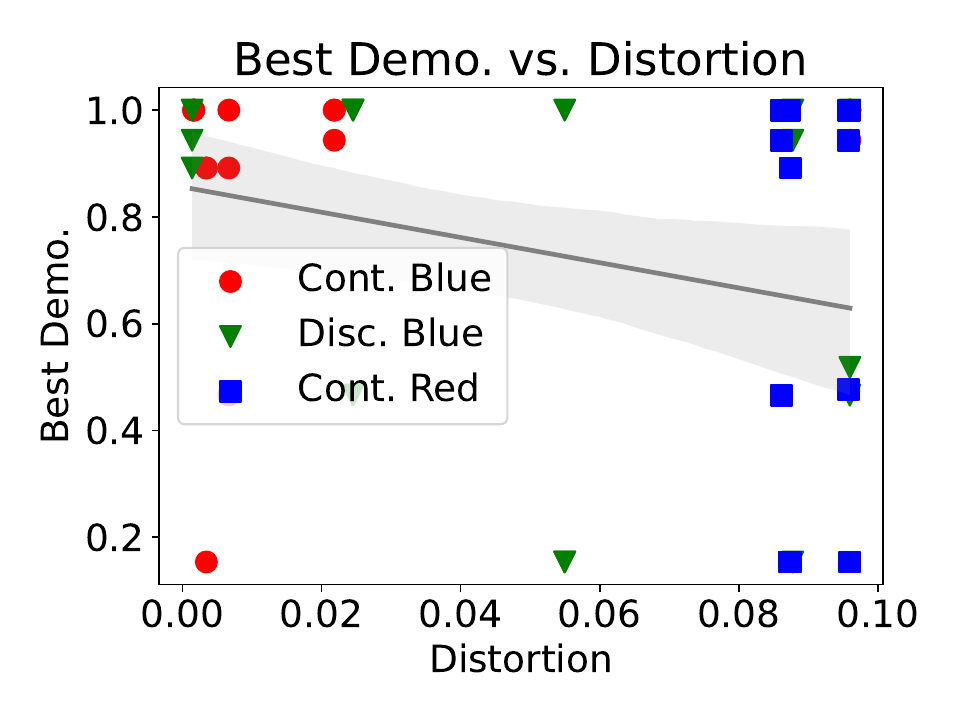}
        \caption{Best Demonstration vs. Distortion}
    \end{subfigure}
    \begin{subfigure}[c]{0.31\textwidth}
        \centering
        \includegraphics[trim={0cm 0cm 0cm 0cm},clip=true,width=\textwidth]{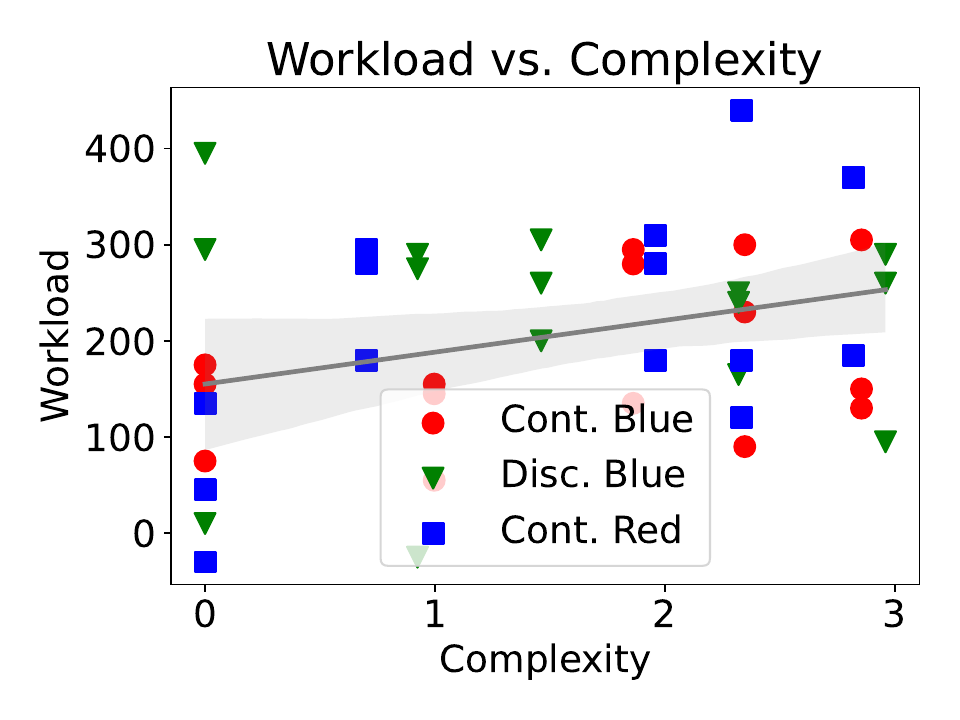}
        \caption{Workload vs. Complexity}
    \end{subfigure}
    \caption{Color domain results, using the continuous reward function. Trends were weaker than those observed in the grid domains, but still reflected the hypothesized directions. Similar results for the discontinuous reward function are included in Figure~\ref{fig:disco_results} in Appendix~\ref{app:color_full}.}
    \label{fig:continuous_results}
\end{figure}

Results from the color domain experiments corroborate the key trends observed in the grid navigation domains, with the exception of the \textit{best demonstration (BD)} understanding assessment. 
First, separating results by continuous and discontinuous reward functions, we established significant Spearman correlations for some of the hypothesized trends. 
\textit{Feature rank (FR)} and distortion were negatively correlated for both reward functions $(p < 0.001)$.
For the continuous reward (shown in Figure \ref{fig:continuous_results} a), $\rho(\text{FR}, \text{dist}) = -0.47$; for discontinuous, $\rho(\text{FR}, \text{dist}) = -0.40$.
Correlations between complexity and workload were positive, but not at the $p = 0.05$ level: for continuous (Figure \ref{fig:continuous_results} c), $\rho(\text{Work.}, \text{Comp.}) = 0.24 \hspace{0.5em} (p = 0.06)$ and for discontinuous, $\rho(\text{Work.}, \text{Comp.}) = 0.18 \hspace{0.5em} (p = 0.11)$.
Lastly, correlations between the \textit{best demonstration (BD)} understanding assessment and distortion were weak, with no significance value lower than 0.15.
We attribute the \textit{best demonstration (BD)} correlation failure to high random chance performance with high distortion: even with completely uninformative abstractions, some participants selected the optimal best path through random guessing. Also of note is that the visualizations of the abstractions within the color domain were not provided in the same space in which the best demonstration task was performed (as opposed to the grid domain, where the abstractions were visualized in the grid itself), so another possible reason for this difference (and the added difficulty with high-complexity abstractions) is the extra step necessary to translate the abstract information into the task space. 

Complementing our Spearman correlation analysis for each domain separately, we again performed LMEM tests on all the joint data, grouped by participant, as we did for our grid navigation experiments. We applied linear mixed effects models of the same form, and found significant main effects for both distortion in the $FR$ model and complexity in the $workload$ model. The linear trends for $FR$ vs. distortion and workload vs. complexity were significant at $p < 0.05$: $\text{FR} = -0.70 * \text{Dist} + 0.34$, $\text{Work.} = 39.43 * \text{Comp.} + 156.23$.
These findings support our two hypotheses that (H1) increasing distortion would decrease understanding and (H2) increasing complexity would increase workload. 
As before, however, the correlation between $BD$ and distortion was not statistically significant: $\text{BD} = -1.11 * \text{Dist.} + 0.73 \hspace{0.5em} (p = 0.12)$. 

\section{Discussion}

Across domains, we found evidence supporting our two hypotheses, with more mixed results in the color domain.  
In every domain, and with every reward function, we observed significantly negative Spearman correlation coefficients between feature-based understanding and distortion. 
While the significance of other trends varied slightly across domains, when we leveraged the within-participant aspect of our experiment design to account for individual differences in participant responses, we similarly found significantly positive correlations between workload and complexity. 
Although weaker than the $FR$ and distortion correlations, we also found significant correlations between the $BD$ measure of understanding (i.e., policy understanding) and distortion in all cases within the grid navigation domain. These weaker correlations track with previous experimental results, which demonstrated that the factor loadings for policy-based assessments of alignment (understanding) were weaker than those for feature-based assessments \cite{sanneman2023validating}; this is likely due to the additional challenge of translating a reward function into an optimal policy within a given environment. Nonetheless, the overall trends in our results support the hypothesized link between human factors constructs and information-theoretic concepts. This enables us to leverage these information-theoretic concepts to mathematically characterize the workload-understanding tradeoff in XAI design and to automatically generate abstraction-based explanations which trade off these factors, allowing us to account for variable informational and workload needs between different users of AI systems.

Overall, we observed a larger number of significant results and stronger correlations in the grid-navigation domain compared with the color domain, particularly for the $BD$ (policy understanding) results. This is likely related to the visualizations of the abstractions in each domain: in the grid-navigation domain, heat maps of square values were provided within the best demonstration task grid itself, while in the color domain, participants interpreted the heat maps and their relation the color grid separately from the task grid, and then had to translate their reward understanding into an optimal policy in the task grid in an additional step. While we identified significant support for our key hypotheses related to the relationships between information-theoretic concepts and human factors constructs in abstract explanation generation across both experiments, the differences in the $BD$ results corresponding to the different types of abstraction visualizations between the experiments highlight the importance of carefully considering how to visualize abstractions for effective communication in future explanation design.

Finally, we observed some evidence that continuous reward functions may be better candidates for abstraction than discontinuous ones. 
The correlations between $FR$ (feature-based reward understanding) and distortion were stronger for the Manhattan grid (with a fundamentally continuous reward structure) than the Random grid in the grid-navigation domain, and for the continuously varying reward function than the discontinuously varying reward function in the color domain. 
This suggests that abstracting such continuously varying reward regions may lead to more natural explanations of reward functions than groupings of discontinuous reward regions. We leave additional exploration of the impact of the structure of reward functions and the inherent ``abstractability" of their feature spaces on explanation efficacy as an area for future work.

\subsection{Limitations and Future Work}
Our work takes an important step toward connecting IB methods to human factors in understanding explanations, but we rely upon several simplifying assumptions. 
First, in our experiments, we used the ``ground truth'' reward function, as well as alternative reward function baselines, to generate abstractions. 
As shown in our results, generating low-distortion explanations is extremely important for understanding explanations, and lacking the ground truth reward function would necessarily lead to higher-distortion abstractions. 
For the purposes of this paper, we therefore scoped our work to assume access to such functions; future work may wish to relax this assumption, but must therefore propose methods for the use of alternative reward functions.

Second, in our experiments, we used exact IB methods for generating abstractions, which may not scale to larger domains. 
For example, as the number of grid locations in the navigation domain increases, standard IB methods will slow down considerably. 
Fortunately, recent complementary work proposed approximated discrete IB methods for rapidly generating abstractions at varying complexity levels~\citep{tucker2022trading}; we anticipate that such methods may be easily combined with our explanation work to incorporate approximate IB abstractions into explanations.

Finally, we scoped our work to study abstract explanations of reward functions in particular, but there are additional concepts related to AI decision making, such as policies, constraints, counterfactual decisions, and decision uncertainty (among others), which might also be effectively explained through the application of automatically generated abstract explanations. Beyond this, such techniques can be extended to explain complex concepts in larger-scale domains, such as reinforcement learning-based robotics applications and large language models (LLMs). In this work, we have established empirical connections between information-theoretic concepts and human factors constructs which we hope will apply to explanation design for this broader scope of AI concepts and domains, and have laid the groundwork for future exploration and confirmation of these relationships in different settings. Future work can explore these relationships and their applicability to human-centered XAI design in an expanded assortment of settings.

\section{Conclusion}
In this work, we established empirical connections between human factors metrics of explanation understanding and workload with Information Bottleneck (IB) concepts of distortion and complexity.
In the standard IB framework, a tradeoff exists between distortion and complexity; we established a similar tradeoff of people improving reward understanding as distortion decreased, but at at the cost of increased workload as complexity increased.
Our findings may be used directly to inform explanation design, especially in accounting for differing informational and workload needs between individual users of AI systems.
For example, given a maximum acceptable workload, one could find the corresponding allowed complexity level for explanations, and, at that complexity level, promote understanding by minimizing distortion.
More generally, our work establishes important connections at the intersection of human factors and information theory, which we hope future work will continue to explore.

\begin{acks}
To Robert, for the bagels and explaining CMYK and color spaces.
\end{acks}

\bibliographystyle{ACM-Reference-Format}
\bibliography{sample-base}

\appendix
\section{Generating IB Abstractions Implementation}
\label{app:ib_impl}
We used the \texttt{embo} package to generate abstractions at different levels of complexity and reward distortion and employed these abstractions in human participant experiments to analyze the role of complexity and distortion in human understanding~\citep{embo}.
Here, we discuss how we generated our abstractions, and demonstrate that we induced meaningful variation in complexity and distortion.\footnote{Code to generate the abstractions in our experiments is available at \href{https://github.com/mycal-tucker/ib-explanations}{https://github.com/mycal-tucker/ib-explanations}.}

\subsection{Grid Domain}
In the grid domains, we numbered each of the 25 grid cells with a unique id, and used four reward functions during the IB process: Manhattan distance, random reward, $x$ coordinate, and $y$ coordinate.
For the Manhattan distance reward, depicted in the main paper in Figure~\ref{fig:domains}~a, reward was set to $+1$ at cell location $(1, 3)$ and decreased by $0.33$ for each increase in Manhattan distance, to a minimum of $-1.0$.
For the random reward, depicted in Figure~\ref{fig:domains}~b, we selected a reward value uniformly at random in the range $[-1, 1]$.
For exact values, we refer the reader to our code which, given the fixed random seed of 0, will exactly reproduce the random values in our experiments.
Lastly, for the $x$ and $y$ coordinate rewards, we set the reward equal to the integer value of the $x$ or $y$ coordinate of each cell, ranging from 0 to 4, inclusive.

For each of the four training objects used during the IB process, we evaluated abstractions with the Manhattan and random grid reward functions.
For example, we generated abstractions via the $x$ coordinate reward function, which divided the grid into vertical regions, and evaluated distortion using such abstractions for the Manhattan and random grid rewards.
Figure~\ref{fig:grid_full}~a in the main paper shows how abstractions generated using different reward functions resulted in different distortion values for the same complexity.

\subsection{Color Domain}
\label{app:color}
In the color domain, we uniquely numbered each of the 122 colors in our experiments, and used three reward functions during the IB process: 1) predict the blue value in the color's RGB representation, 2) predict the discontinuous reward (defined in the next paragraph) based on the blue value of the color, or 3) predict the red value of the color's RGB representation.

The continuous reward functions (predicting the blue or red value) were simply defined as the continuous value of the color, ranging from 0 to 1.0.
We additionally used a discontinuous reward function, which divided the blue color range into eight bins of equal sizes (in $[0, 0.125)$, $[0.125, 0.25)$, etc.) with values of $[0.5, -0.5, 0.0, 0.75, 1.0, -1.0, 0.25, -0.75]$.
We chose these values to intentionally cause similar continuous blue values (e.g., $0.124$ and $0.126$) to have dissimilar rewards (e.g., $0.5$ and $-0.5$).

\begin{figure}
    \centering
    \includegraphics[trim={0cm 0.5cm 0.5cm 2cm},clip=true, width=\textwidth]{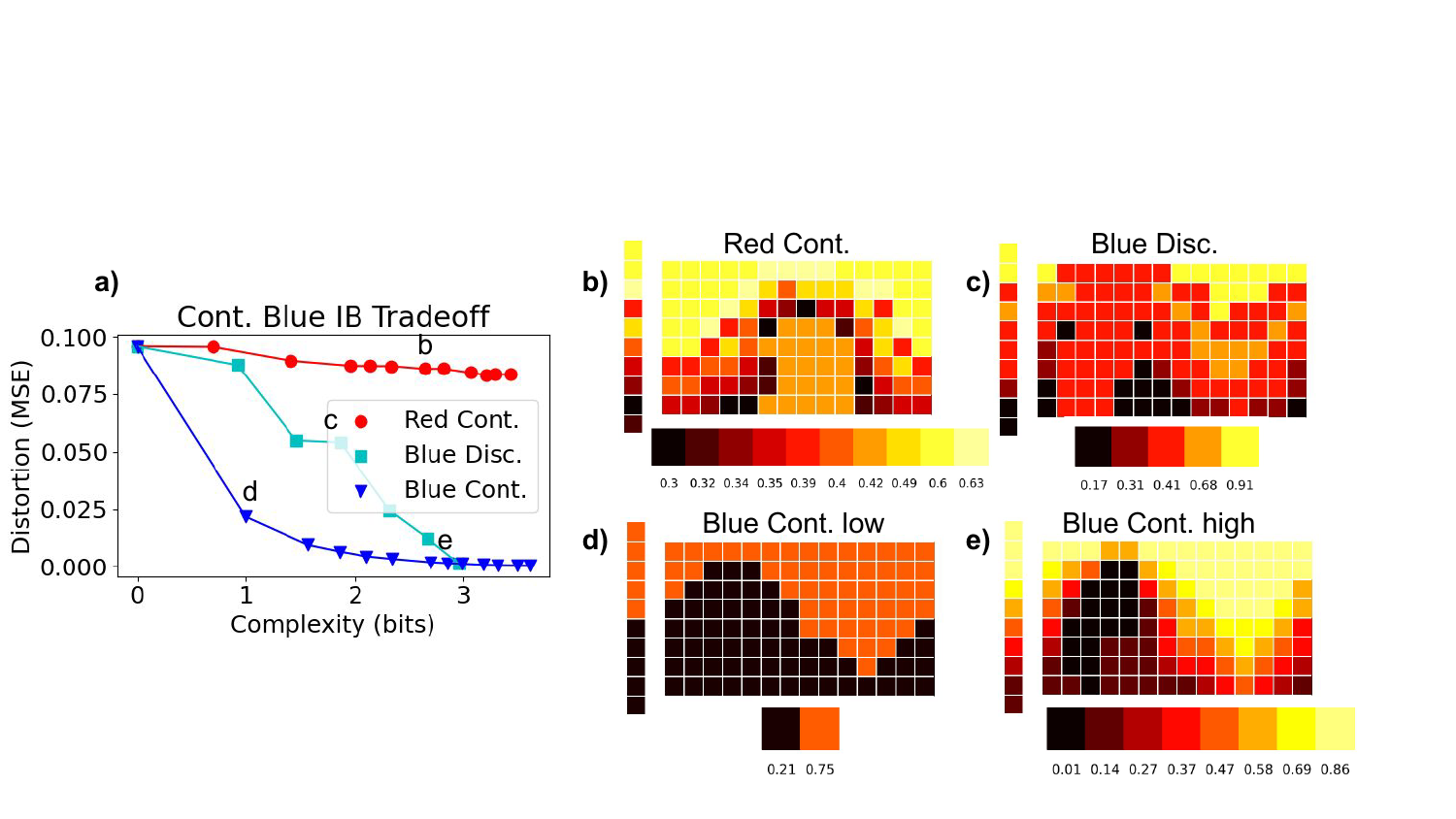}
    \caption{Complexity-distortion curves (\textbf{a)}) and corresponding abstractions (\textbf{b-e}) for the continuous reward function in the color domain. Using different rewards to generate abstractions (different curves) led to varying distortions for the same complexity. Using the continuous reward function led to optimal distortion-complexity tradeoffs, and varying complexity increased the number of abstractions (\textbf{d-e}).}
    \label{fig:app_color_full}
\end{figure}

We include results of evaluating all such abstractions using the continuous blue reward function in Figure~\ref{fig:app_color_full}.
Unsurprisingly, using the continuous blue reward to generate abstractions resulted in lower distortion for the same complexity compared with using the other two training objectives.
Concretely, we observed that increasing the complexity of abstractions used for predicting a color's red value had virtually no effect on the distortion for predicting the color's blue value.
Given the orthogonal nature of blue and red representations in RGB colors, this is unsurprising.
Overall, we used these three training objects for generating abstractions to decouple changes in complexity and distortion, while exploring meaningful ranges for each value.

\section{Color Domain Results}
\label{app:color_full}

In the main paper, we omitted some of the graphs from the color domain experiments for clarity; we include them here for completeness.

\begin{figure}
    \centering
    \begin{subfigure}[c]{0.31\textwidth}
        \centering
        \includegraphics[trim={0cm 0cm 0cm 0cm},clip=true,width=\textwidth]{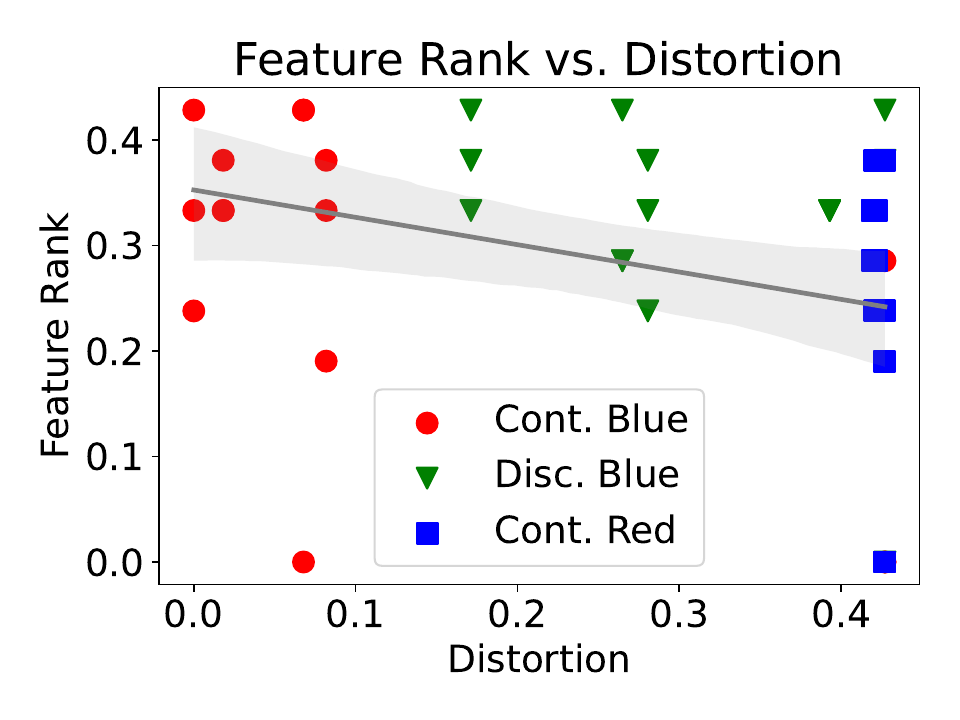}
        \caption{Feature Rank vs. Distortion}
    \end{subfigure}
    \begin{subfigure}[c]{0.31\textwidth}
        \centering
        \includegraphics[trim={0cm 0cm 0cm 0cm},clip=true,width=\textwidth]{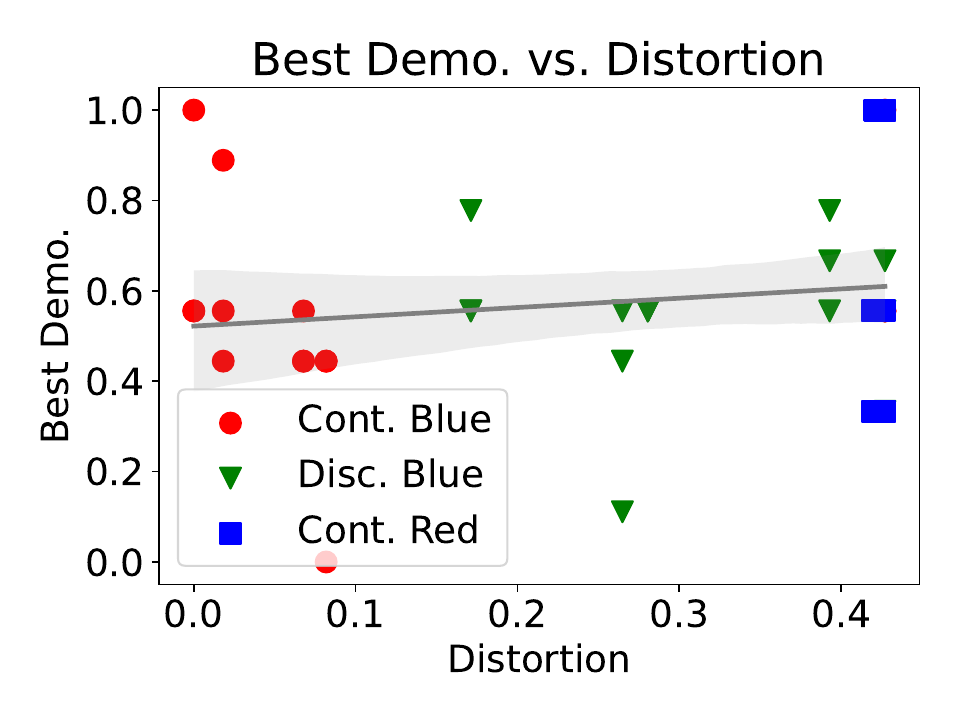}        \caption{Best Demonstration vs. Distortion}
    \end{subfigure}
    \begin{subfigure}[c]{0.31\textwidth}
        \centering
        \includegraphics[trim={0cm 0cm 0cm 0cm},clip=true,width=\textwidth]{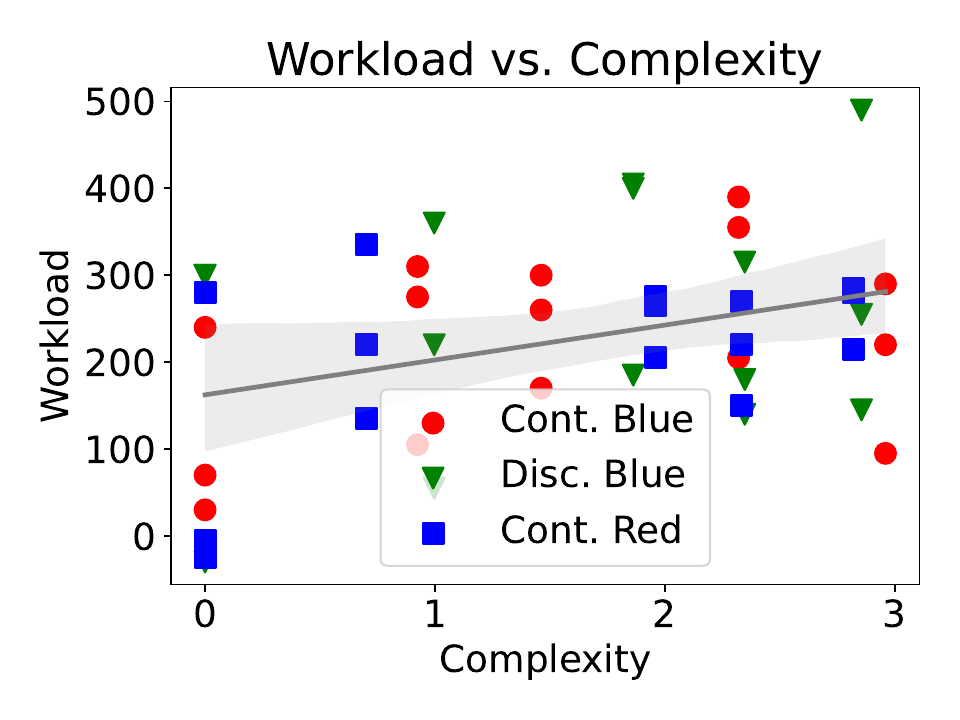}
        \caption{Workload vs. Complexity}
    \end{subfigure}
    \caption{Color domain results using the discontinuous blue reward function. Trends were weaker than those observed with the continuous reward, although there was still a significant negative correlation between feature rank and distortion (a).}
    \label{fig:disco_results}
\end{figure}

Figure~\ref{fig:disco_results} shows the key trends between Feature Rank and Distortion (a), Best Demonstration and Distortion (b), and Workload and Complexity (c) using the discontinuous reward function.
In the main paper, we noted that the FR-distortion trend was significant but the Workload-Complexity trend was not, while the Best Demonstration results were never significant in any color domain.
The plots in Figure~\ref{fig:disco_results} visually corroborate these statistical tests.
The Best Demonstration trend line is nearly flat, including some participants predicting the exact right demonstration at a distortion of over 0.4, which indicates high performance using uninformative abstractions.
At the same time, the trend between complexity and workload is positive, as we hypothesized.

\section{Abstraction Visualizations}
\label{app:abs_viz}

Throughout our experiments, we used abstractions at different complexity and distortion levels and typically reported the complexity and distortion metrics.
Here, to provide intuition about the types of abstractions used, we provide visualizations for both domains, at various complexity and distortion values.

\begin{figure}
    \centering
    \begin{subfigure}[t]{0.31\textwidth}
        \centering
        \includegraphics[trim={1.1cm 0.3cm 1.1cm 0.5cm},clip=true,width=0.8\textwidth]{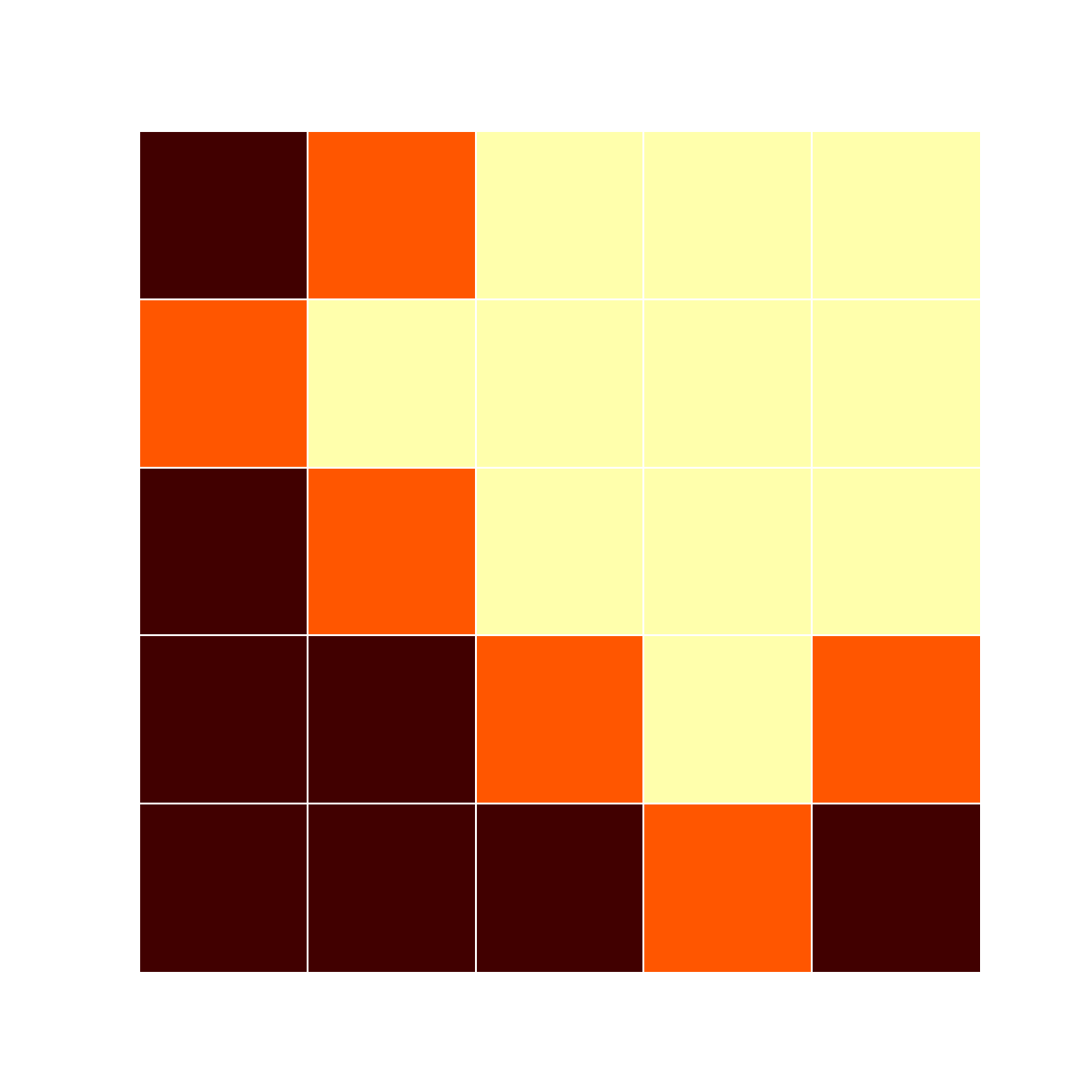}
    \end{subfigure}
    \begin{subfigure}[t]{0.31\textwidth}
        \centering
        \includegraphics[trim={1.1cm 0.3cm 1.1cm 0.5cm},clip=true,width=0.8\textwidth]{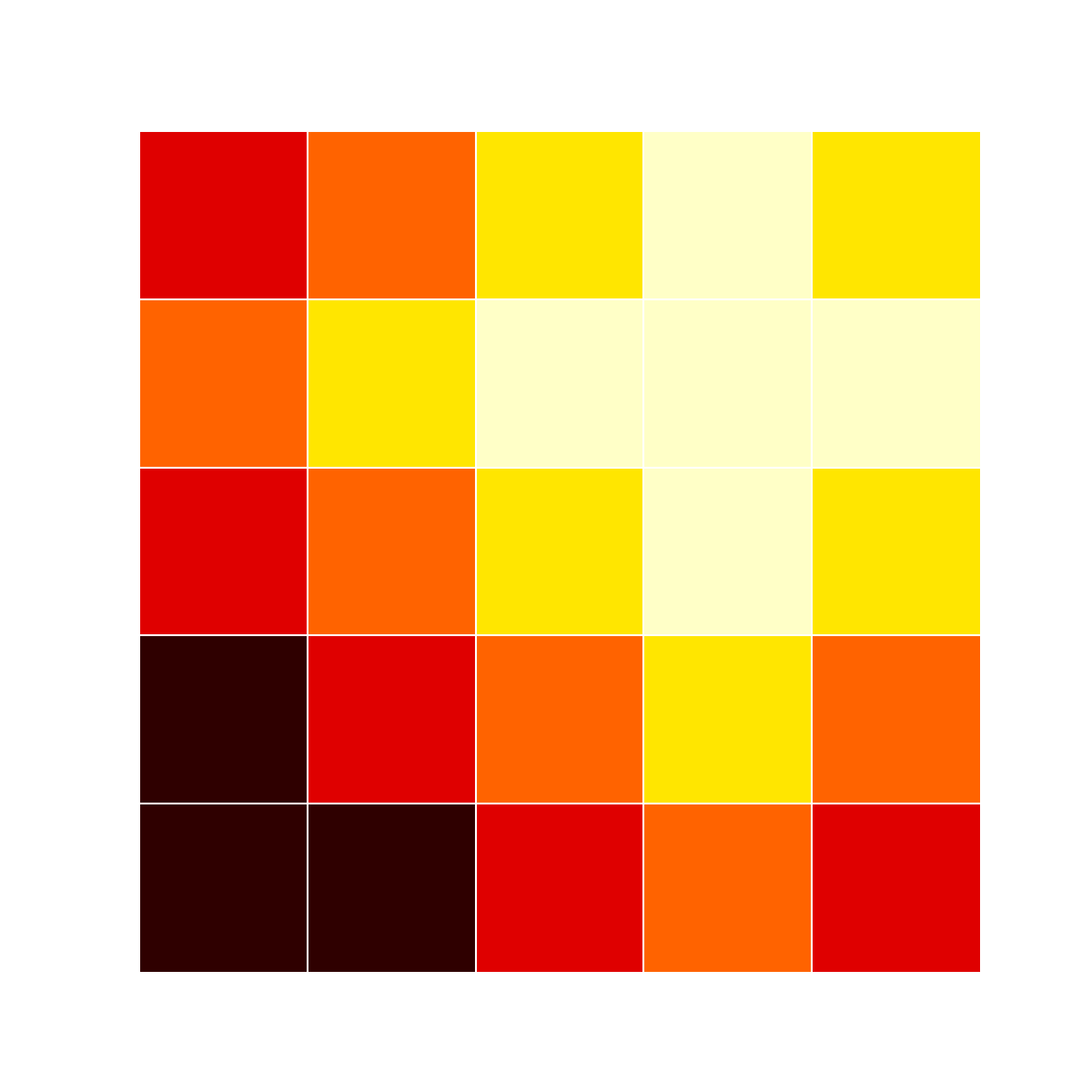}
    \end{subfigure}
    \begin{subfigure}[t]{0.31\textwidth}
        \centering
        \includegraphics[trim={1.1cm 0.3cm 1.1cm 0.5cm},clip=true,width=0.8\textwidth]{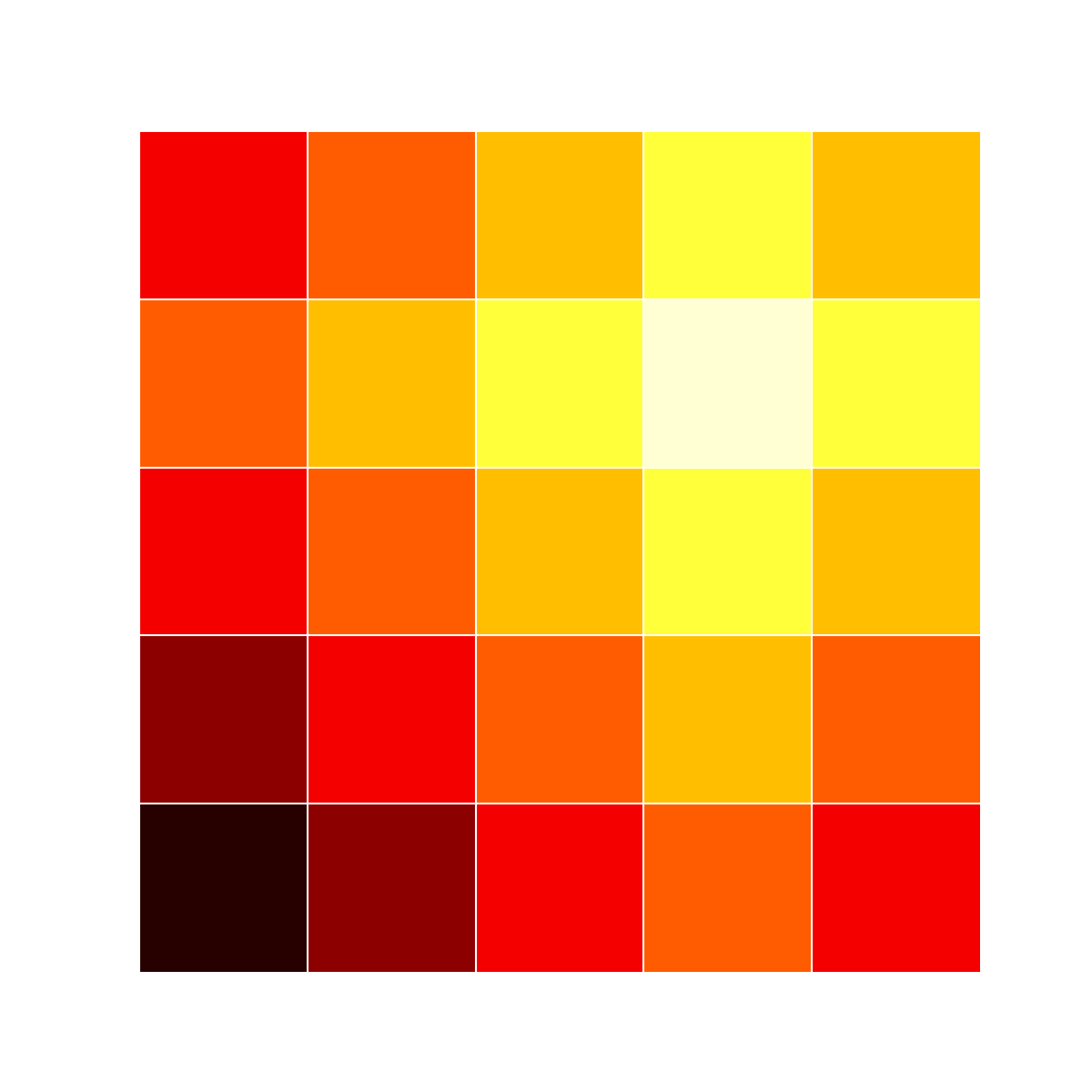}
    \end{subfigure}

    \begin{subfigure}[t]{0.31\textwidth}
        \centering
        \includegraphics[trim={0.5cm 0cm 0.2cm 0cm},clip=true,height=1.1cm]{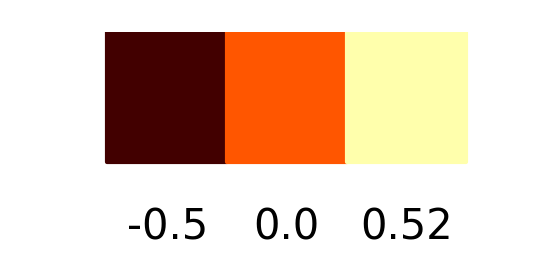}
        \caption{Manhattan Low}
    \end{subfigure}
    \begin{subfigure}[t]{0.31\textwidth}
        \centering
        \includegraphics[trim={0.5cm 0cm 0.2cm 0cm},clip=true,height=1.1cm]{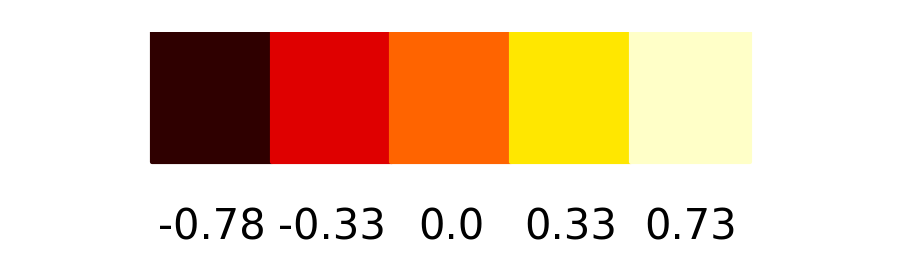}
        \caption{Manhattan Middle}
    \end{subfigure}
    \begin{subfigure}[t]{0.31\textwidth}
        \centering
        \includegraphics[trim={1.3cm 0cm 1.6cm 0cm},clip=true,height=1.1cm]{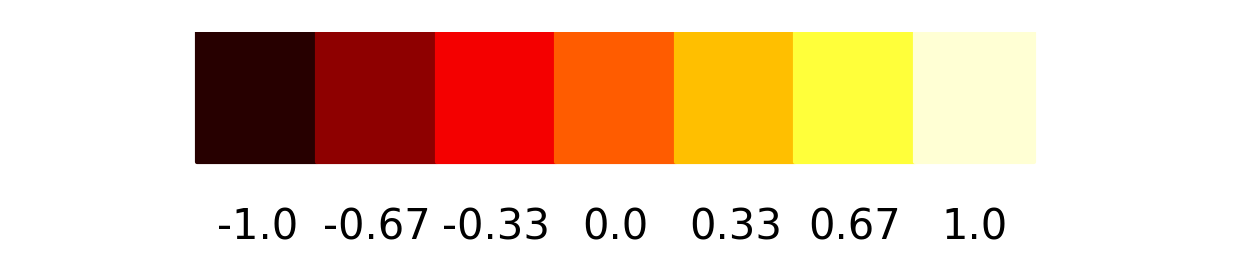}
        \caption{Manhattan High}
    \end{subfigure}

    \begin{subfigure}[t]{0.31\textwidth}
        \centering
        \includegraphics[trim={1.1cm 0.3cm 1.1cm 0.5cm},clip=true,width=0.8\textwidth]{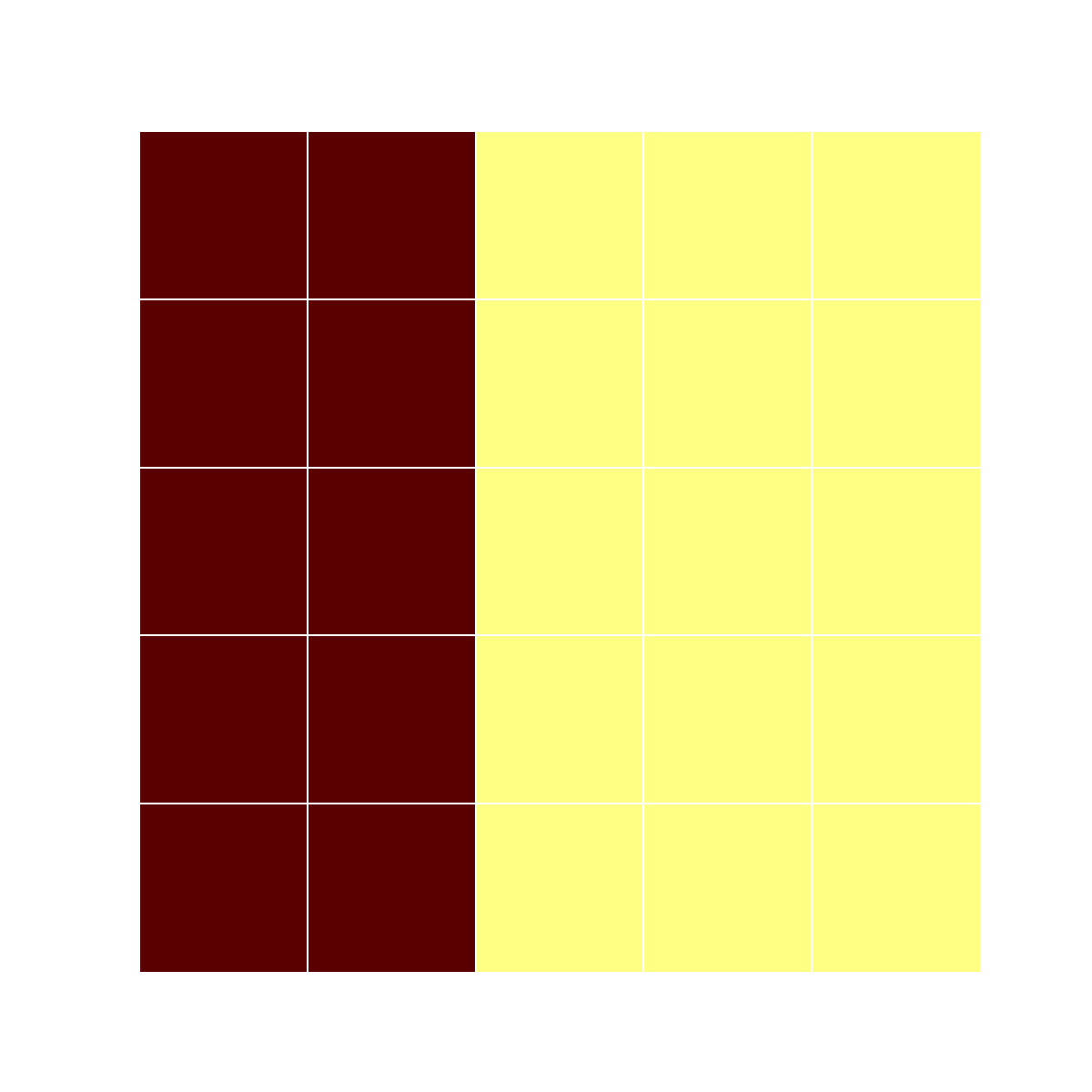}
    \end{subfigure}
    \begin{subfigure}[t]{0.31\textwidth}
        \centering
        \includegraphics[trim={1.1cm 0.3cm 1.1cm 0.5cm},clip=true,width=0.8\textwidth]{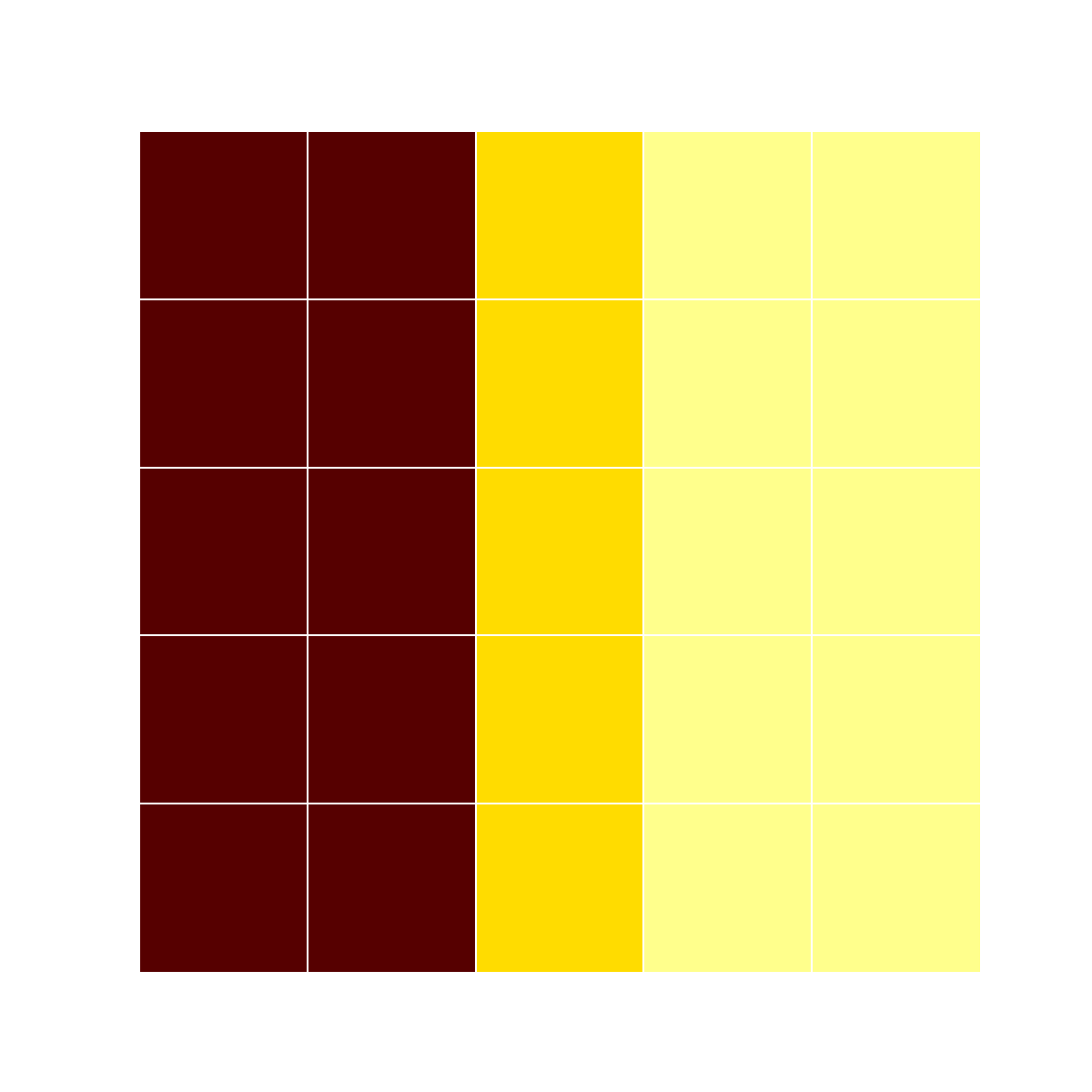}
    \end{subfigure}
    \begin{subfigure}[t]{0.31\textwidth}
        \centering
        \includegraphics[trim={1.1cm 0.3cm 1.1cm 0.5cm},clip=true,width=0.8\textwidth]{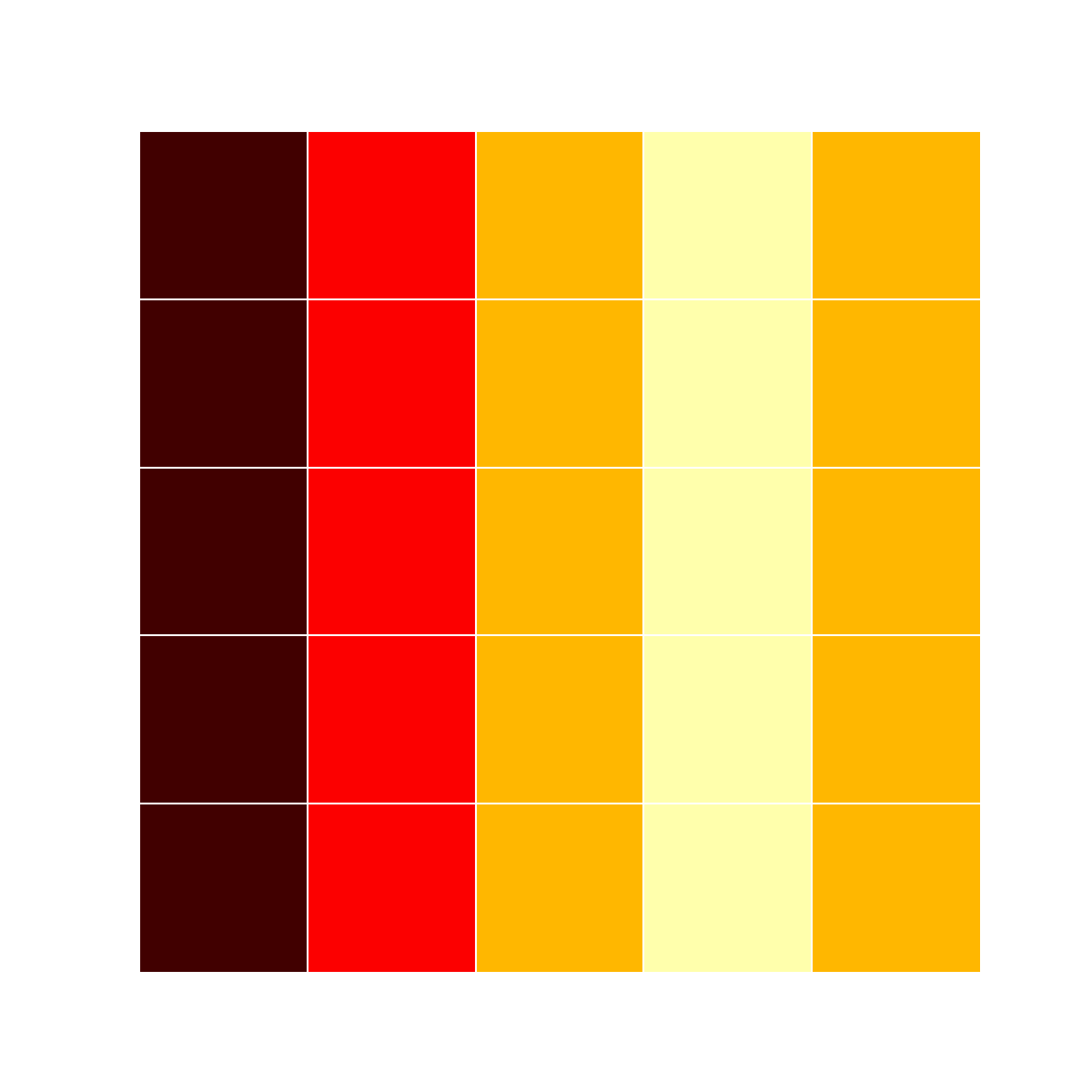}
    \end{subfigure}

    \begin{subfigure}[t]{0.31\textwidth}
        \centering
        \includegraphics[trim={0.5cm 0cm 0.2cm 0cm},clip=true,height=1.1cm]{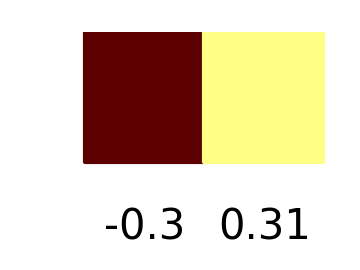}
        \caption{$X$ Coord. Low}
    \end{subfigure}
    \begin{subfigure}[t]{0.31\textwidth}
        \centering
        \includegraphics[trim={0.5cm 0cm 0.2cm 0cm},clip=true,height=1.1cm]{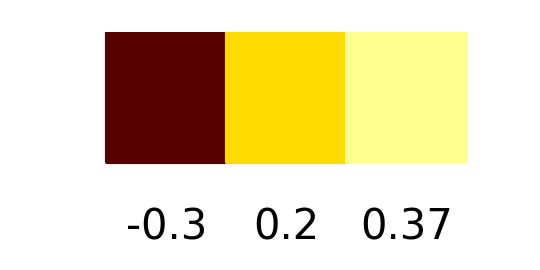}
        \caption{$X$ Coord. Middle}
    \end{subfigure}
    \begin{subfigure}[t]{0.31\textwidth}
        \centering
        \includegraphics[trim={0.7cm 0cm 0.9cm 0cm},clip=true,height=1.1cm]{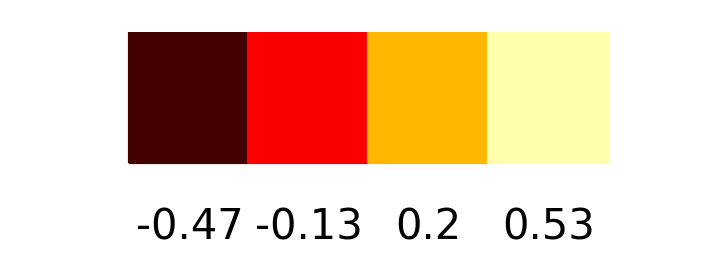}
        \caption{$X$ Coord. High}
    \end{subfigure}
    
    \begin{subfigure}[t]{0.31\textwidth}
        \centering
        \includegraphics[trim={1.1cm 0.3cm 1.1cm 0.5cm},clip=true,width=0.8\textwidth]{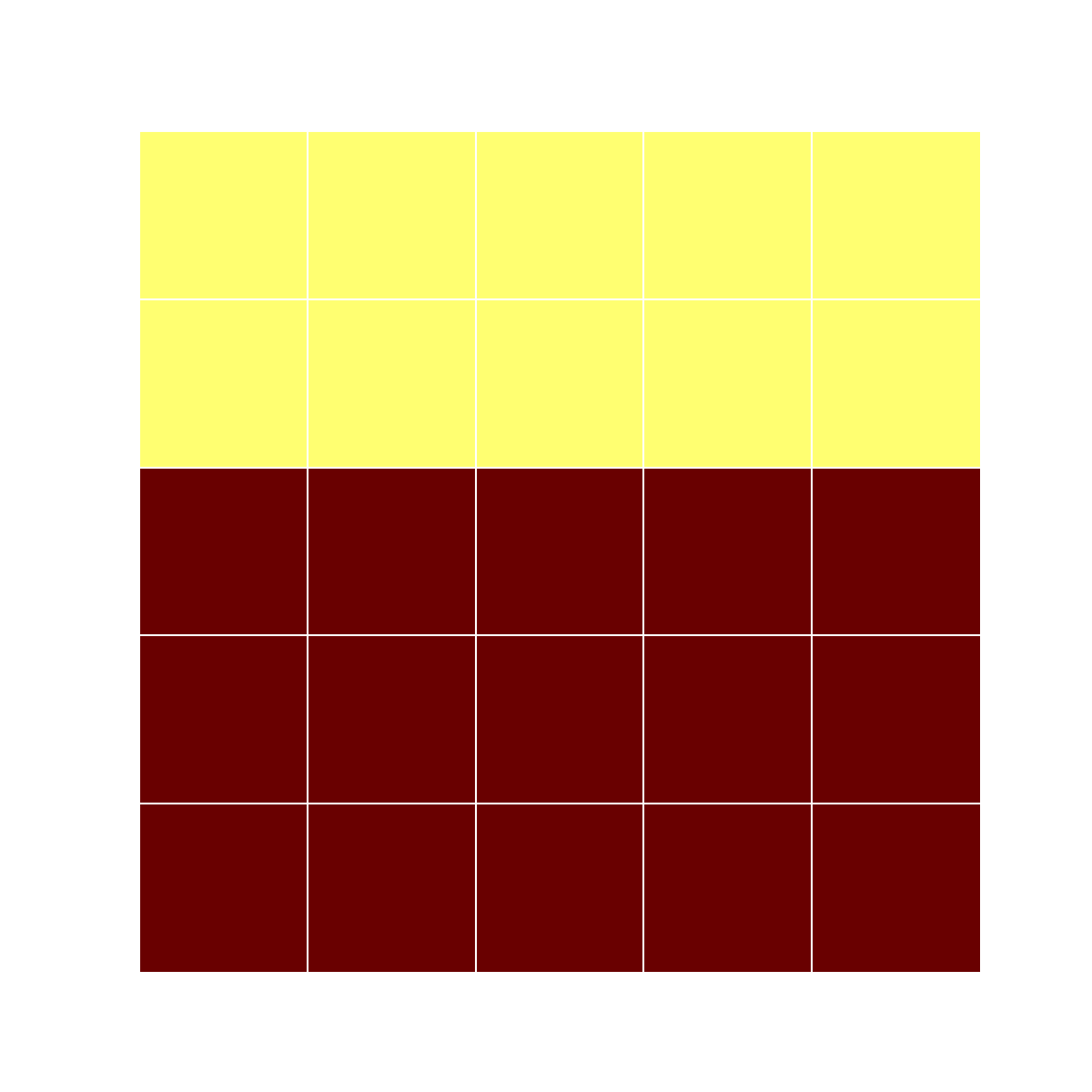}
    \end{subfigure}
    \begin{subfigure}[t]{0.31\textwidth}
        \centering
        \includegraphics[trim={1.1cm 0.3cm 1.1cm 0.5cm},clip=true,width=0.8\textwidth]{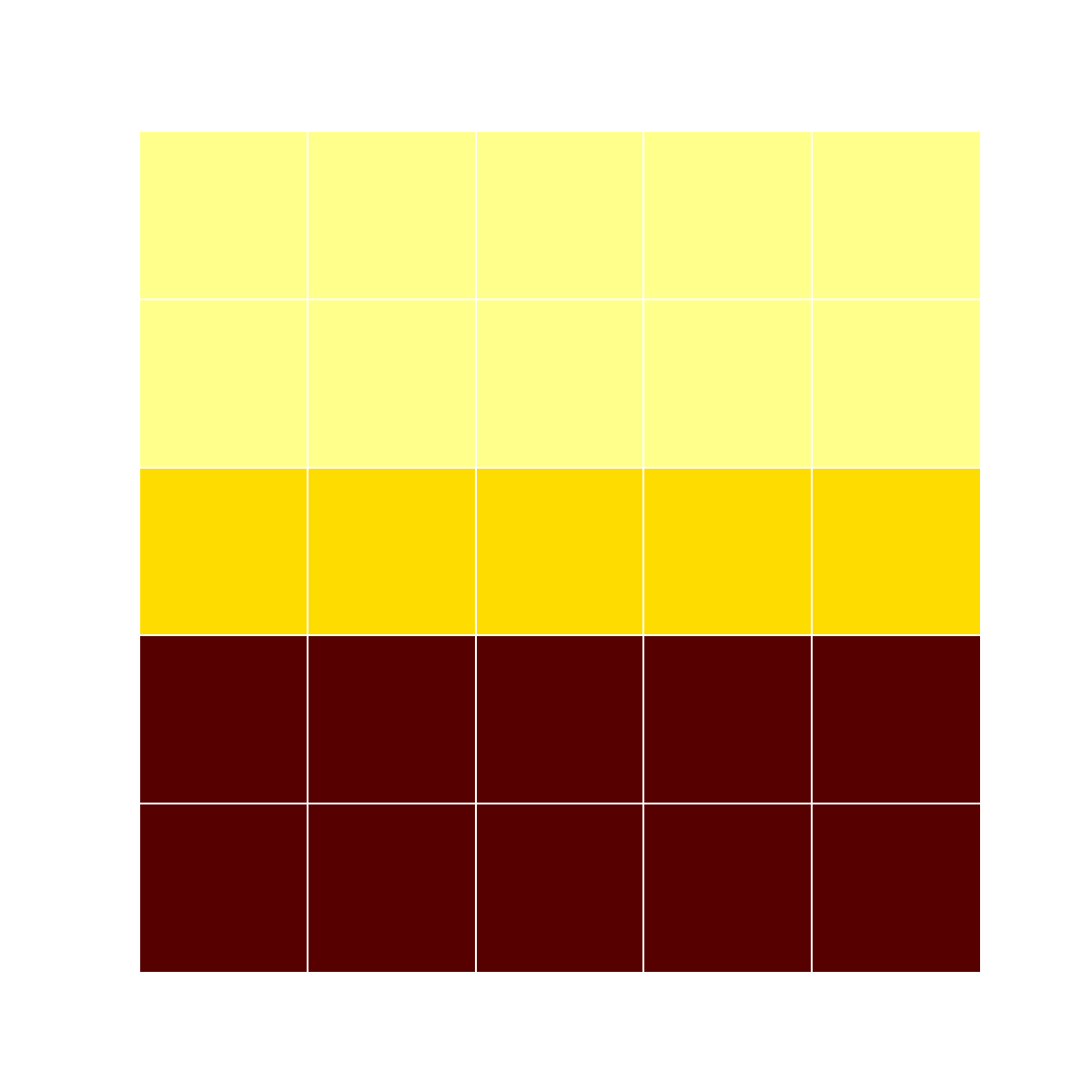}
    \end{subfigure}
    \begin{subfigure}[t]{0.31\textwidth}
        \centering
        \includegraphics[trim={1.1cm 0.3cm 1.1cm 0.5cm},clip=true,width=0.8\textwidth]{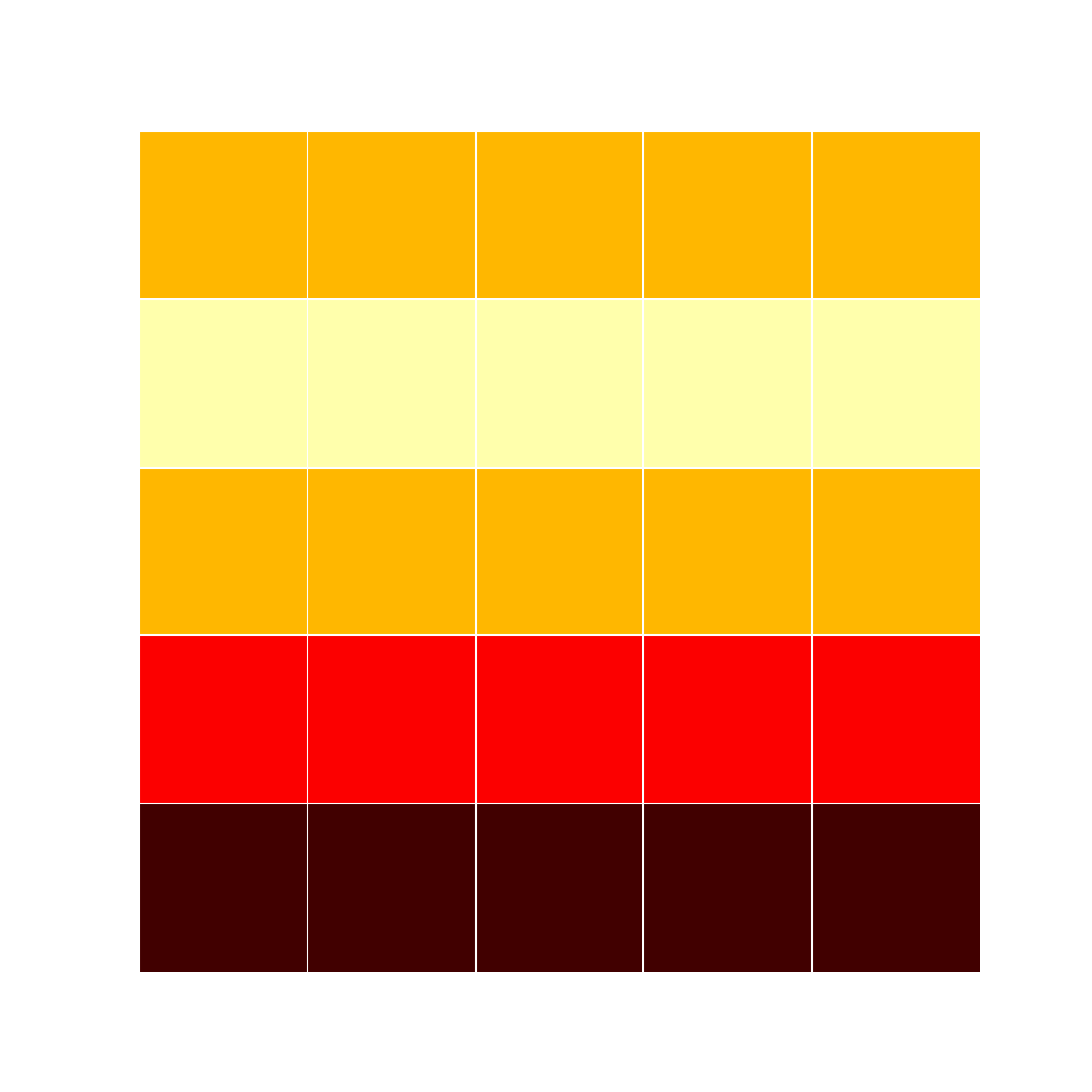}
    \end{subfigure}

    \begin{subfigure}[t]{0.31\textwidth}
        \centering
        \includegraphics[trim={0.5cm 0cm 0.2cm 0cm},clip=true,height=1.1cm]{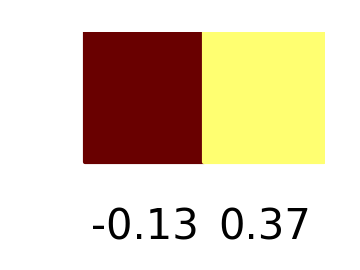}
        \caption{$Y$ Coord. Low}
    \end{subfigure}
    \begin{subfigure}[t]{0.31\textwidth}
        \centering
        \includegraphics[trim={0.5cm 0cm 0.2cm 0cm},clip=true,height=1.1cm]{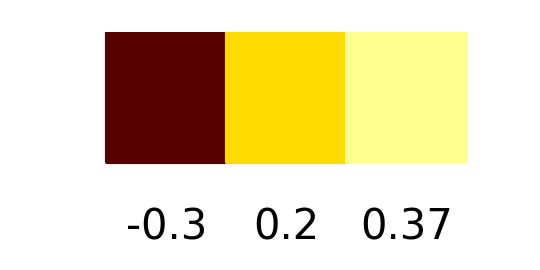}
        \caption{$Y$ Coord.Middle}
    \end{subfigure}
    \begin{subfigure}[t]{0.31\textwidth}
        \centering
        \includegraphics[trim={0.7cm 0cm 0.9cm 0cm},clip=true,height=1.1cm]{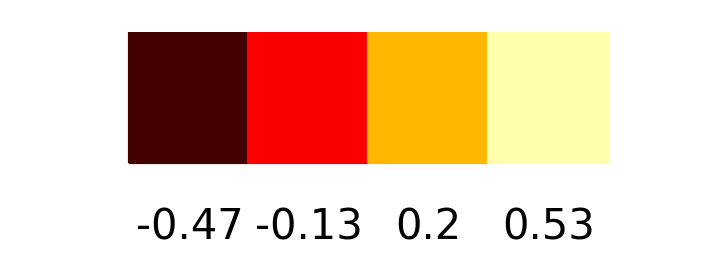}
        \caption{$Y$ Coord. High}
    \end{subfigure}
    \caption{Manhattan grid abstractions for various training objectives (different rows) and complexity levels (different columns). Increasing complexity led to more and finer-grained abstractions. Some abstractions led to low distortion (top row), whereas others removed important information, leading to high distortion (bottom two rows).}
    \label{fig:app_manhattan_grid_viz}
\end{figure}

\begin{figure}
    \centering
    \begin{subfigure}[t]{0.31\textwidth}
        \centering
        \includegraphics[trim={1.1cm 0.3cm 1.1cm 0.5cm},clip=true,width=0.8\textwidth]{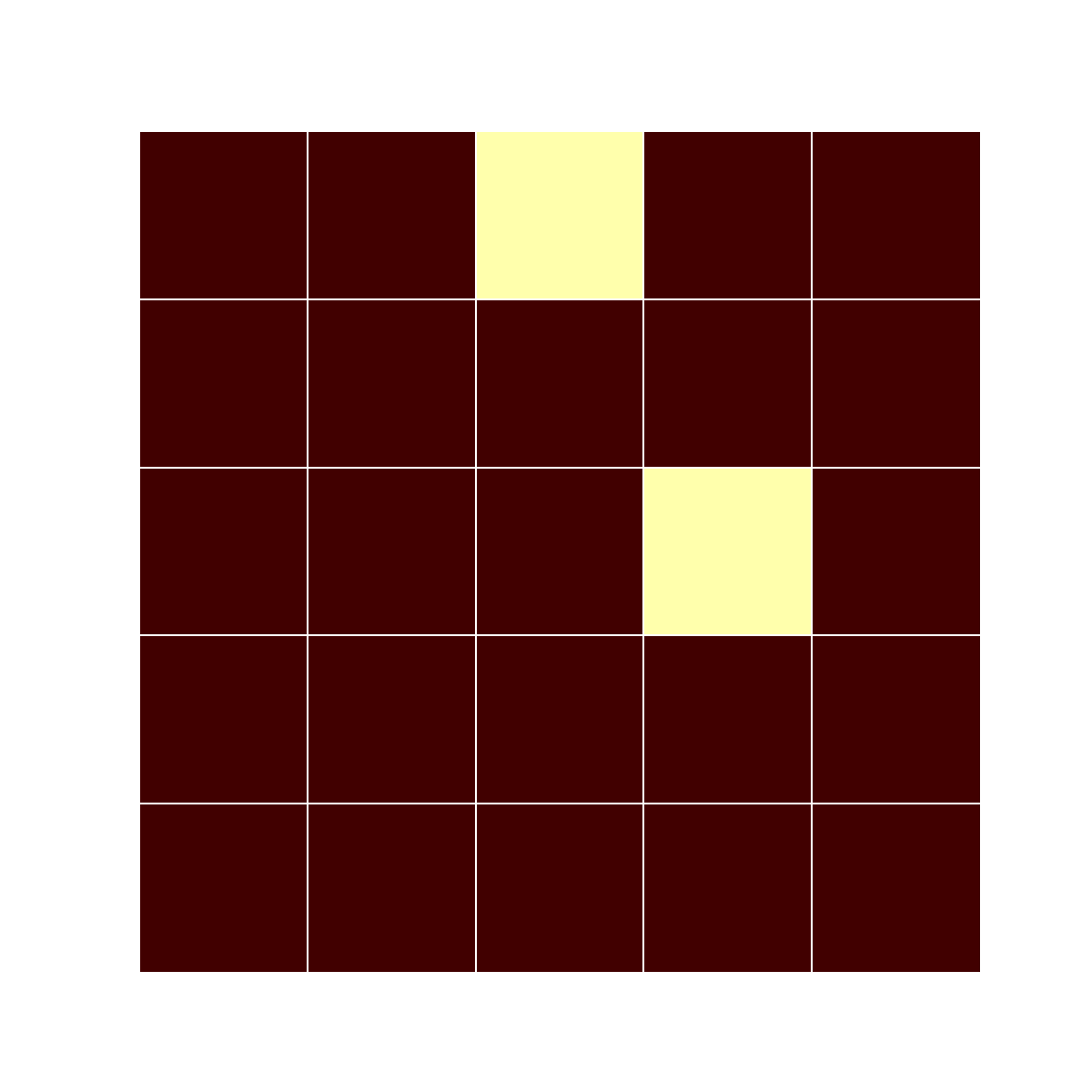}
    \end{subfigure}
    \begin{subfigure}[t]{0.31\textwidth}
        \centering
        \includegraphics[trim={1.1cm 0.3cm 1.1cm 0.5cm},clip=true,width=0.8\textwidth]{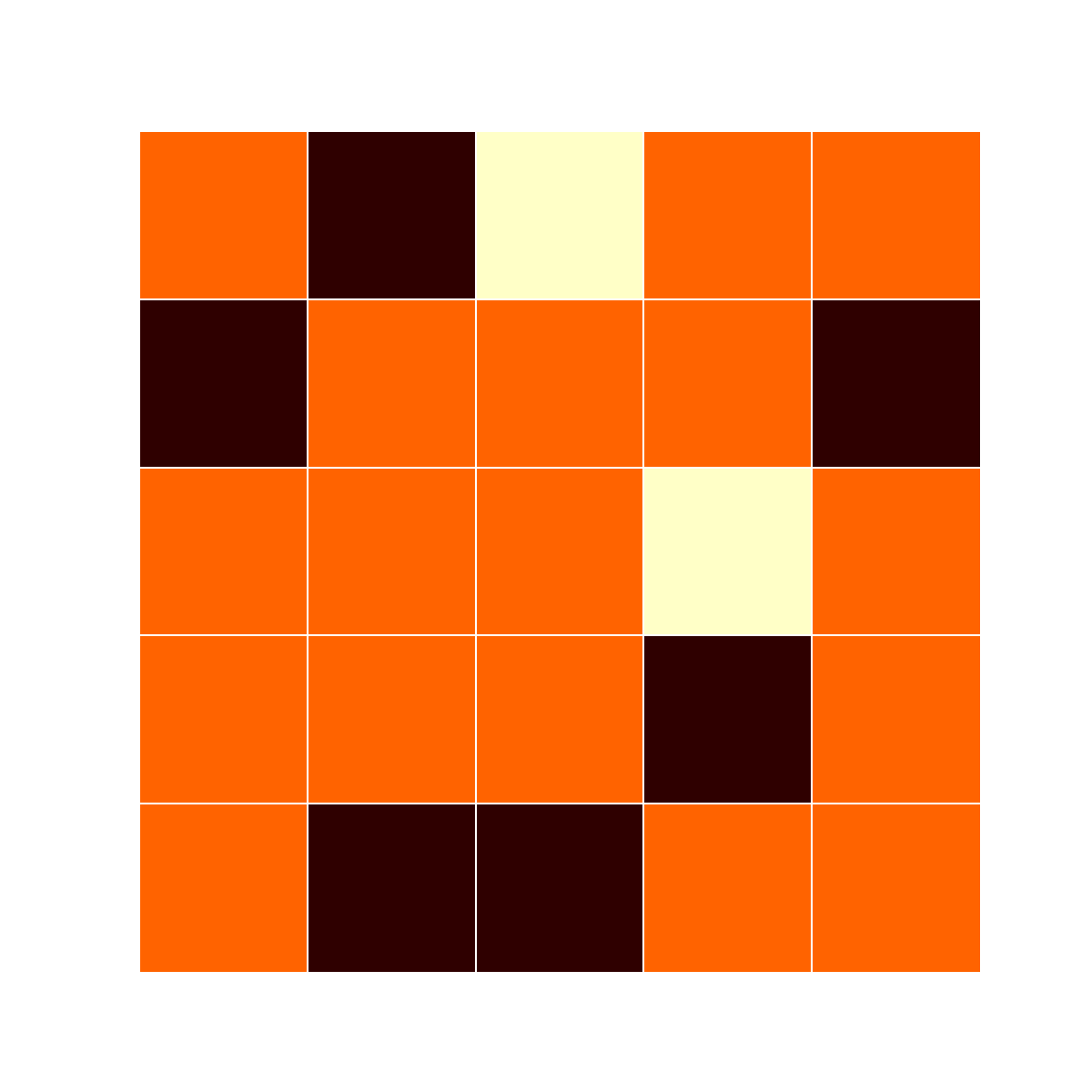}
    \end{subfigure}
    \begin{subfigure}[t]{0.31\textwidth}
        \centering
        \includegraphics[trim={1.1cm 0.3cm 1.1cm 0.5cm},clip=true,width=0.8\textwidth]{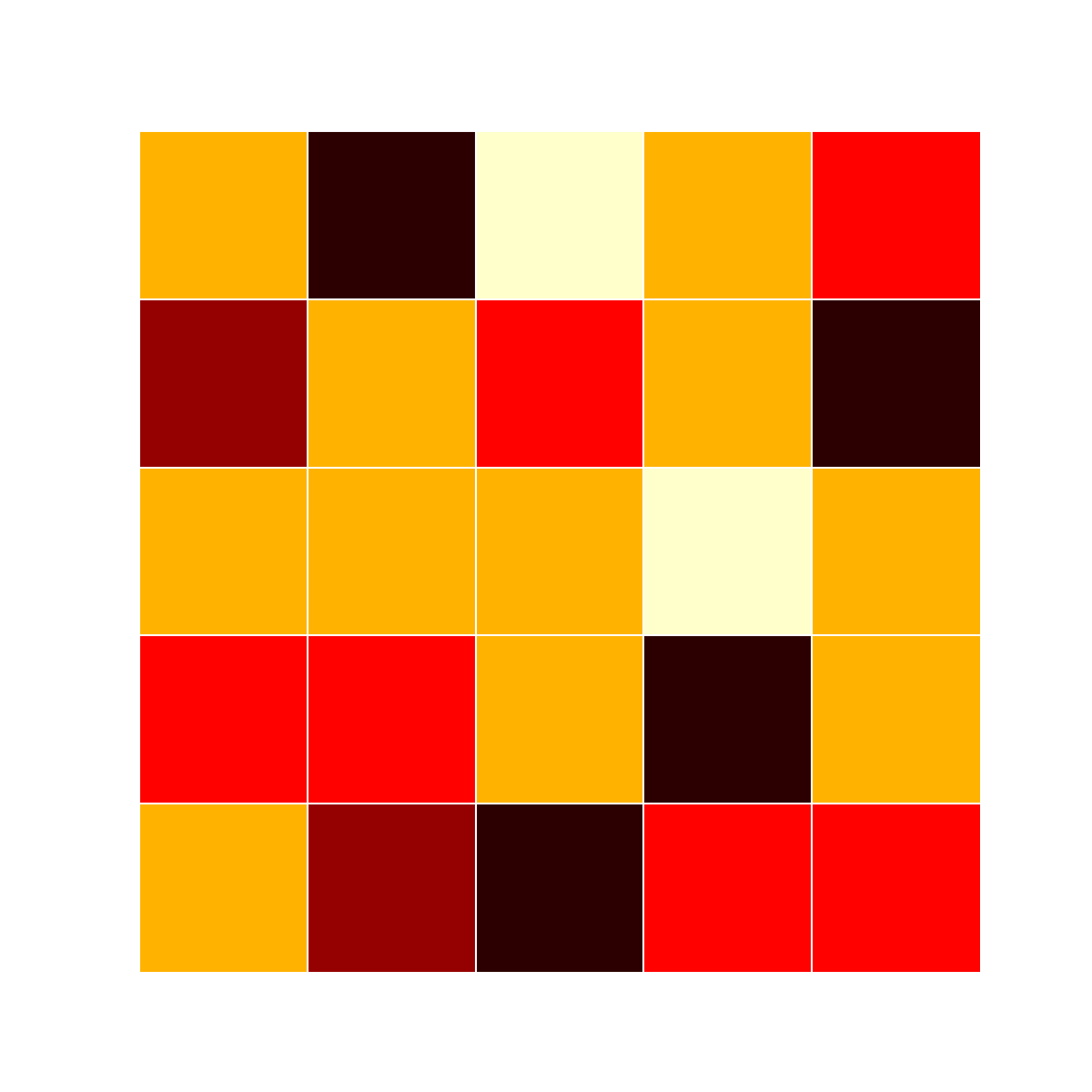}
    \end{subfigure}

    \begin{subfigure}[t]{0.31\textwidth}
        \centering
        \includegraphics[trim={0.5cm 0cm 0.2cm 0cm},clip=true,height=1.2cm]{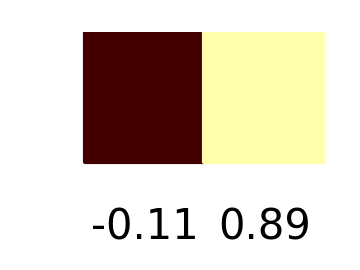}
        \caption{Random Grid Low}
    \end{subfigure}
    \begin{subfigure}[t]{0.31\textwidth}
        \centering
        \includegraphics[trim={0.5cm 0cm 0.2cm 0cm},clip=true,height=1.1cm]{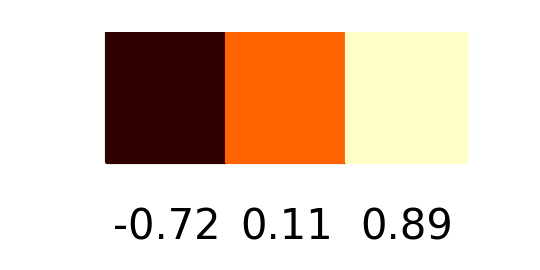}
        \caption{Random Grid Middle}
    \end{subfigure}
    \begin{subfigure}[t]{0.31\textwidth}
        \centering
        \includegraphics[trim={1.3cm 0cm 1.6cm 0cm},clip=true,height=1.1cm]{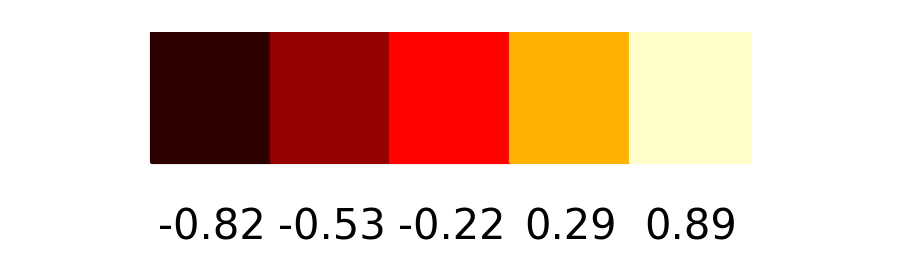}
        \caption{Random Grid High}
    \end{subfigure}

    \begin{subfigure}[t]{0.31\textwidth}
        \centering
        \includegraphics[trim={1.1cm 0.3cm 1.1cm 0.5cm},clip=true,width=0.8\textwidth]{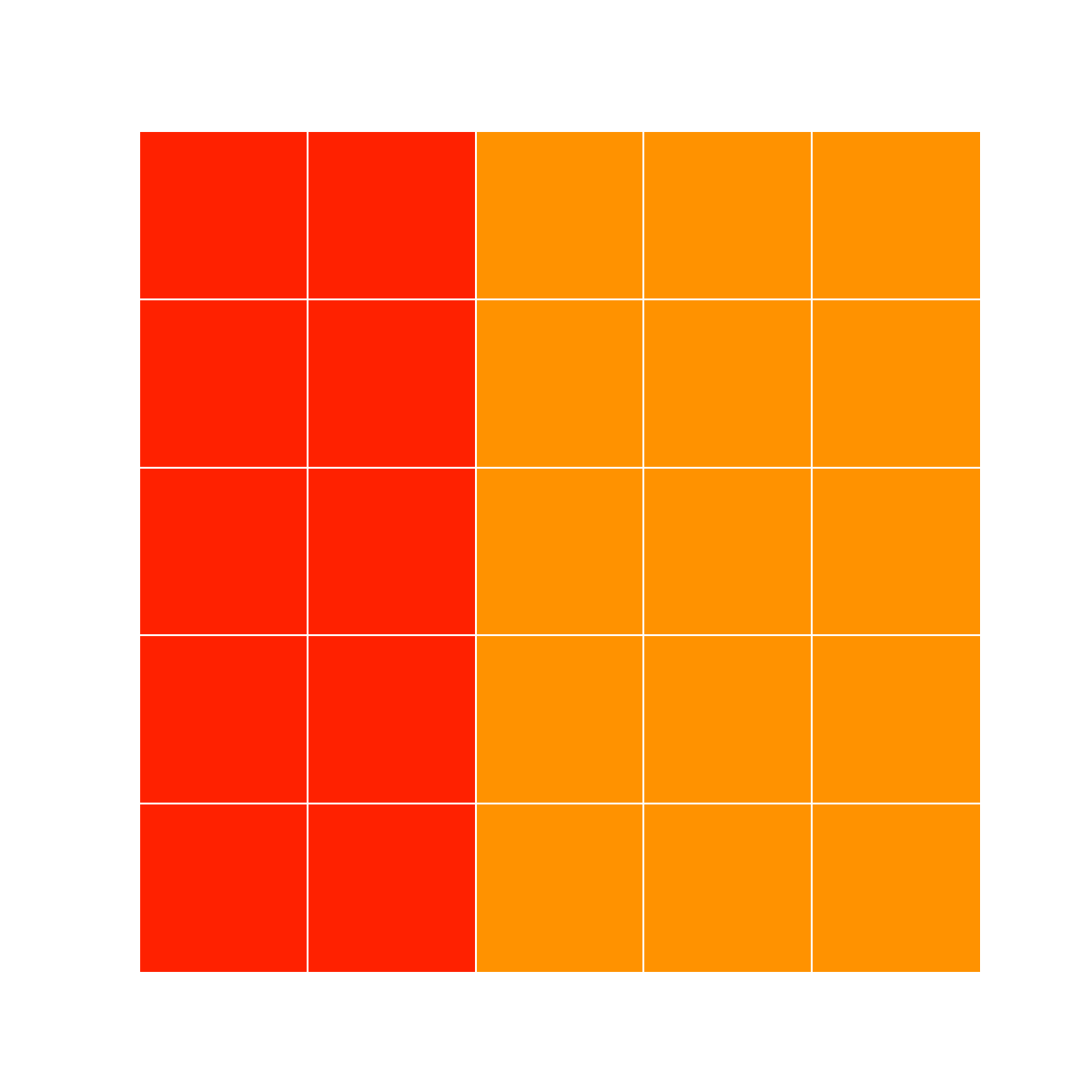}
    \end{subfigure}
    \begin{subfigure}[t]{0.31\textwidth}
        \centering
        \includegraphics[trim={1.1cm 0.3cm 1.1cm 0.5cm},clip=true,width=0.8\textwidth]{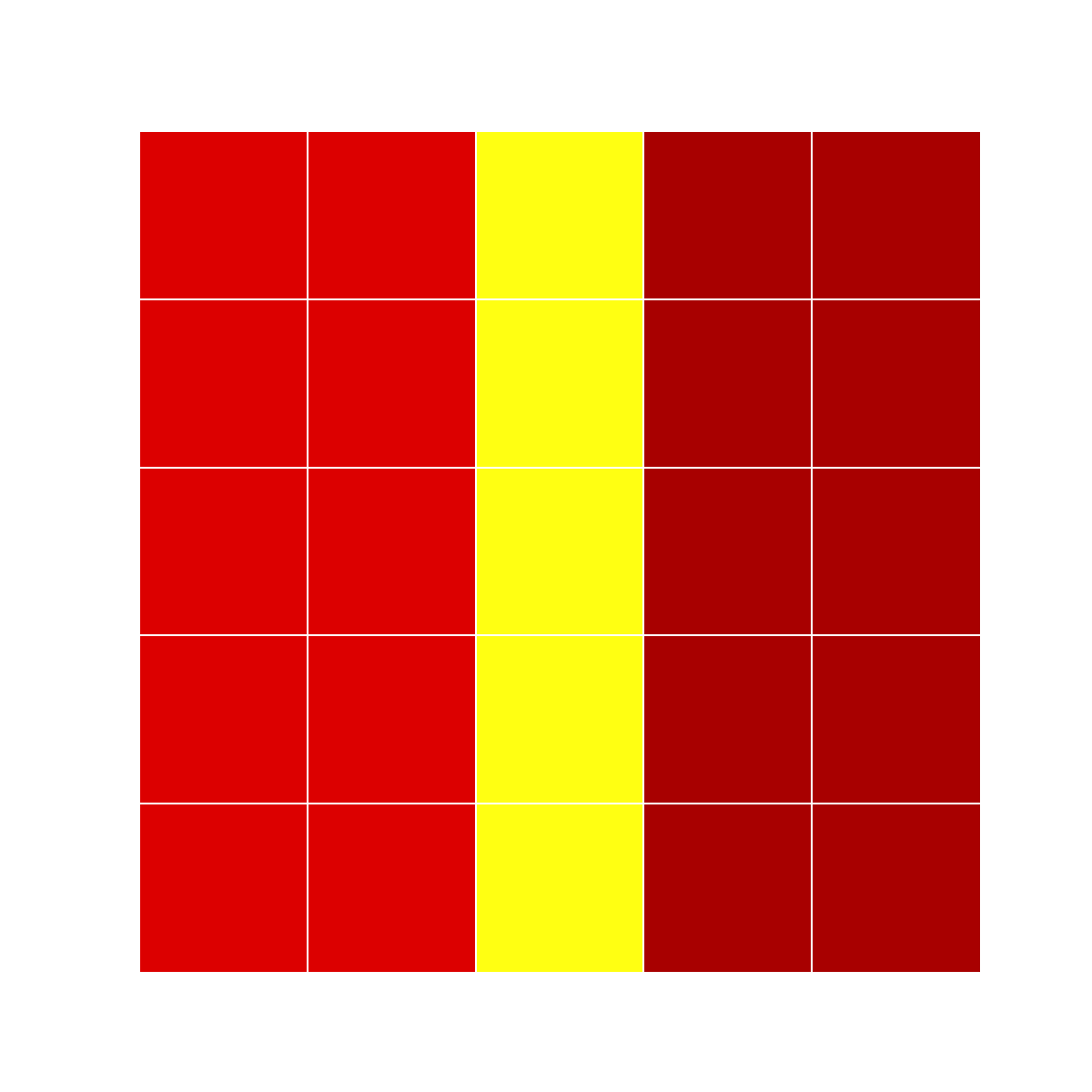}
    \end{subfigure}
    \begin{subfigure}[t]{0.31\textwidth}
        \centering
        \includegraphics[trim={1.1cm 0.3cm 1.1cm 0.5cm},clip=true,width=0.8\textwidth]{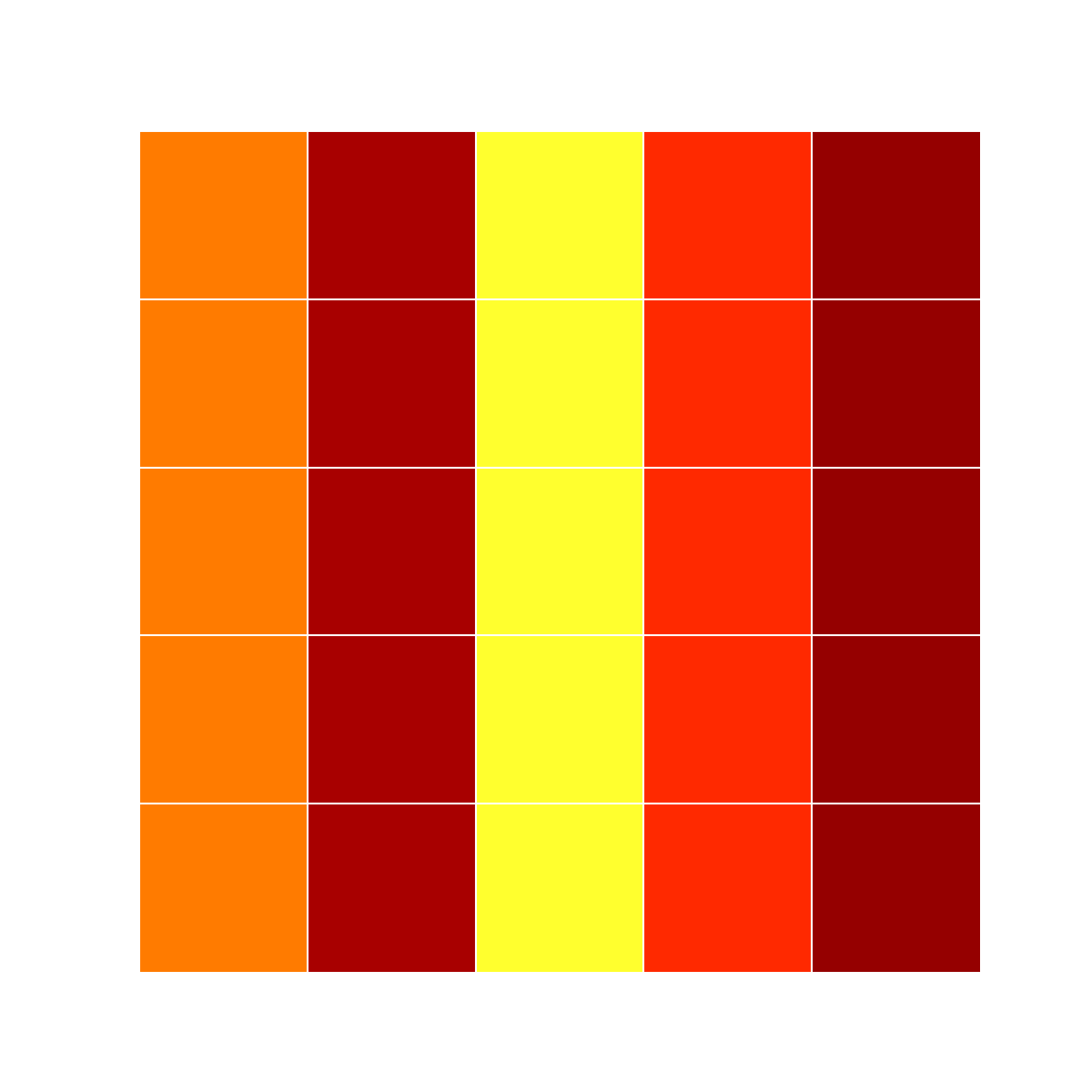}
    \end{subfigure}

    \begin{subfigure}[t]{0.31\textwidth}
        \centering
        \includegraphics[trim={0.5cm 0cm 0.2cm 0cm},clip=true,height=1.1cm]{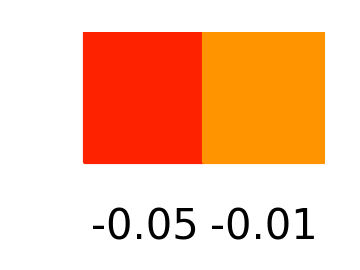}
        \caption{$X$ Coord. Low}
    \end{subfigure}
    \begin{subfigure}[t]{0.31\textwidth}
        \centering
        \includegraphics[trim={0.5cm 0cm 0.2cm 0cm},clip=true,height=1.1cm]{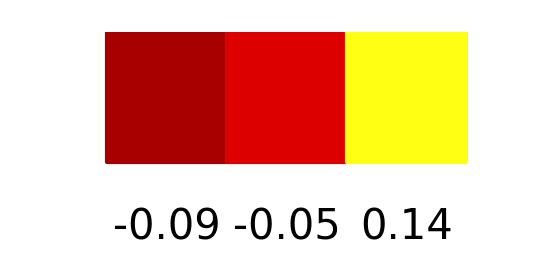}
        \caption{$X$ Coord. Middle}
    \end{subfigure}
    \begin{subfigure}[t]{0.31\textwidth}
        \centering
        \includegraphics[trim={0.7cm 0cm 0.9cm 0cm},clip=true,height=1.1cm]{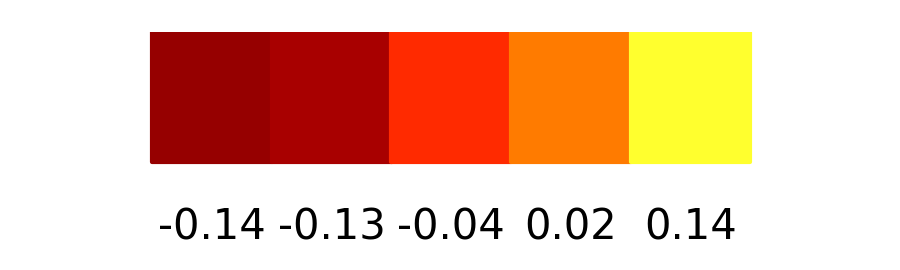}
        \caption{$X$ Coord. High}
    \end{subfigure}
    
    \begin{subfigure}[t]{0.31\textwidth}
        \centering
        \includegraphics[trim={1.1cm 0.3cm 1.1cm 0.5cm},clip=true,width=0.8\textwidth]{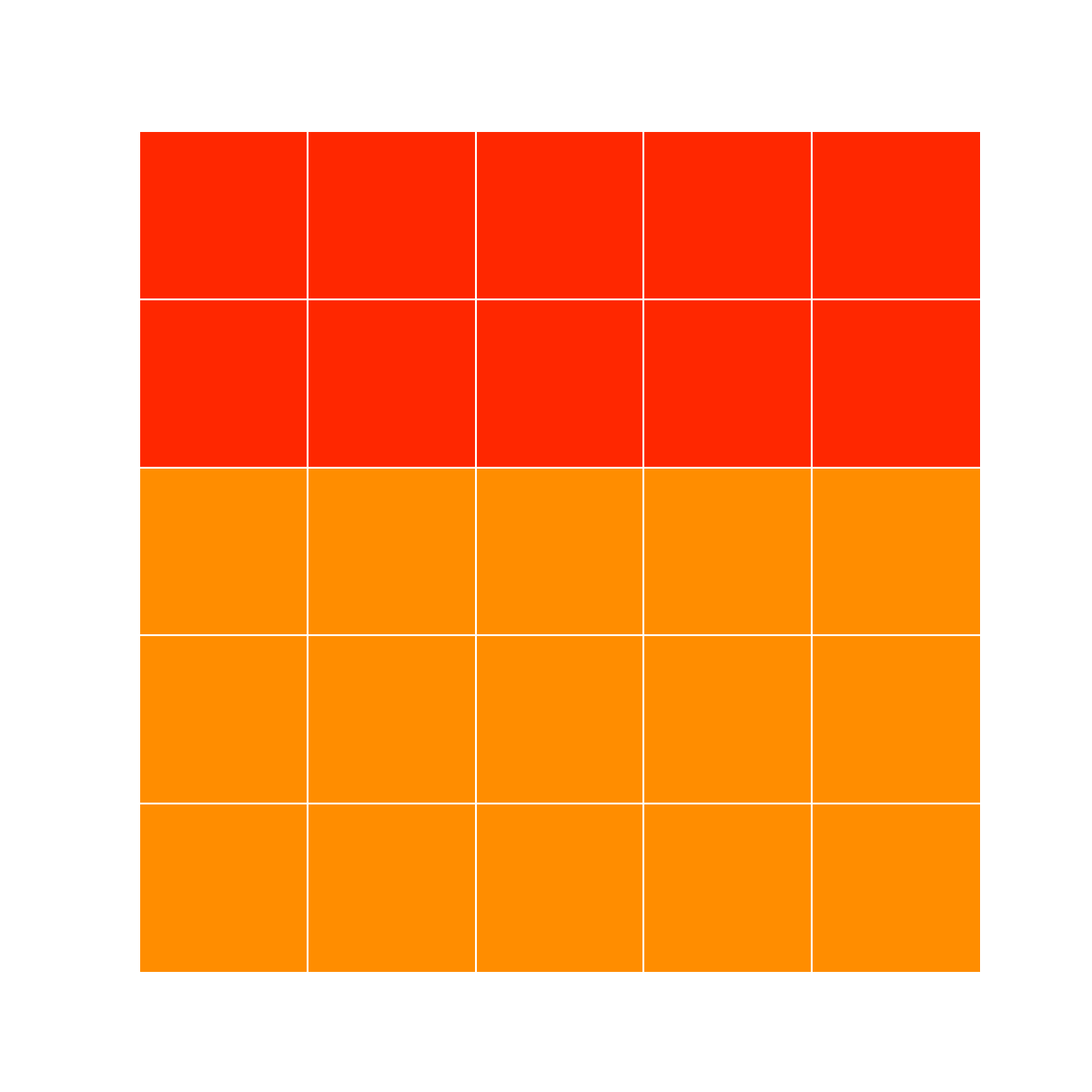}
    \end{subfigure}
    \begin{subfigure}[t]{0.31\textwidth}
        \centering
        \includegraphics[trim={1.1cm 0.3cm 1.1cm 0.5cm},clip=true,width=0.8\textwidth]{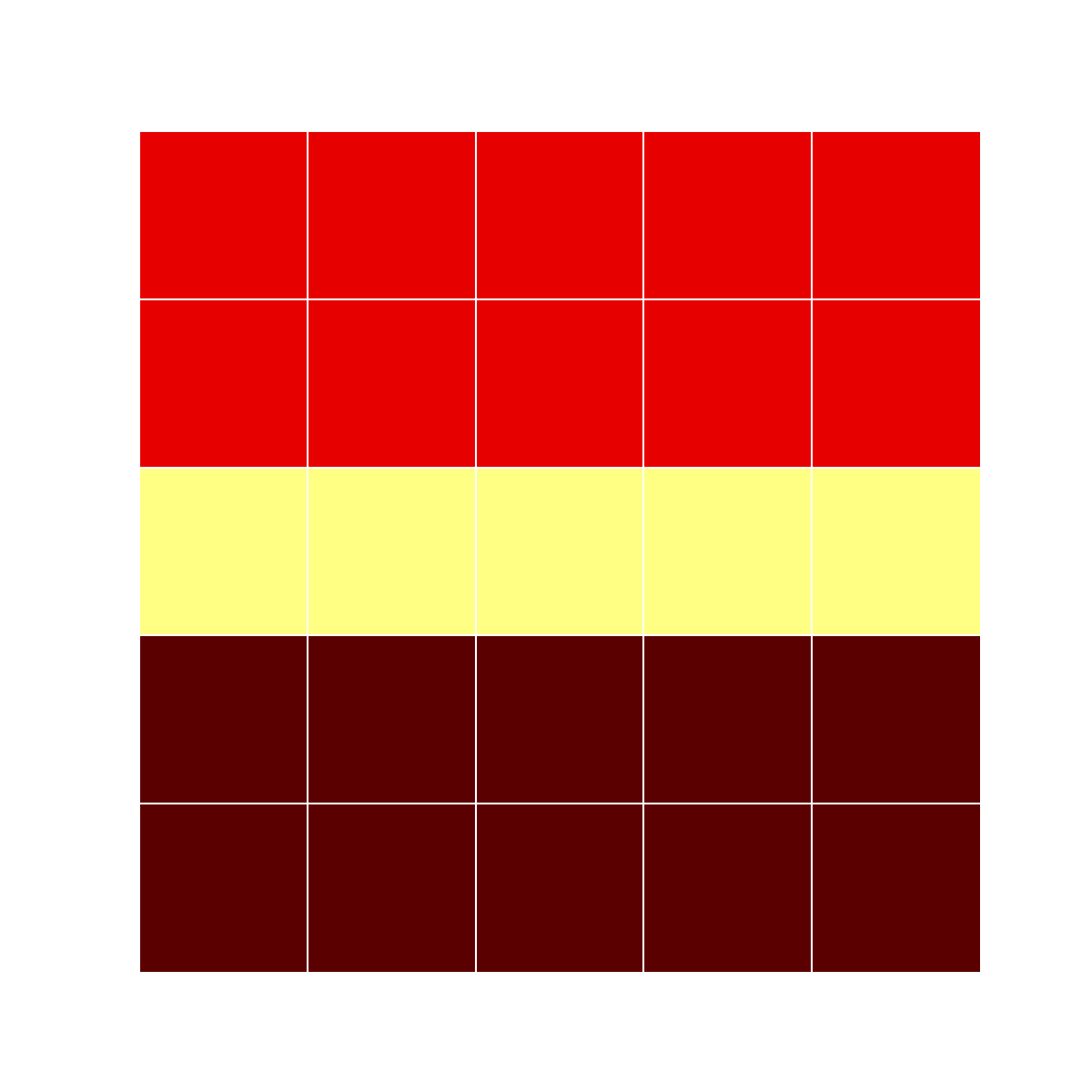}
    \end{subfigure}
    \begin{subfigure}[t]{0.31\textwidth}
        \centering
        \includegraphics[trim={1.1cm 0.3cm 1.1cm 0.5cm},clip=true,width=0.8\textwidth]{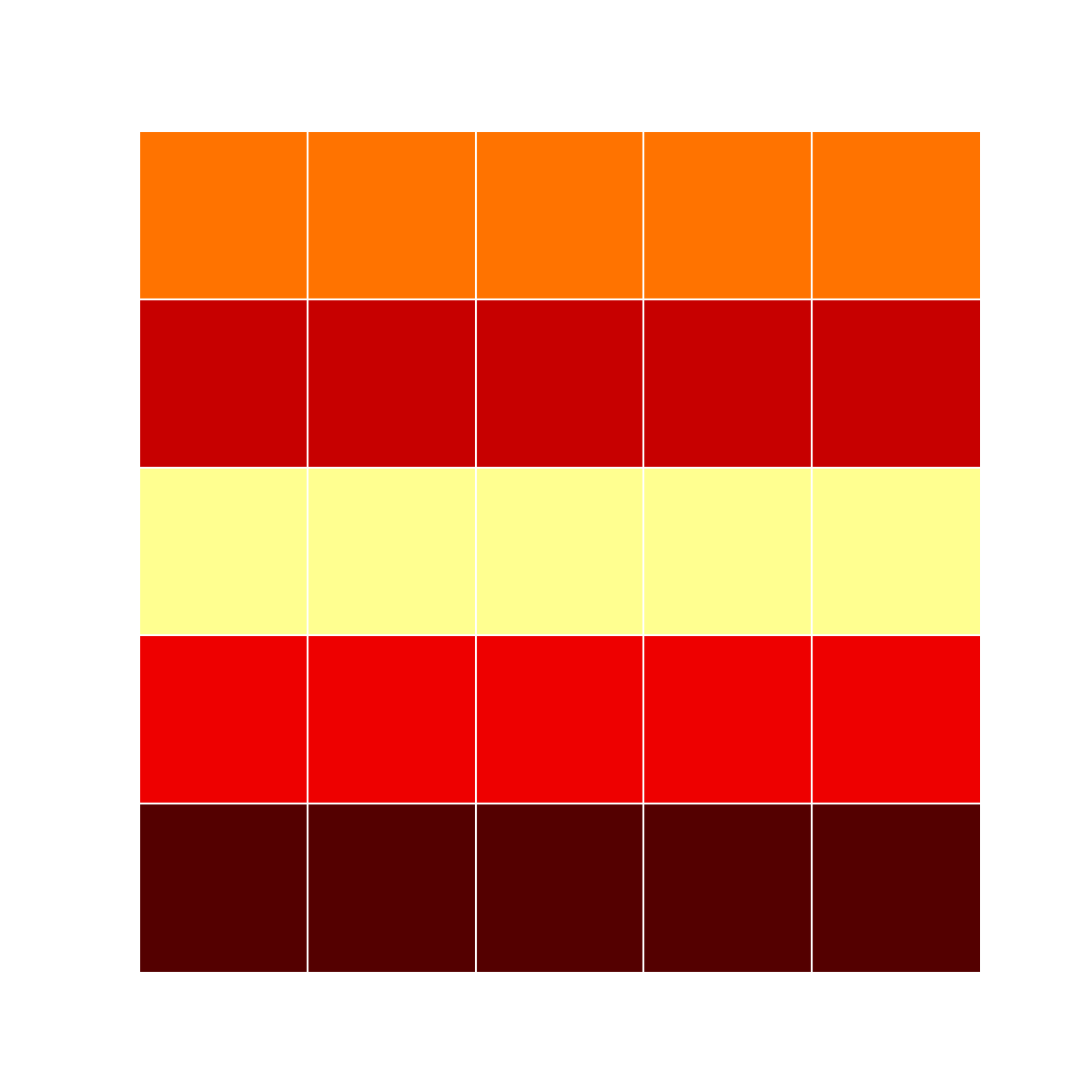}
    \end{subfigure}

    \begin{subfigure}[t]{0.31\textwidth}
        \centering
        \includegraphics[trim={0.5cm 0cm 0.2cm 0cm},clip=true,height=1.1cm]{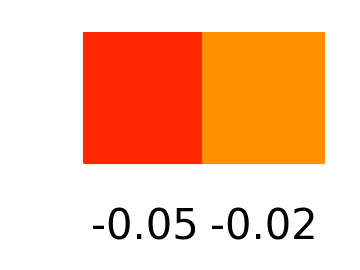}
        \caption{$Y$ Coord. Low}
    \end{subfigure}
    \begin{subfigure}[t]{0.31\textwidth}
        \centering
        \includegraphics[trim={0.5cm 0cm 0.2cm 0cm},clip=true,height=1.1cm]{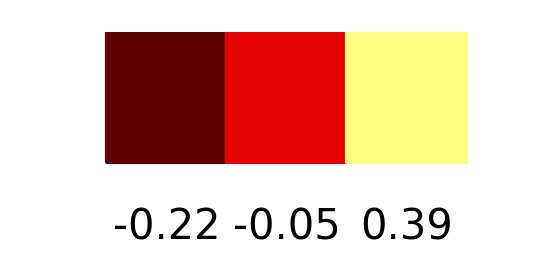}
        \caption{$Y$ Coord. Middle}
    \end{subfigure}
    \begin{subfigure}[t]{0.31\textwidth}
        \centering
        \includegraphics[trim={0.7cm 0cm 0.9cm 0cm},clip=true,height=1.1cm]{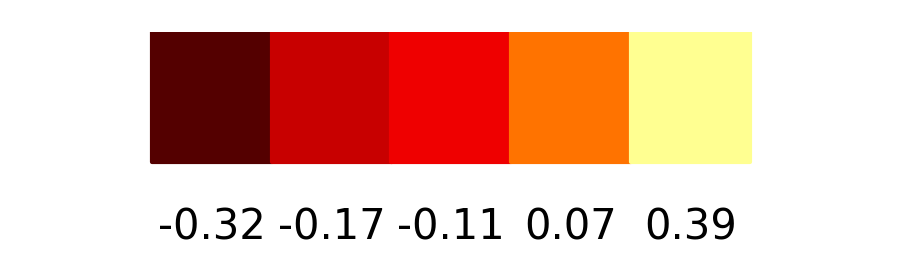}
        \caption{$Y$ Coord. High}
    \end{subfigure}
    \caption{Random grid abstractions for various training objectives (different rows) and complexity levels (different columns). As in the Manhattan grid abstractions, more complex abstractions captured finer-grained information, but the distortion for a given complexity level depended upon the training objective.}
    \label{fig:app_random_grid_viz}
\end{figure}

Figures~\ref{fig:app_manhattan_grid_viz} and \ref{fig:app_random_grid_viz} include visualizations of abstractions for the Manhattan and random grid domains, respectively.
Within each figure, we selected three checkpoints at low, medium, and high complexity (corresponding to different columns).
Note how the low-complexity checkpoints used fewer abstractions than higher-complexity checkpoints.

The different rows in each figure reflect abstractions generated using different training objectives in the IB process.
Recall from Section~\ref{sec:ib_method} that we used different training objectives when generating abstractions to induce different distortions for the same complexity.
Abstractions based on different reward functions are included as different rows in Figures~\ref{fig:app_manhattan_grid_viz} and \ref{fig:app_random_grid_viz}.
For example, the top row of Figure~\ref{fig:app_manhattan_grid_viz} shows abstractions generated using the Manhattan reward function; the second row, however, depicts abstractions generated using the $x$ coordinate of each location.
Even as the $x$-based abstractions increased in complexity, until each $x$ value was represented separately, such abstractions remained poor for predicting the actual Manhattan reward in the grid.
Similar patterns were true in the random grid as well (Figure~\ref{fig:app_random_grid_viz}).
Overall, the visualizations of abstractions in these grid domains confirm intuitions that 1) increasing complexity led to finer-grained abstractions and 2) abstractions generated using sub-optimal training objectives led to poor reward prediction, and therefore high distortion.

\newpage

\begin{figure}
    \centering
    \begin{subfigure}[t]{0.31\textwidth}
        \centering
        \includegraphics[trim={1.1cm 0.3cm 1.1cm 0.5cm},clip=true,width=\textwidth]{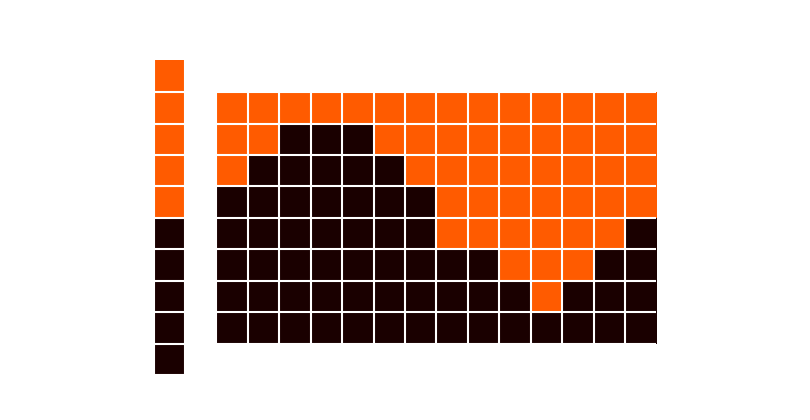}
    \end{subfigure}
    \begin{subfigure}[t]{0.31\textwidth}
        \centering
        \includegraphics[trim={1.1cm 0.3cm 1.1cm 0.5cm},clip=true,width=\textwidth]{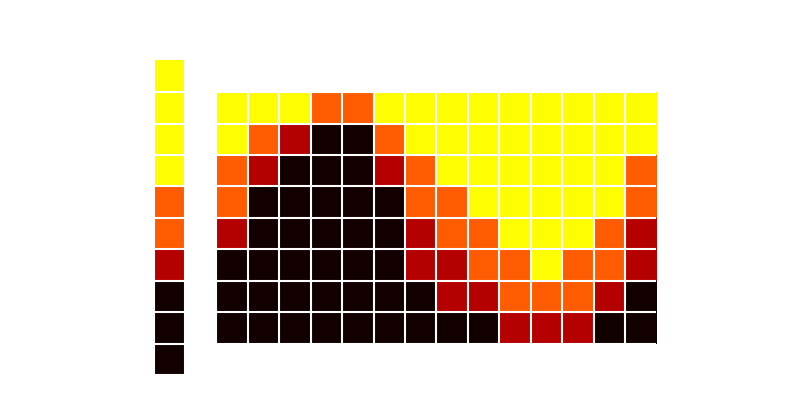}
    \end{subfigure}
    \begin{subfigure}[t]{0.31\textwidth}
        \centering
        \includegraphics[trim={1.1cm 0.3cm 1.1cm 0.5cm},clip=true,width=\textwidth]{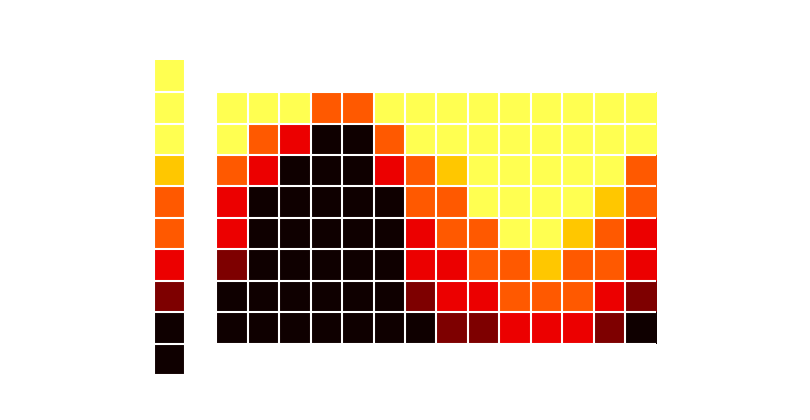}
    \end{subfigure}

    \begin{subfigure}[t]{0.31\textwidth}
        \centering
        \includegraphics[trim={0.5cm 0cm 0.2cm 0cm},clip=true,height=1.5cm]{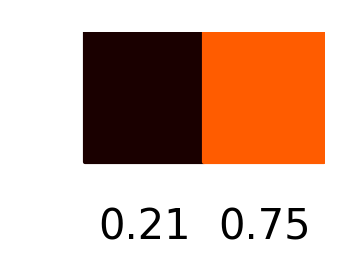}
        \caption{Blue Cont. Low}
    \end{subfigure}
    \begin{subfigure}[t]{0.31\textwidth}
        \centering
        \includegraphics[trim={0.5cm 0cm 0.2cm 0cm},clip=true,height=1.5cm]{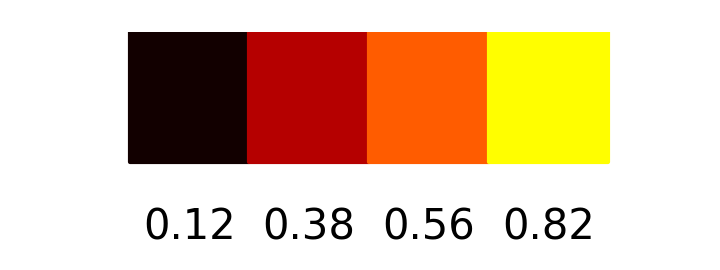}
        \caption{Blue Cont. Middle}
    \end{subfigure}
    \begin{subfigure}[t]{0.31\textwidth}
        \centering
        \includegraphics[trim={1.3cm 0cm 1.6cm 0cm},clip=true,height=1.5cm]{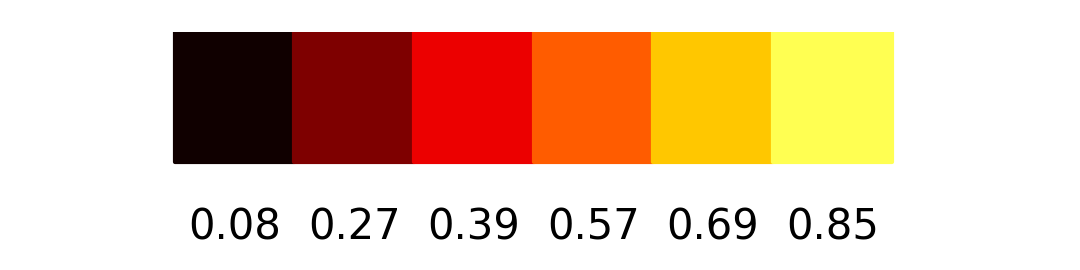}
        \caption{Blue Cont. High}
    \end{subfigure}

    \begin{subfigure}[t]{0.31\textwidth}
        \centering
        \includegraphics[trim={1.1cm 0.3cm 1.1cm 0.5cm},clip=true,width=\textwidth]{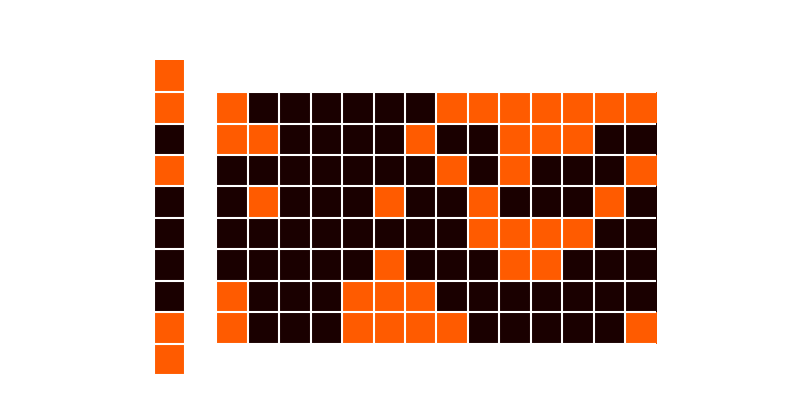}
    \end{subfigure}
    \begin{subfigure}[t]{0.31\textwidth}
        \centering
        \includegraphics[trim={1.1cm 0.3cm 1.1cm 0.5cm},clip=true,width=\textwidth]{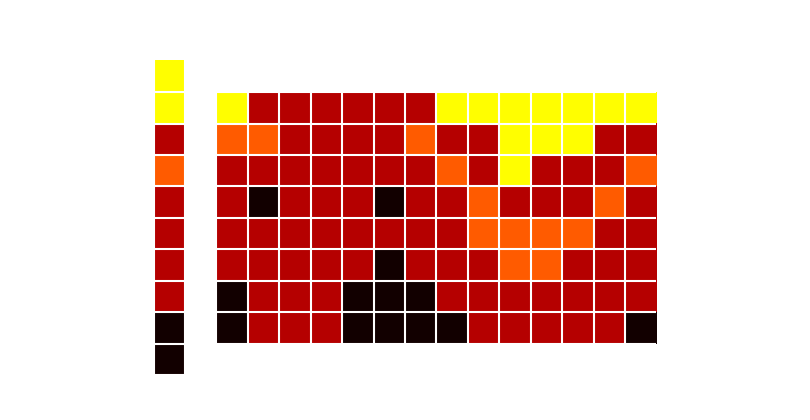}
    \end{subfigure}
    \begin{subfigure}[t]{0.31\textwidth}
        \centering
        \includegraphics[trim={1.1cm 0.3cm 1.1cm 0.5cm},clip=true,width=\textwidth]{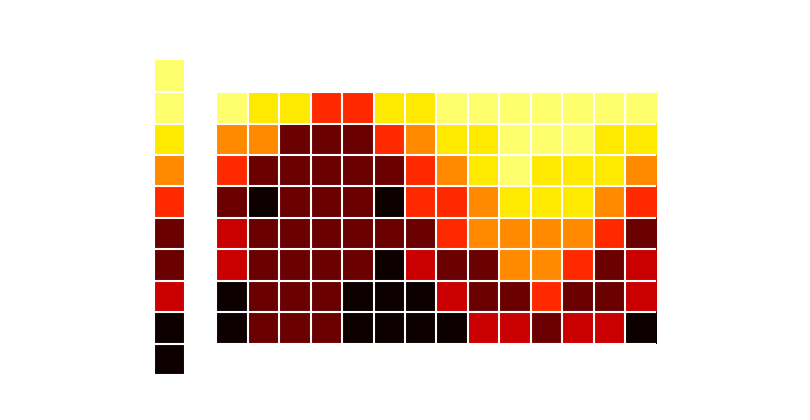}
    \end{subfigure}

    \begin{subfigure}[t]{0.31\textwidth}
        \centering
        \includegraphics[trim={0.5cm 0cm 0.2cm 0cm},clip=true,height=1.5cm]{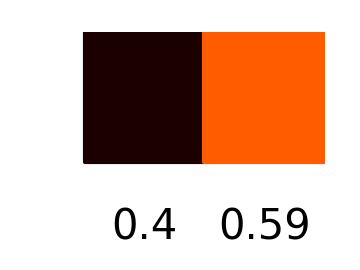}
        \caption{Blue Disc. Low}
    \end{subfigure}
    \begin{subfigure}[t]{0.31\textwidth}
        \centering
        \includegraphics[trim={0.5cm 0cm 0.2cm 0cm},clip=true,height=1.5cm]{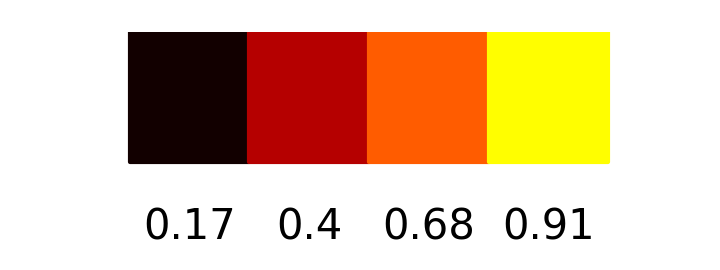}
        \caption{Blue Disc. Middle}
    \end{subfigure}
    \begin{subfigure}[t]{0.31\textwidth}
        \centering
        \includegraphics[trim={1.7cm 0cm 1.9cm 0cm},clip=true,height=1.5cm]{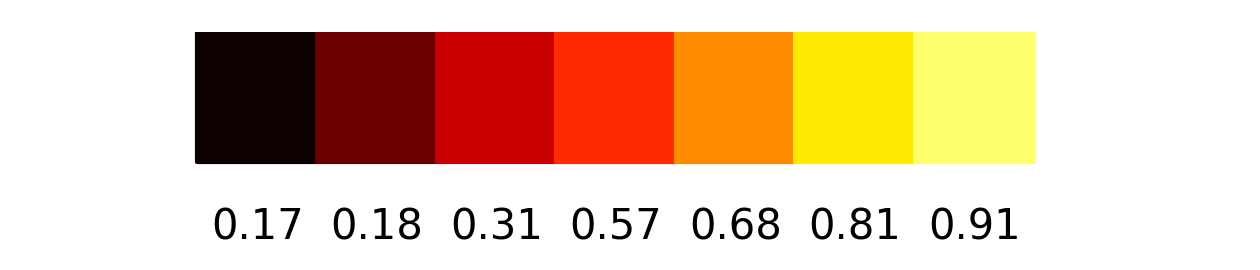}
        \caption{Blue Disc. High}
    \end{subfigure}
    
    \begin{subfigure}[t]{0.31\textwidth}
        \centering
        \includegraphics[trim={1.1cm 0.3cm 1.1cm 0.5cm},clip=true,width=\textwidth]{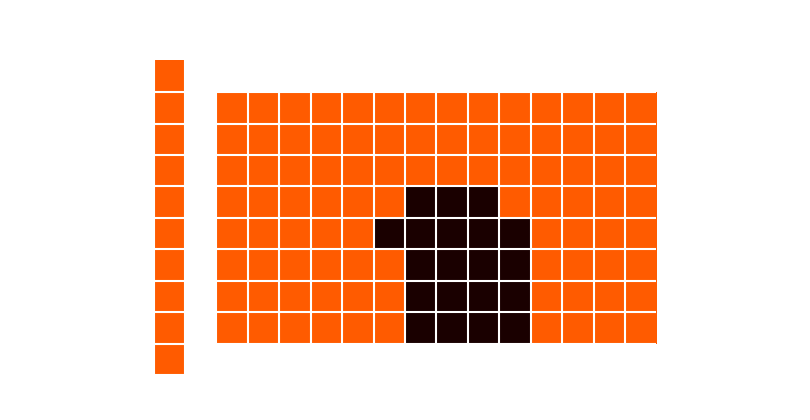}
    \end{subfigure}
    \begin{subfigure}[t]{0.31\textwidth}
        \centering
        \includegraphics[trim={1.1cm 0.3cm 1.1cm 0.5cm},clip=true,width=\textwidth]{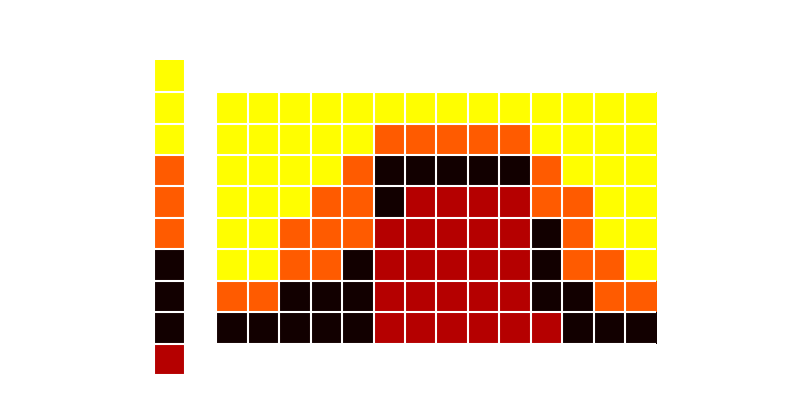}
    \end{subfigure}
    \begin{subfigure}[t]{0.31\textwidth}
        \centering
        \includegraphics[trim={1.1cm 0.3cm 1.1cm 0.5cm},clip=true,width=\textwidth]{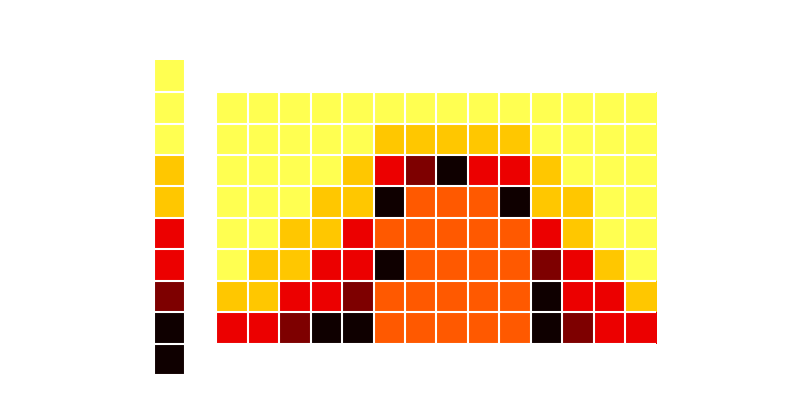}
    \end{subfigure}

    \begin{subfigure}[t]{0.31\textwidth}
        \centering
        \includegraphics[trim={0.5cm 0cm 0.2cm 0cm},clip=true,height=1.5cm]{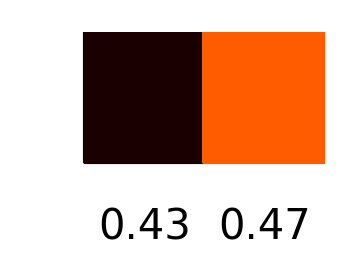}
        \caption{Red Cont. Low}
    \end{subfigure}
    \begin{subfigure}[t]{0.31\textwidth}
        \centering
        \includegraphics[trim={0.5cm 0cm 0.2cm 0cm},clip=true,height=1.5cm]{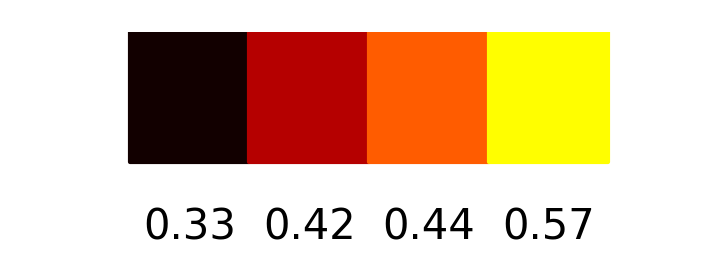}
        \caption{Red Cont. Middle}
    \end{subfigure}
    \begin{subfigure}[t]{0.31\textwidth}
        \centering
        \includegraphics[trim={1.3cm 0cm 1.7cm 0cm},clip=true,height=1.5cm]{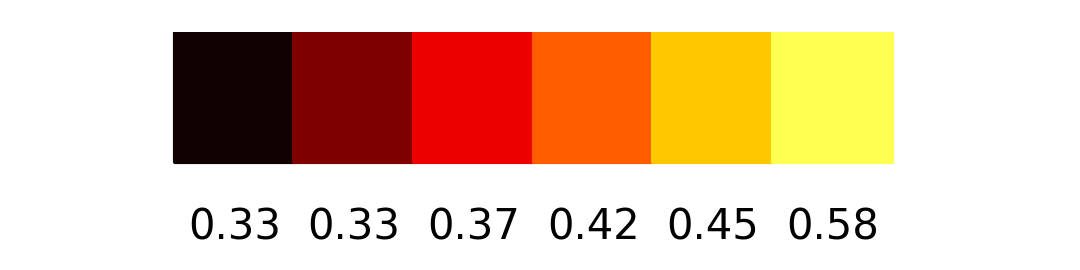}
        \caption{Red Cont. High}
    \end{subfigure}
    \caption{Color abstractions evaluated on the continuous blue reward function. Increasing complexity (left to right) increased the number of abstractions. The distortion associated with such abstractions varied greatly, however, depending upon the training objective used when generating abstractions. Using the continuous blue reward to generate abstraction (top row) led to evenly spaced abstractions with little distortion. Different training objectives (discontinuous blue in the second row, continuous red in the third) led to suboptimal abstractions with higher distortion.}
    \label{fig:app_color_viz}
\end{figure}

Figure~\ref{fig:app_color_viz} includes visualizations of abstractions from the color domain much like the previous visualizations of grid-based abstractions.
Abstractions were evaluated according to the continuous blue reward function; the heatmap used for visualization shows the average blue value of each abstraction.
As before, we selected three checkpoints at low, medium, and high complexity (corresponding to different columns).

The different rows in Figure~\ref{fig:app_color_viz} reflect abstractions generated using different training objectives.
In the top row, we used the continuous blue reward function -- the same function used for evaluation.
In the second row, we used the discontinuous blue reward function, and in the third row we used a color's red value to generate abstractions.

As in the grid domains, we found that 1) increasing complexity led to more fine-grained abstractions and 2) using different training objectives in the IB process led to sub-optimal abstractions that resulted in high distortion.
For example, using the discontinuous blue reward function (second row) at high complexity (Figure~\ref{fig:app_color_viz}~f), we observed that two abstractions have nearly identical average rewards of 0.17 and 0.18.
Thus, the additional complexity incurred by having more abstractions comes with barely any decrease in distortion.
This suboptimality is even more pronounced for abstractions generated with the continuous red reward function (bottom row).
Using just two abstractions at low complexity (Figure~\ref{fig:app_color_viz}~g), the color space is partitioned into two groups that are meaningful for predicting the redness of a color, but with almost no difference in blue value (mean values of 0.43 and 0.47).

\clearpage
\section{Survey Questions}
\label{app:survey_questions}
Here, we include examples of each question asked of participants in the grid navigation survey and the color survey. Each participant answered two of each question within the particular survey they were responding to: one for a continuous reward function and one for a discontinuous reward function.

\subsection{Grid Navigation Domain}
\label{app:survey_questions_grid}
\begin{figure}
    \centering
    \includegraphics[width=\textwidth]{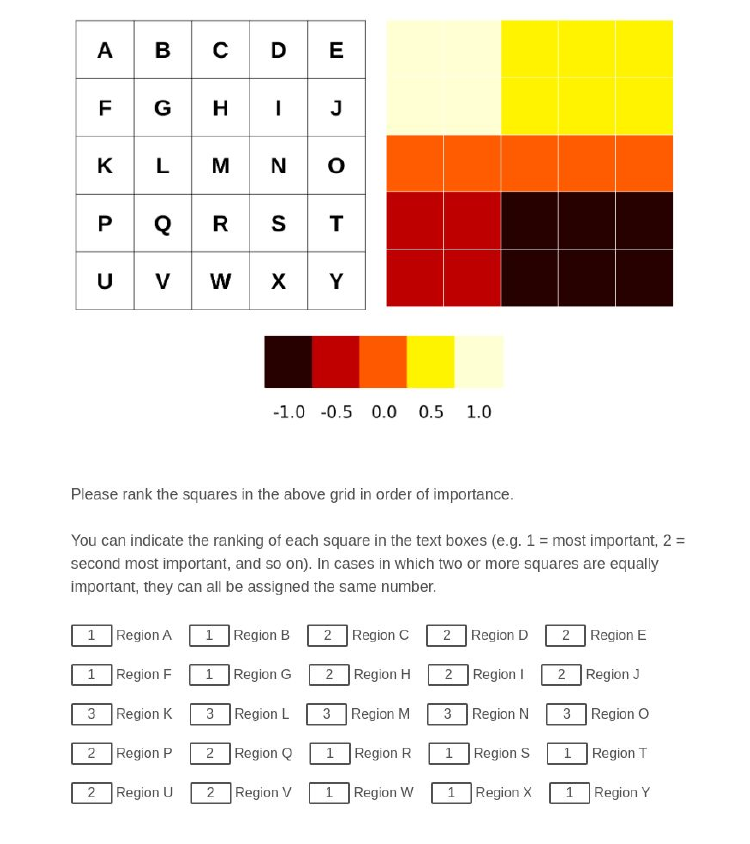}
        \caption{Example of the \textit{feature rank} question in the grid navigation domain. The top left grid designates regions of the grid which are ranked by participants when answering the question. The top right grid depicts the abstract reward regions. The reward values associated with each reward region are depicted in the five numbered swatches below the grids. At the bottom, sample responses for the \textit{feature rank} question are provided based on the given abstract grid.}
        \label{fig:grid_nav_fr}
\end{figure}

\begin{figure}

    \begin{subfigure}[t]{0.6\textwidth}
    \centering
    \includegraphics[width=\linewidth]{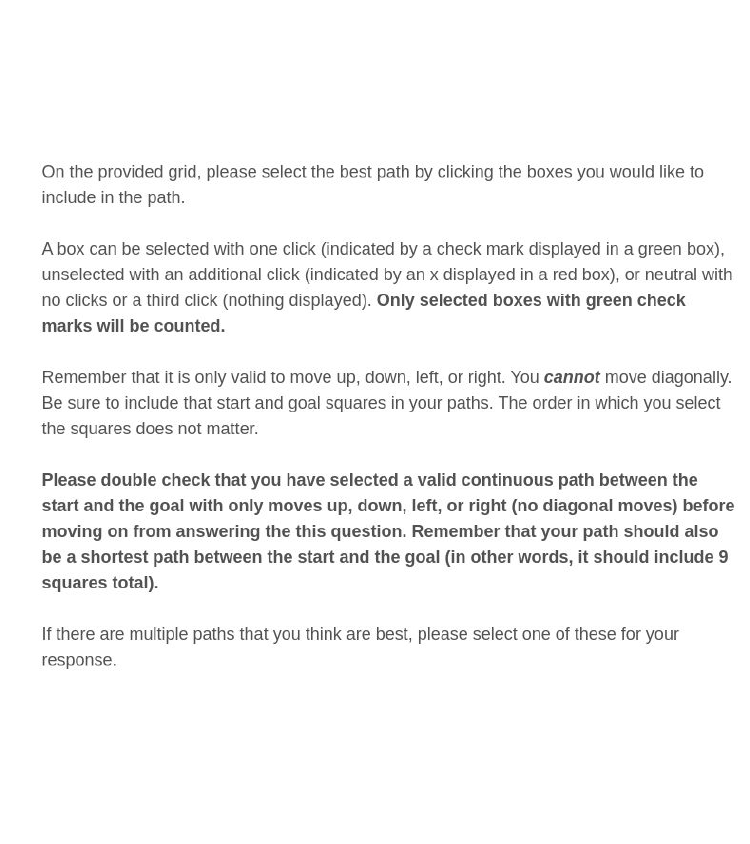}
    \end{subfigure}
    
    \begin{subfigure}[t]{0.45\textwidth}
    \includegraphics[trim={0cm 0cm 0cm 0cm},clip=true,width=\linewidth]{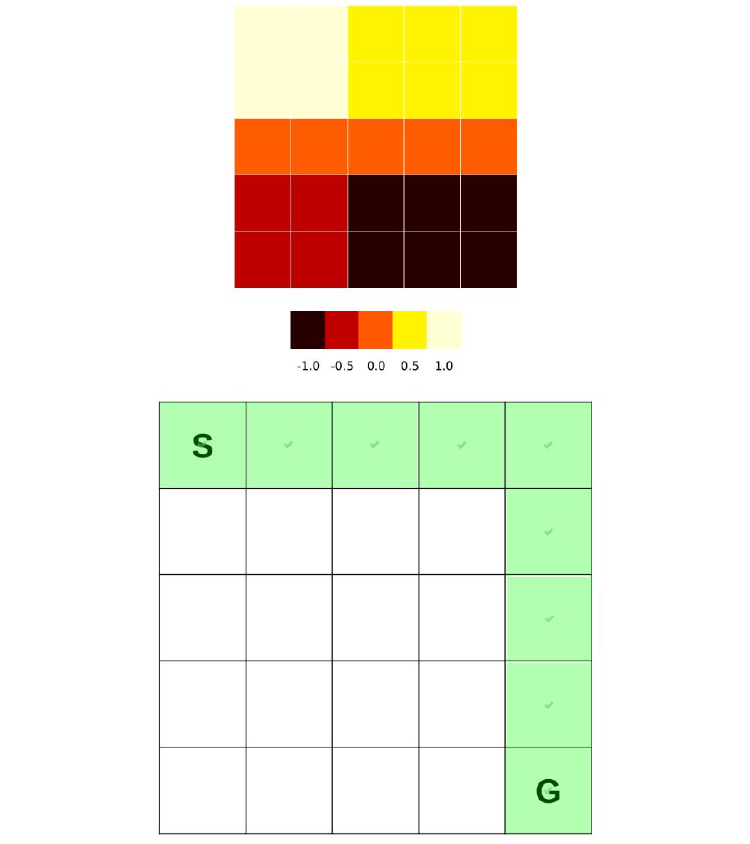}
    \end{subfigure}
    \caption{Example of the \textit{best demonstration} question in the grid navigation domain. The same explanation (grid with the abstract reward regions) is shown to participants as for the \textit{feature rank} question. For this question, participants must select the reward-maximizing path through the grid at the bottom of the figure (depicted through the selected green boxes with check marks).}
    \label{fig:grid_nav_bp}
\end{figure}

Figures \ref{fig:grid_nav_fr} and \ref{fig:grid_nav_bp} depict the \textit{feature rank (FR)} and the \textit{best demonstration (BD)} questions as they were presented to participants in the grid navigation domain. Here, the abstract explanation shown in each question (via a heat map indicating the values of the grid squares) represents the ground truth reward function. In this case, grid squares in the upper left corner have values of +1.0, those in the upper right have values of +0.5, those in the middle row have values of 0.0, those in the lower left have values of -0.5, and those in the lower right have values of -1.0.

\clearpage
\subsection{Color Domain}
\label{app:survey_questions_color}
\begin{figure}
    \begin{subfigure}[t]{0.6\textwidth}
    \centering
    \includegraphics[width=\linewidth]{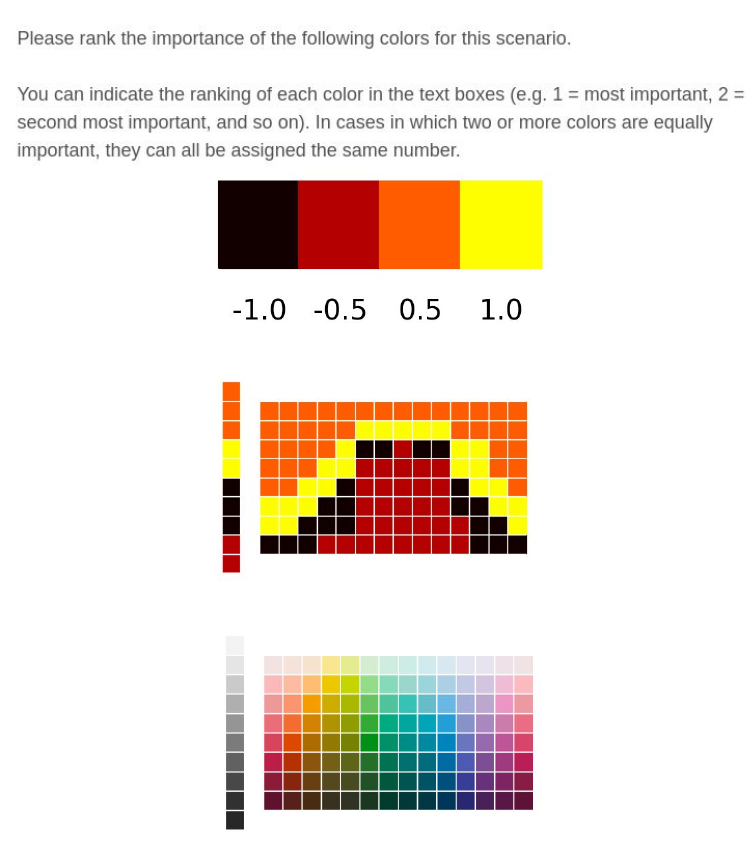}
    \end{subfigure}
    
    \begin{subfigure}[t]{0.6\textwidth}
    \centering
    \includegraphics[trim={0cm 7cm 0cm 0cm},clip=true,width=\textwidth]{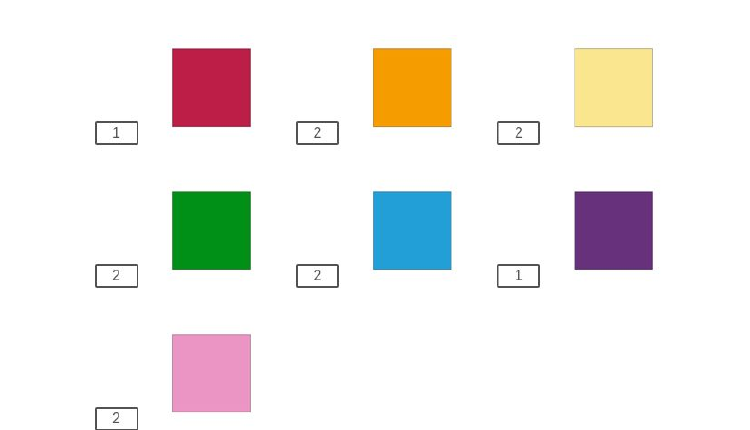}
    \end{subfigure}
    \caption{Example of the \textit{feature rank} question in the color domain. The heat map and the associated numbered color swatches at the top indicate the abstract regions of colors from the color grid below and their corresponding reward values. At the bottom, sample responses for the \textit{feature rank} question (a set of ranked colors from the color grid) are provided based on the given abstraction of the color grid.}
    \label{fig:color_fr}
\end{figure}

\begin{figure}
    \begin{subfigure}[t]{0.65\textwidth}
    \centering
    \includegraphics[width=\linewidth]{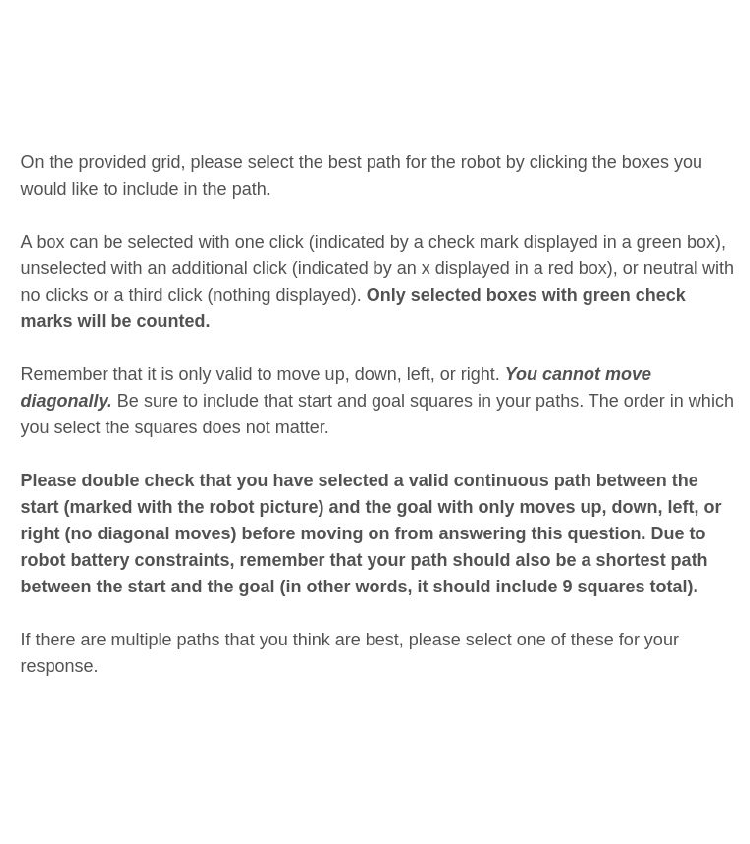}
    \end{subfigure}
    
    \begin{subfigure}[t]{0.35\textwidth}
    \includegraphics[width=\linewidth]{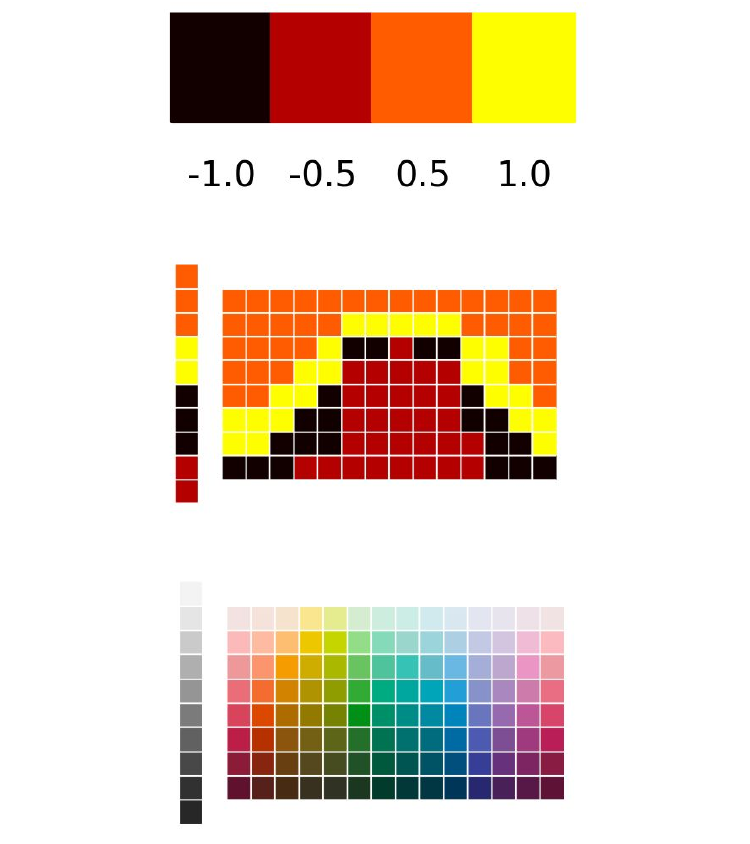}
    \end{subfigure}
    ~
    \begin{subfigure}[t]{0.35\textwidth}
    \includegraphics[width=\linewidth]{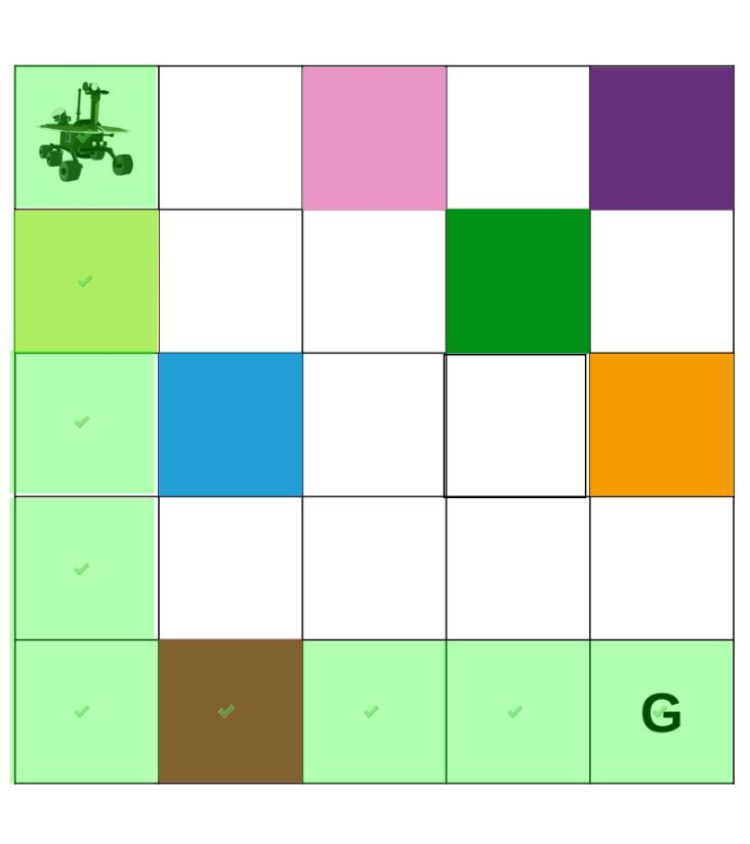}
    \end{subfigure}
    \caption{Example of the \textit{best demonstration} question in the color domain. The same explanation (color grid with the abstract reward regions) is shown to participants as for the \textit{feature rank} question. Here, participants must select the path through the grid at the bottom of the figure which maximizes the value of the collected samples, which are indicated by different colors from the original color grid. The selected path is, again, marked by green boxes with check marks.}
    \label{fig:color_bp}
\end{figure}

Figures \ref{fig:color_fr} and \ref{fig:color_bp} depict the \textit{feature rank (FR)} and the \textit{best demonstration (BD)} questions as they were presented to participants in the color domain. The abstract explanation shown in each question (via a heat map indicating the values of each color in the color grid) represents the ground truth reward function. Colors in the top part of the grid have values of +0.5, those below that have values of +1.0, those below that have values of -1.0, and those at the bottom have values of -0.5.

\clearpage
\subsection{Workload Questions}
\label{app:survey_questions_workload}
\begin{figure}
    \centering
    \includegraphics[width=\textwidth]{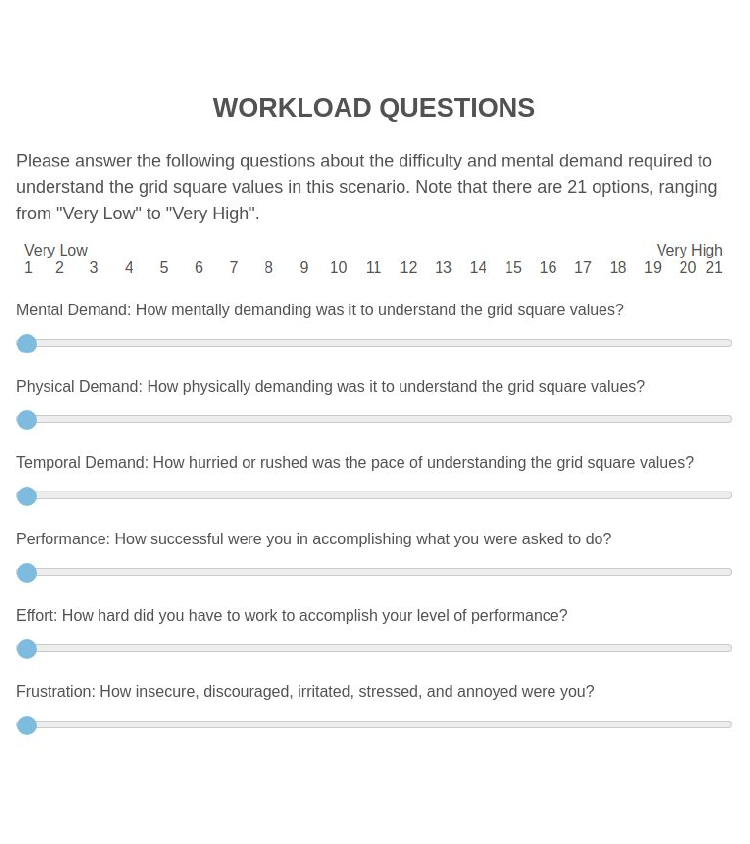}
        \caption{Questions from the NASA TLX survey \cite{hart1988development}. The questions above are depicted for the grid navigation domain.}
    \label{fig:grid_nav_workload}
\end{figure}

The workload questions that we applied in our surveys were drawn from the NASA TLX survey \cite{hart1988development}. The set of questions as they were displayed in the grid navigation domain are depicted in Figure \ref{fig:grid_nav_workload}. The questions shown to participants in the color domain were nearly identical with minor wording changes based on the differences between the two domains.

\clearpage
\subsection{Subjective Assessment Questions}
\label{app:survey_questions_subjective}
\begin{figure}
    \centering
    \includegraphics[width=\textwidth]{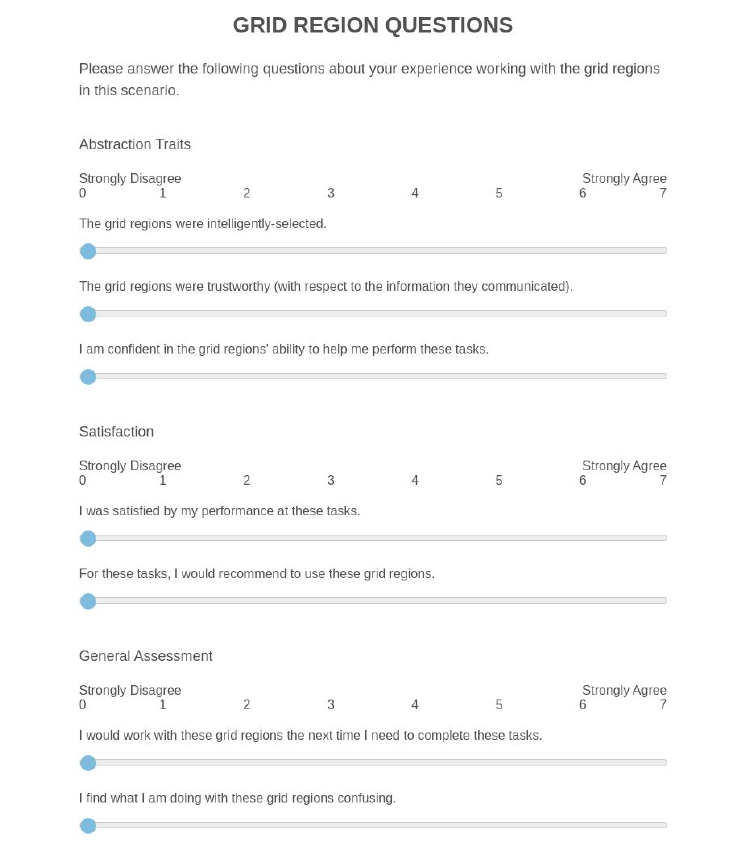}
        \caption{Subjective assessment questions adapted from \citet{hoffman2019evaluating}. The questions above are depicted for the grid navigation domain, and were largely similar in the color domain.}
        \label{fig:grid_nav_subj}
\end{figure}

The subjective assessment questions that we applied in our surveys were adapted from the scale for team fluency proposed by  \citet{hoffman2019evaluating}. The set of questions as they were displayed in the grid navigation domain are depicted in Figure \ref{fig:grid_nav_subj}. The questions shown to participants in the color domain were, again, nearly identical with minor wording changes based on the differences between the two domains.

\end{document}